\documentclass[twoside]{article}
\usepackage{microtype}
\usepackage{graphicx}
\usepackage{subcaption}
\usepackage{booktabs} % for professional tables
\usepackage[utf8]{inputenc}
\usepackage[ruled]{algorithm2e}
\usepackage{algorithmic}
\usepackage{amsmath}
\usepackage{mathtools}  % for :=
\usepackage{amsfonts}

\DeclareMathOperator*{\argmin}{arg\,min}        % Argmin
\DeclareMathOperator*{\argmax}{arg\,max}        % Argmax
\newcommand{\iter}[2][\bocount]{{#2}^{(#1)}}    % Iteration specific instance of variable/function/anything

\newcommand{\prob}[0]{p}                        % Probability p
\newcommand{\cost}[0]{c}                        % Cost
\newcommand{\x}[0]{\mathbf{x}}                              % Input vector x
\newcommand{\y}[0]{y}                                       % Output y
\newcommand{\surro}[0]{\hat{\cost}}             % Surrogate model

\usepackage{comment}

\usepackage{verbatim}
\usepackage{tikz}
\usetikzlibrary{positioning}

\usepackage[round]{natbib}

\usepackage{hyperref}

\usepackage{aistats2022}

\title{Searching in the Forest for Local Bayesian Optimization}

\usepackage[subpreambles=true]{standalone}
\usepackage{xr}
\usepackage{natbib}
\usepackage{graphicx}
\usepackage{amsmath}
\usepackage{amsfonts}
\usepackage{amssymb}
\usepackage{mathtools}
\usepackage{bm}
\usepackage{wrapfig}

\newcommand{\observe}{\mathbf{X}}
\newcommand{\candidate}{\bm{x}}
\newcommand{\pcs}{\mathcal{X}}
\newcommand{\covvect}{\mathbf{k}}
\newcommand{\covmat}{\mathbf{K}}
\newcommand{\idenmat}{\mathbf{I}}
\renewcommand{\cost}{f}

\newcommand{\inside}{\mathbf{i}}
\newcommand{\outside}{\mathbf{o}}

\newcommand{\ndims}{\mathit{d}}

\newcommand{\xout}{\mathbf{X_\outside}}
\newcommand{\fout}{\mathbf{f_\outside}}
\newcommand{\yout}{\mathbf{y_\outside}}
\newcommand{\xin}{\mathbf{X_\inside}}
\newcommand{\fin}{\mathbf{f_\inside}}

\newcommand{\ftest}{\mathbf{f_\star}}
\newcommand{\inducing}{\mathbf{u}}

\begin{document}

\twocolumn[

\aistatstitle{Searching in the Forest for Local Bayesian Optimization}

\aistatsauthor{ Difan Deng\And Marius Lindauer}

\aistatsaddress{ Leibniz University Hannover \And  Leibniz University Hannover} ]

%\icmltitle{Searching in the Forest for Local Bayesian Optimization}
% Bi-Level Bayesian Optimization
% Locally Focused Bayesian Optimization 
% Local Bayesian Optimization with GP-Trees
% Tree-Structured Local Bayesian Optimization
% Searching in the Forest for Local Bayesian Optimization
%
%]

\begin{abstract}
Because of its sample efficiency, Bayesian optimization (BO) has become a popular approach dealing with expensive black-box optimization problems, such as hyperparameter optimization (HPO). Recent empirical experiments showed that the loss landscapes of HPO problems tend to be more benign than previously assumed, i.e. in the best case uni-modal and convex, such that a BO framework could be more efficient if it can focus on those promising local regions. In this paper, we propose BOinG, a two-stage approach that is tailored toward mid-sized configuration spaces, as one encounters in many HPO problems. 
In the first stage, we build a scalable global surrogate model with a random forest to describe the overall landscape structure. Further, we choose a promising subregion via a bottom-up approach on the upper-level tree structure. In the second stage, a local model in this subregion is utilized to suggest the point to be evaluated next. %To capture the local distribution well while preserving the information of the global structure, we further propose a partial sparse Gaussian Process model. 
Empirical experiments show that BOinG is able to exploit the structure of typical HPO problems and performs particularly well on mid-sized problems from synthetic functions and HPO.
\end{abstract}

\section{Introduction}
Hyperparameter optimization (HPO) is considered to be a tedious, error-prone and expensive problem, but nevertheless crucial for achieving peak performance of a machine learning algorithm on a given dataset~\citep{bergstra-jmlr12a,feurer-bookchapter19a}. Here we assume that a user-defined cost function $\cost$ is optimized over a space $\pcs$ of possible hyperparameter configurations $\candidate \in \pcs$:
$\candidate^* \in \argmin_{\candidate \in \pcs} \cost(\candidate)$.
Typical cost functions $\cost$ include, for example, accuracy for classification or RMSE for regression either over a hold-out validation set or a $k$-fold cross validation on the training set~\citep{thornton-kdd13a}.
Since HPO is often treated as an expensive black-box problem, where only a few function evaluations can be afforded and no gradient is available, Bayesian Optimization~(BO; \citep{%mockus-jgo94,
jones-jgo98a}) is a common and efficient approach for obtaining well-performing hyperparameter configurations~\citep{snoek-nips12a,hutter-lion12a,bergstra-icml13a, eriksson-nips19a}.

%NeurIPS

% \begin{wrapfigure}{r}{0.5\textwidth}
%     \centering
%     \vspace{-1em}
%     \begin{subfigure}{0.25\textwidth}
%     \centering
%     \includegraphics[width=1.0\textwidth]{images/bayesmark/ada_iris_acc.png}
%     \end{subfigure}\hfill
%     \begin{subfigure}{0.25\textwidth}
%     \centering
%     \includegraphics[width=1.0\textwidth]{images/bayesmark/linear_wine_nll.png}
%     \end{subfigure}
%     \caption{Exemplary hyperparameter loss landscape generated on BayesMark~\cite{bayesmark}.}
%     \label{fig:loss_land_scape}
%     \vspace{-1em}
% \end{wrapfigure}

%AAAI

\begin{figure}
    \centering
    \begin{subfigure}{0.235\textwidth}
    \centering
    \includegraphics[width=1.0\textwidth]{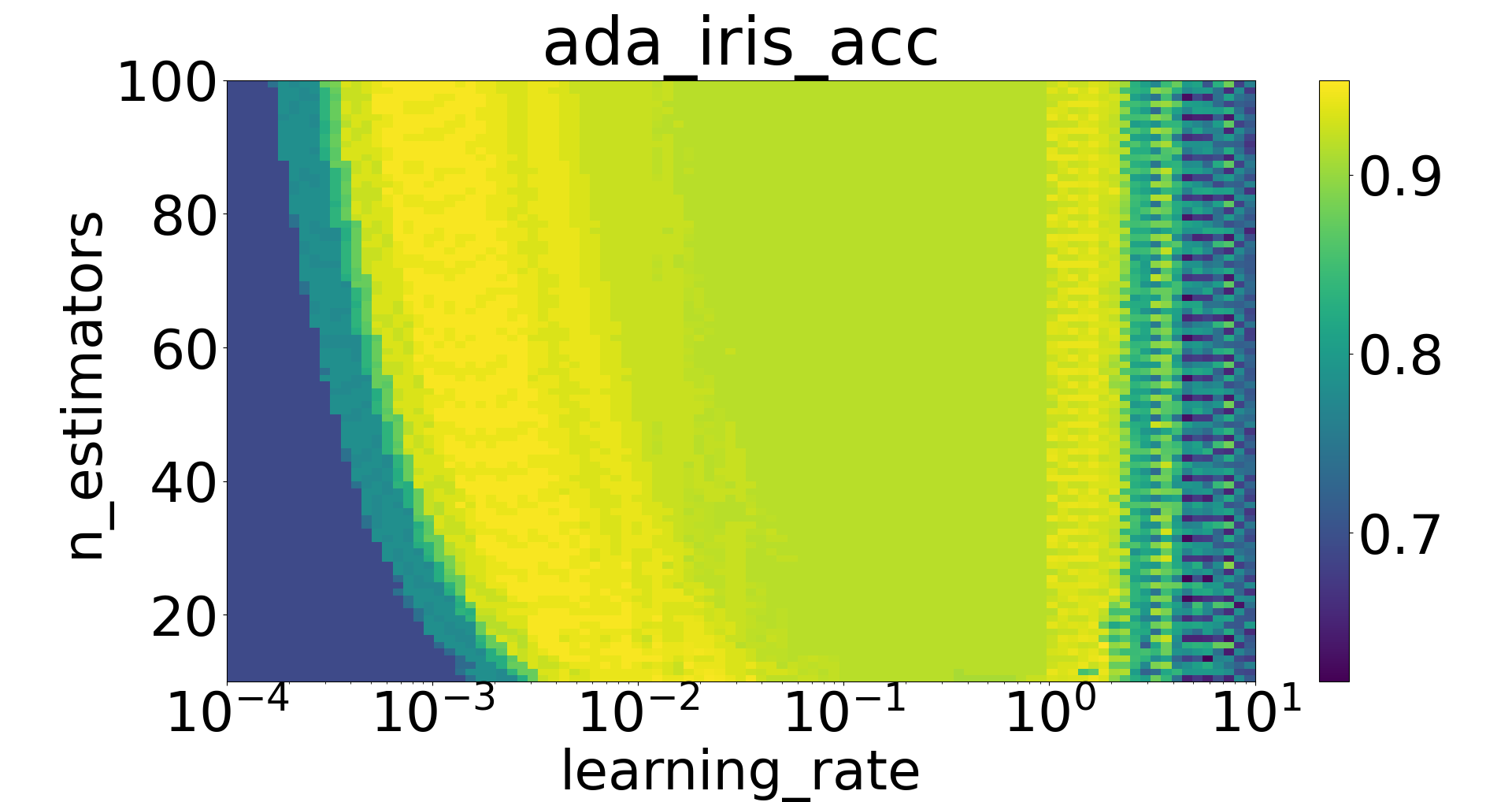}
    \end{subfigure}
    \begin{subfigure}{0.235\textwidth}
    \centering
    \includegraphics[width=1.0\textwidth]{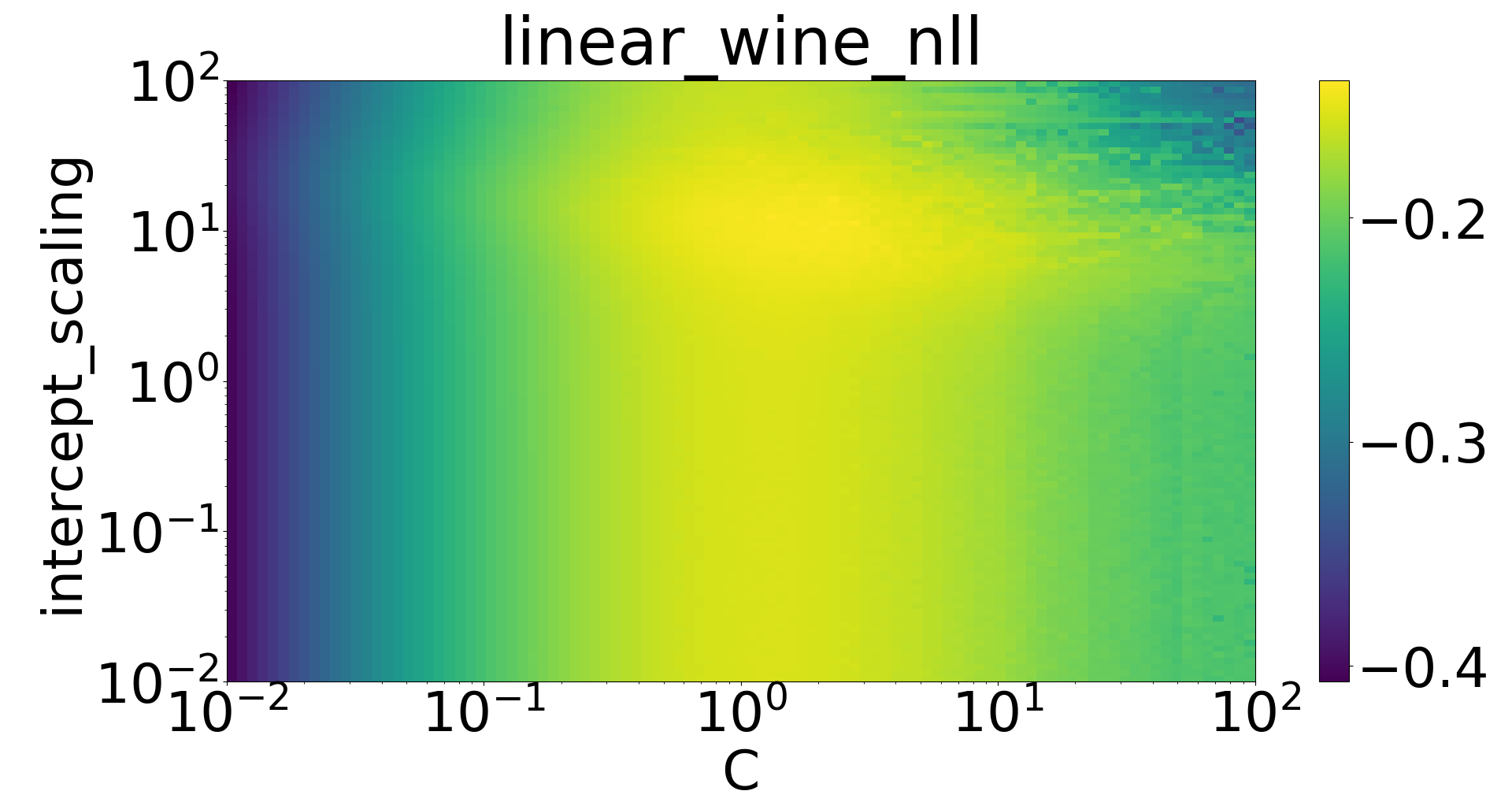}
    \end{subfigure}
    \caption{Exemplary hyperparameter loss landscape generated on BayesMark~\citep{bayesmark}. % (Left) the accuracy values of ada-boosting models with two hyperparameters: number of estimators and learning rate. (Right) the negative log likelihood values of linear classifiers with two hyperparameters: C (regularization strength) and intercept\_scaling
    }
    \label{fig:loss_land_scape}
\end{figure}

%However, previous work on HPO showed two important insights: First, the loss landscape of a hyperparameter optimization problem is more benign than what one would expect~\cite{klein-ejs17,Pushak-ppsn18a,loshchilov-iclr19a,pimenta-evocc20a}. In most cases, the loss landscape in well-performing regions is quite flat, sometimes convex and the best performing region is fairly well defined, see for example Figure~\ref{fig:loss_land_scape}. This in turn implies that we can focus on promising regions and can ignore potentially poor regions early on to increase the sample efficiency of BO. Second, different surrogate models, used to trade-off exploration and exploitation, perform well depending on the task at hand~\cite{eggensperger-bayesopt13}. 

However, previous work on HPO showed two important insights: First, the loss landscape of a hyperparameter optimization problem is more benign than what one would expect~\citep{klein-ejs17,Pushak-ppsn18a,pimenta-evocc20a}. In most cases, the loss landscape in well-performing regions is quite flat and the best performing region is fairly well defined, see for example Figure~\ref{fig:loss_land_scape}. Given a limited HPO budget, we prefer to focus more on identified best performing regions and to avoid unnecessary exploration in the later stages of the optimization. Second, different surrogate models, used to trade-off exploration and exploitation, perform well depending on the task at hand~\citep{eggensperger-bayesopt13}.

%Standard BO approaches are guided by a surrogate model, e.g. a Gaussian Process (GP)~\cite{snoek-nips12a}, a random forest (RF)~\cite{hutter-lion12a}, or Bayesian Neural Network (BNN)~\cite{snoek-icml15a}, estimate the possible distribution of the target function holistically. 
%However, given a limited number of function evaluations for the optimization, these approaches might exhibit a poor exploration-exploitation trade-off by overly exploring regions with high uncertainties regarding the true function value. 

Inspired by recent advances in exploiting local structures in BO~\citep{eriksson-nips19a,wang2020, wan2021think}, we address these insights by proposing a two-stage BO algorithm. Our approach treats the optimization problem on different scales: in the first stage, i.e. the global level, we execute a full BO iteration with all observed points that are evaluated previously. Then we extract a subregion based on the trained surrogate model and the suggested point given by the global optimizer. In the second stage, i.e. the local level, we perform another BO iteration with only the points inside the subregion and train a local surrogate model. Finally, we propose the sample to be evaluated next on the local model. Since we combine the best of two surrogates by extracting the subregion on the global level from a random forest (RF) and performing BO with a Gaussian Process (GP) inside this subregion, our robust HPO method is hence named as \emph{Bayesian Optimization inside a Grove (BOinG)}.

Our contributions are as follows: 
\begin{enumerate}
    \item We propose the two-stage Bayesian Optimization approach BOinG that allows to locate a promising subregion that should be explored more.
    \item By using a scalable global model (e.g., RF), BOinG reduces the computational burden on the local model (e.g., GP), which scales better to more iterations  than vanilla GP-BO.
    \item We propose to augment the local Gaussian Process with global data distribution such that the model captures the local loss landscape while preserving the influence of the global data distribution at minimal cost.
    \item We show the robustness and state-of-the-art performance of BOinG on several synthetic functions and HPO benchmarks for deep learning and reinforcement learning. %and provide insights into the exploration-exploitation tradeoffs on the global and local level.
    %\item Finally, our approach is surrogate model and acquisition function agnostic, and we can apply various meta-parameters to different stages to control different degrees of exploration-exploitation trade-offs in different phases of the optimization process. 
\end{enumerate}

\section{Related Work \& Background}

  %\subsection{Bayesian Optimization}
  Bayesian Optimization (BO) became a promising approach in solving expensive black-box functions~\citep{shahriari-ieee16a}. Recent progress on BO focuses on extending BO to large scale and high dimensional search space~\citep{kandasamy-icml15a,wang-aistats18a,eriksson-nips19a,wang2020}. 
  BO needs to employ a surrogate model to describe the possible data distribution of the target function. A GP model is the  commonly utilized surrogate model. The predicted mean and variance of a GP for given configuration $\candidate$ is given by
  \begin{align}
      \mu(\candidate) &= \covvect_{*}^{T}(\covmat+\sigma^2\idenmat)^{-1}\mathbf{y} \label{eq:gp_mean} \\
      \sigma^2(\candidate) &= \covmat(\candidate, \candidate) - \covvect_{*}^{T}(\covmat+\sigma^2\idenmat)^{-1}\covvect_* \label{eq:gp_var}
  \end{align}
  where $\covvect_*$ is the covariance vector between $\candidate$ and all previous observations, while $\covmat$ is the covariance matrix of all  previously evaluated points. 
  
  One drawback of GPs is its $\mathcal{O}(n^3)$-complexity for fitting the $n$ previous observations and $\mathcal{O}(n^2)$ for predicting means and variances at query points. Several approximate models such as sparse Gaussian processes~\citep{candela-jmlr6a} have been proposed to alleviate this issue by only using a set of additional points as inducing points to approximate the data distributions. Sparse GP reduces the complexity to $\mathcal{O}(m^2n)$ and $\mathcal{O}(mn)$ for fitting and predicting respectively, where $m$ is the number of the inducing points. Additionally, the introduction of variational inference~\citep{titsias-pmlr09a} brings great benefit to training a sparse GP, e.g., one can optimize a sparse GP w.r.t. each points individually~\citep{hensman-auai2013}, apply natural gradient~\citep{salimbeni-pmlr18a} for faster optimization, etc. However, in exchange, sparse GP leads to poor variance estimation that could mislead the optimizer~\citep{shahriari-ieee16a}. Despite its various complexity issues, GPs remain the most widely used surrogate model in BO frameworks.  
  
  An alternative surrogate model is a random forest~(RF \citep{breimann-mlj01a,hutter-lion11a}). An RF predicts the mean and variance values from the empirical mean and variance of each individual tree's predictions. Additionally, the tree structure allows it to easily deal with various data types as well as conditional hyperparameters, which is a common case in complex hyperparameter spaces. Still, RF's empirical mean and variance predictions might cause poor variance estimation when extrapolation is required~\citep{shahriari-ieee16a}.

  Partitioning the entire search space into several subregions with trees has been applied to reduce the computational complexity \citep{wang-aistats18a} or work with heteroscedasticity \citep{gramacy-icml04, Assael14} problems. The inborn hierarchy structure of trees makes them especially suitable for selecting the region that satisfies our requirements ~\citep{wang-aistats14}. We follow the same idea in BOinG to build a single local model on the most promising subregion that is obtained by the trees in an RF model. 
  %Nevertheless, these approaches train multiple local models for each partition to fully explore each subregion. Furthermore, new suggestions might never be selected from every region and it is not worth building a local model for these regions. BOinG follows a similar idea of space partitioning, but it only builds a single local model on the most promising subregions. 

  %TuRBO
  Similar to BOinG, TuRBO ~\citep{eriksson-nips19a} maintains a local model and expands or shrinks the subregion based on the evaluated result of the suggested point. TuRBO expands the subregion if it finds a point that is better than the current one and shrinks the subregion if the suggested point is worse than the current best evaluated point. TuRBO discards all previously evaluated observations if the subregion shrinks to be smaller than a threshold. Thus, TuRBO does not make full use of the previous evaluations and might sample repeatedly in the same region. Furthermore, since the local model and the subregion controller do not get access to the information about the entire search space, TuRBO might discard a promising subregion too early and hence cannot dig deeper within a limited budget. In contrast, BOinG maintains a global model and provides an estimation of the entire search space that guides the choice of the local subregion. Similarly, \cite{wang2020} improved TuRBO by utilizing Monte Carlo Tree Search to guide the choice and size of the subregion. However, their method only restarts TuRBO in a similar manner as the method proposed in TuRBO. Thus their method might still waste too many resources until TuRBO restarts.  %Given that the goal of BO is to optimize expensive functions, we might not afford the cost of too many function evaluations.
  
  While TuRBO only considers BO as a local optimizer, ~\cite{McLeodRO-icml18} relies BO on finding the most promising subregion and only exploiting inside that subregion. BOinG  considers the exploration and exploitation trade-off in both global and local levels. Hence it can quickly adjust to the local model while avoiding falling into local optimum too early.

\section{A General Approach of Two-Stage Bayesian Optimization}

% NeurIPS

% \begin{wrapfigure}[15]{r}{0.45\textwidth}
%     \centering
%     \vspace{-2em}
%     \begin{subfigure}[b]{0.2\textwidth}
%     \centering
%     \includegraphics[width=1.0\textwidth]{images/introduction/traj_GP.png}
%     \end{subfigure}\hfill
%     \begin{subfigure}[b]{0.2\textwidth}
%         \centering
%     \includegraphics[width=1.0\textwidth]{images/introduction/traj_BOinG.png}
%     \end{subfigure}\vfill
%     \begin{subfigure}[b]{0.4\textwidth}
%     \centering
%         \includegraphics[width=1.0\textwidth]{images/introduction/legend_traj.png}
%     \end{subfigure}
%     \caption{\textbf{Left}: The points suggested by a GP that is trained on the entire search space.  
%     \textbf{Right}: The subregions and points suggested by BOinG over time. Transparency visualizes subregions extracted in different phases. }
%     \label{fig:overview}
% \end{wrapfigure}

%AAAI

\begin{figure}
    \centering
    \begin{subfigure}[b]{0.2\textwidth}
    \centering
    \includegraphics[width=1.0\textwidth]{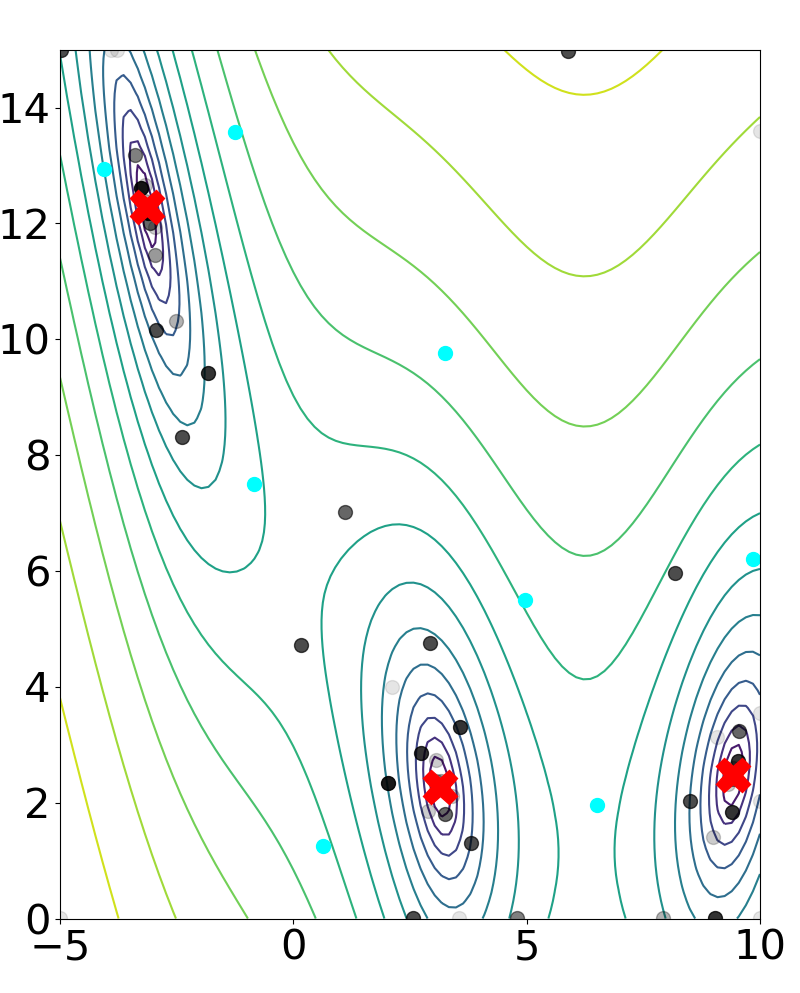}
    \end{subfigure}\hfill
    \begin{subfigure}[b]{0.2\textwidth}
        \centering
    \includegraphics[width=1.0\textwidth]{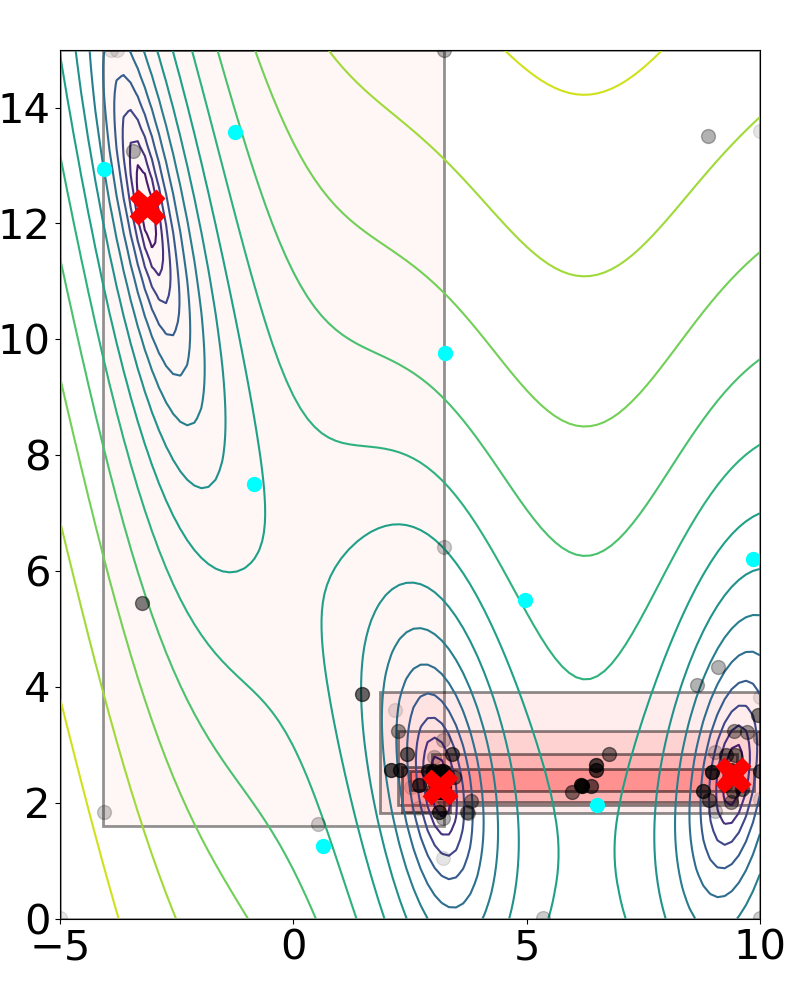}
    \end{subfigure}\vfill
    \begin{subfigure}[b]{0.4\textwidth}
    \centering
        \includegraphics[width=1.0\textwidth]{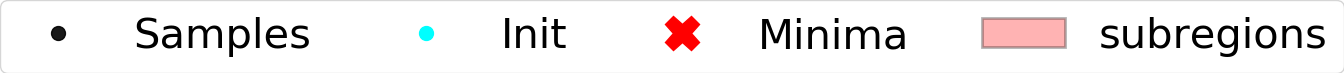}
    \end{subfigure}
    \caption{\textbf{Left}: The points suggested by a GP that is trained on the entire search space.  
    \textbf{Right}: The subregions and points suggested by BOinG over time. Transparency visualizes subregions extracted in different phases. }
    \label{fig:overview}
\end{figure}

Algorithm~\ref{alg:bibo} outlines the general idea of BOinG. In each iteration, we first train a global surrogate model~$\surro_{g}$ (Line~5) on all observations $\iter[t-1]{\observe}$ to estimate the possible regions that are worth being explored. Using an acquisition function $\alpha_g$ on $\surro_g$, we select a promising configuration (Line~6) that will guide the extraction of a subregion (Line~7; see Section~\ref{sec:first_level} for details). Then a new local model $\surro_l$ is fitted to the observations inside the selected subregion $\iter[t-1]{\observe_{in}}$ (see Section~\ref{sec:psgp} for details). Based on $\surro_l$, the maximum of a local acquisition function $\alpha_l$ (Line~9) decides the next configuration to be evaluated on the real cost function~$\cost$ (Line~10).

To be applicable to more complex problems with more reasonable evaluation budgets $T$,  the global model needs to scale well to many observations, e.g., an RF is suitable~\citep{hutter-lion11a}.
Since the number of points inside the subregion is rather small and nearly constant to some degree, we can afford the cost of accurate but expensive models (e.g., a GP) even if the number of the total evaluated points increases by a high degree.

\begin{algorithm}[tb]
 \caption{Two-stage Bayesian Optimization with BOinG}\label{alg:bibo}
\begin{algorithmic}[1]
 \STATE {\bfseries Input:}  a black-box function $\cost$; search space $\pcs$; predictive models $\surro_{g}$ and $\surro_{l}$ that fit global and local observation distributions respectively; acquisition functions $\alpha_g$ and $\alpha_l$; evaluation budget $T$
 
 \STATE {\bfseries Output:} global minimizer of $\cost$: $\candidate \in \argmin_{\candidate \in \pcs} \cost(\candidate)$
 
 \STATE{\bfseries Initialization: } Initialize data $\iter[0]{\observe}$ with initial observations
 
 \FOR{$t=1, 2, \cdots T$}
   \STATE fit global model $\iter{\surro_{g}}$ on $\iter[t-1]{\observe}$ 
   \STATE select a candidate point $\candidate_g$ with global acquisition function $\alpha_g$: \mbox{$\candidate_g \in \argmax_{\candidate \in \pcs} \alpha_g(\candidate; \iter[t-1]{\observe}, \iter{\surro_g})$}
   
   \STATE extract a subregion $\pcs_{sub} \subseteq \pcs$ based on global candidate $\candidate_g$ and model $\iter{\surro_g}$
   
   \STATE fit local model $\surro_l$ with the points inside the subregion $ \iter[t-1]{\observe_{\inside}} \in \pcs_{sub}$ together with the points outside the subregion $\iter[t-1]\observe_{\outside} = \iter[t-1]{\observe} \setminus \ \iter[t-1]{\observe}_{\inside}$ 
   
   \STATE determine final sampling point based on local acquisition function $\alpha_l$: \mbox{$\iter{\candidate}\in \argmax_{\candidate \in \pcs_{sub}} \alpha_l(\candidate; \iter[t-1]{\observe_{\inside}}, \iter[t-1]{\observe_{\outside}},\iter{\surro_l})$}
   
   \STATE query $\iter{y}:=\cost(\iter{\candidate})$ 
   \STATE update data: $\iter{\observe} \leftarrow \iter[t-1]{\observe} \cup \{\langle \iter{\candidate}, \cost(\iter{\candidate}) \rangle\}$
 \ENDFOR
 \STATE{\bfseries Return:} $\hat{\candidate} \in \argmin_{\candidate \in \pcs}\cost(\candidate)$
\end{algorithmic}
\end{algorithm}

Figure \ref{fig:overview} (right) illustrates how the subregion and the suggested points vary during the optimization. As the optimization progresses, the subregion extracted by the global model gradually shrinks toward a potential optimum. BOinG mostly focuses on the regions that look promising---here the lower part of the figure---and invests more evaluations inside these regions to focus the exploration of the landscape. Please note that here BOinG hardly explores the region that is shown to be sub-optimal (the upper and right part of the image). It only takes a few evaluations in the beginning to indicate the unfitness of that region. As a comparison, the GP being trained on the entire space invests many evaluations on sub-optimal regions; even worse, the GP suggests lots of points on the boundary; see Figure \ref{fig:overview} (left). BOinG hence focuses more on the regions that seem to be more promising. Thus, there is a good chance for our method to find a better solution within a limited budget.

%%%%%%%%%%%%%%%%%%%%%%%%%%%%%%%%%%%%%%%%%%%%%%%%%%%%%%%%%%%%%%%%%%%%%%%%%%
\subsection{Subregion Extraction with RF\label{sec:first_level}}
%%%%%%%%%%%%%%%%%%%%%%%%%%%%%%%%%%%%%%%%%%%%%%%%%%%%%%%%%%%%%%%%%%%%%%%%%%
We build the global model on all previous observations and then select the most promising region to be exploited in the next stage.
%We desire a cheap model at this stage to provide a rough estimation of the global loss landscape. 
We propose to leverage the scalable and hierarchical structure of a tree-based model, such as an RF, s.t. we can split the global search space into subregions that contain sufficient data points to fit a local model. 

First, we do a single BO iteration to determine a global candidate $\candidate_g$ based on a global model $\surro_g$ and acquisition function $\alpha_g$. Then starting from the root node of each tree and a subregion $\pcs_{sub}=\pcs$ , we iterate toward the leaf node that contains $\candidate_g$ and get a split of the current node. Each split will shrink $\pcs_{sub}$ and exclude several points from $\observe_{in}$. We repeatedly continue the split until it contains at least $n_{min}$ points. We stop further exploring one node if that split makes the number of the points in the subregion smaller than $n_{min}$. Finally we stop shrinking the subregion if we cannot step further in any of the trees without keeping the number of the points in the subregion greater than $n_{min}$. For the corresponding algorithm, we refer to the appendix \ref{sec:appendix_subregion_extar}.

%As described in Algorithm \ref{alg:bibo}, we do a single BO iteration to determine a global candidate $\candidate_g$ based on a global model $\surro_g$ and acquisition function $u_g$; the subregion then grows around this single point. We determine the size and position of the subregion as shown in Algorithm~\ref{alg:subregion}: For each individual tree of a RF, we find the leaf node in which $\candidate_g$ is located (Line~3). Then we traverse inversely toward the root node of each tree until the union of the subspaces represented by those nodes contain enough points (Line~4 onward and Line~12); latter defined by the hyperparameter $n_{min}$ of our approach. The final selected tree node with at least $n_{min}$ observations defines a subregion $\pcs_{sub}$ by the all splits above it. Eventually, we take the union of these spaces across trees as the extracted subregion.

\begin{figure}[h]
       \vspace{-1em}

 \centering
 \begin{subfigure}[b]{0.2\textwidth}
 \centering
 \scalebox{0.8}{
   \begin{tikzpicture}
   \usetikzlibrary{arrows}
    \usetikzlibrary{shapes}
     \tikzset{treenode/.style={draw, circle, font=\small}}
     \tikzset{line/.style={draw, thick}}
     \node [treenode, draw=red] (a0) {$a_0$};
     \node [treenode, below=0.75cm of a0, xshift=-1cm]  (a1) {$a_1$};
     \node [treenode, draw=red, below=0.75cm of a0, xshift=1cm]  (a2) {$a_2$};
     
     \node [treenode, draw=red, below=0.75cm of a2, xshift=-1cm] (a3) {$a_3$};
     \node [treenode, below=0.75cm of a2, xshift=1cm]  (a4) {$a_4$};
     
     \node [treenode, below=0.75cm of a3, xshift=-1cm] (a5) {$a_5$};
     \node [treenode, below=0.75cm of a3, xshift=1cm]  (a6) {$a_6$};
     
     \path [line] (a0.south) -- + (0,-0.4cm) -| (a1.north) node [midway, above] {$\candidate_1<0.3$};
     \path [line] (a0.south) -- +(0,-0.4cm) -|  (a2.north) node [midway, above] {$\candidate_1\geq0.3$};
     
     \path [line] (a2.south) -- + (0,-0.4cm) -| (a3.north) node [midway, above] {$\candidate_1<0.6$};;
     \path [line] (a2.south) -- +(0,-0.4cm) -|  (a4.north) node [midway, above] {$\candidate_1\geq0.6$};

     \path [line] (a3.south) -- + (0,-0.4cm) -| (a5.north) node [midway, above] {$\candidate_2<0.2$};;
     \path [line] (a3.south) -- +(0,-0.4cm) -|  (a6.north) node [midway, above] {$\candidate_2\geq0.2$};

   \end{tikzpicture}
   }
 \end{subfigure}\hfill
 \begin{subfigure}[b]{0.2\textwidth}
  \centering
       \scalebox{0.8}{
   \begin{tikzpicture}
   \usetikzlibrary{arrows}
    \usetikzlibrary{shapes}
     \tikzset{treenode/.style={draw, circle, font=\small}}
     \tikzset{line/.style={draw, thick}}
     \node [treenode, draw=red] (b0) {$b_0$};
     \node [treenode, draw=red, below=0.75cm of b0, xshift=-1cm]  (b1) {$b_1$};
     \node [treenode, below=0.75cm of b0, xshift=1cm]  (b2) {$b_2$};
     \node [treenode, draw=red, below=0.75cm of b1, xshift=-1cm] (b3) {$b_3$};
     \node [treenode, draw=red, below=0.75cm of b1, xshift=1cm]  (b4) {$b_4$};
     
     \node [treenode, draw=red, below=0.75cm of b4, xshift=-1cm] (b5) {$b_5$};
     \node [treenode, below=0.75cm of b4, xshift=1cm]  (b6) {$b_6$};
     
     \path [line] (b0.south) -- + (0,-0.4cm) -| (b1.north) node [midway, above] {$\candidate_2<0.4$};
     \path [line] (b0.south) -- +(0,-0.4cm) -|  (b2.north) node [midway, above] {$\candidate_2\geq0.4$};
     
     \path [line] (b1.south) -- + (0,-0.4cm) -| (b3.north) node [midway, above] {$\candidate_1<0.2$};
     \path [line] (b1.south) -- +(0,-0.4cm) -|  (b4.north) node [midway, above] {$\candidate_1\geq0.2$};

     \path [line] (b4.south) -- + (0,-0.4cm) -| (b5.north) node [midway, above] {$\candidate_1<0.7$};
     \path [line] (b4.south) -- +(0,-0.4cm) -|  (b6.north) node [midway, above] {$\candidate_1\geq0.7$};

   \end{tikzpicture}
     }
 \end{subfigure}
 \hfill
  \begin{subfigure}[b]{0.2\textwidth}
    \centering
  \begin{tikzpicture}[scale=3]
    \draw[->] (0,0) -- (1.05,0) node[below] {$x_1$};
    \draw[->] (0,0) -- (0, 1.05) node[left] {$x_2$};
  \draw (0,0) -- (1,0) -- (1,1) -- (0,1) -- (0,0);
  \draw (0.3, 0) -- (0.3, 1);
  \draw (0.6, 0) -- (0.6, 1);
  \draw (0.3, 0.2) -- (0.6, 0.2);
  \node[circle,inner sep=0.8pt,fill=black, label=right:{$A$}] at  (0.45,0.27) {};
  \node[circle,inner sep=0.8pt,fill=black, label=right:{$B$}] at  (0.36,0.1) {};
  \node[circle,inner sep=0.8pt,fill=red, label=above right:{$\candidate_g$}] at  (0.35,0.3) {};
  \node[circle,inner sep=0.8pt,fill=black, label=right:{$C$}] at  (0.57,0.12) {};
  \node[circle, inner sep=0.8pt,fill=black, label=right:{$D$} ] at (0.13, 0.72) {};
  \node[circle, inner sep=0.8pt,fill=black, label=right:{$F$} ] at (0.83, 0.17) {};
  \node[circle, inner sep=0.8pt,fill=black, label=right:{$E$} ] at (0.05, 0.25) {};
  \node[circle,inner sep=0.3pt,label=below:{$0.3$}] at (0.3,0) {};
  \node[circle,inner sep=0.3pt,label=below:{$0.6$}] at (0.6,0) {};
  
  \draw [thin , draw=black, fill=red, opacity=0.2]
       (0.3,0) -- (0.6,0) -- (0.6,0.4) -- (0.3, 0.4);
  \end{tikzpicture}
  \end{subfigure}\hfill
  \begin{subfigure}[b]{0.2\textwidth}
    \centering
  \begin{tikzpicture}[scale=3]
      \draw[->] (0,0) -- (1.05,0) node[below] {$x_1$};
    \draw[->] (0,0) -- (0, 1.05) node[left] {$x_2$};
  \draw (0,0) -- (1,0) -- (1,1) -- (0,1) -- (0,0);
  \draw (0, 0.4) -- (1, 0.4);
  \draw (0.2, 0) -- (0.2, 0.4);
  \draw (0.7, 0) -- (0.7, 0.4);
  \node[circle,inner sep=0.8pt,fill=black, label=right:{$A$}] at  (0.45,0.27) {};
  \node[circle,inner sep=0.8pt,fill=black, label=right:{$B$}] at  (0.36,0.1) {};
  \node[circle,inner sep=0.8pt,fill=black, label=right:{$C$}] at  (0.57,0.12) {};
  \node[circle,inner sep=0.8pt,fill=red, label=below:{$\candidate_g$}] at  (0.35,0.3) {};
  \node[circle, inner sep=0.8pt,fill=black, label=right:{$D$} ] at (0.13, 0.72) {};
  \node[circle, inner sep=0.8pt,fill=black, label=right:{$E$} ] at (0.05, 0.25) {};
  \node[circle, inner sep=0.8pt,fill=black, label=right:{$F$} ] at (0.83, 0.17) {};
  \node[circle,inner sep=0.3pt,label=below:{$0.2$}] at (0.2,0) {};
  \node[circle,inner sep=0.3pt,label=below:{$0.7$}] at (0.7,0) {};

  \draw [thin , draw=black, fill=red, opacity=0.2]
      (0.3,0) -- (0.6,0) -- (0.6,0.4) -- (0.3, 0.4);
  \end{tikzpicture}
  \end{subfigure}
  \caption{\label{fig:spaceagg}Subregion selection. The final extracted subregion is illustrated in the red-shaded region.}
 \end{figure}
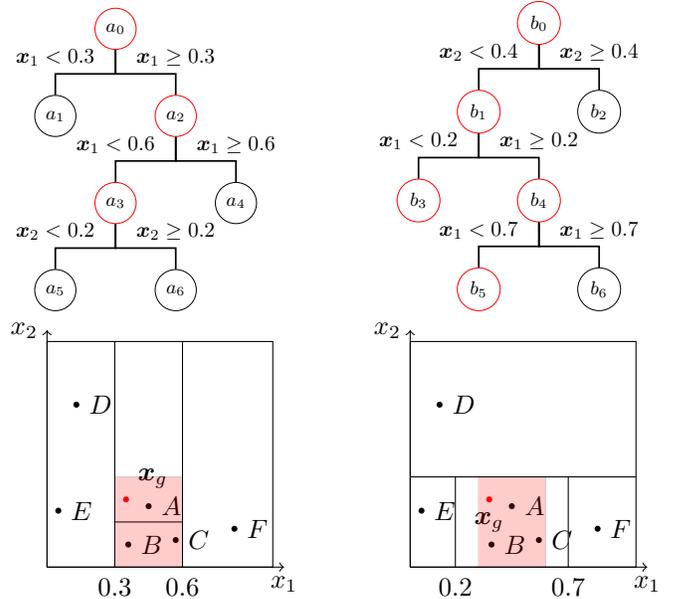

 We illustrate a minimal toy example in Figure \ref{fig:spaceagg}. We train an RF with two trees which split the space accordingly. Suppose that at least $n_{min}=3$ points are required to be included in the subregion and the subregion grows from $\candidate_g$ (marked in red). We note that, here $\candidate_g$ is not an actual sampled point, but it only indicates the position that the subregion should contain. We first step from $a_0$ to $a_2$ as it contains $\candidate_g$ and hence $D$ and $E$ are excluded from the subspace. Then we split the subspace with the split of $b_0$, where all the points are preserved. We further go deeper into the first tree and arrive at node $a_3$, in which case only 3 points lay in the subregion and we stop exploiting tree A. However, for tree B, as each split will not further shrink the subregion or exclude the existing points from the subregion, we arrive at leaf node $b_{5}$ and stop further shrinking the subregion.

%We first locate the leaf nodes $a_6$ and $b_5$ containing $\candidate$. Additionally, $a_6$ contains the observed point $A$ and $b_5$ contains $A$ and $C$. Hence, we will only have 2 observed points inside the initial subregion ($A$ and $C$). Next, we grow the current subregion of the first tree from $a_6$ to $a_3$, where another data point $B$ is included. Since the set of our subregions ($a_3$ and $b_5$) already contains enough observed points, we stop the subregion selection and simply take the union of spaces defined until $a_3$ and $b_5$. %To make the subregion a hypercube, we extract the subregion as the smallest hypercube that contains the union of all the subspaces. 

 %\begin{algorithm}[tb]
 %\caption{Subregion Selection with Random Forest}\label{alg:subregion}

%\begin{algorithmic}[1]
% \STATE {\bfseries Input:} Search Space $\pcs \in \mathbb{R}^n$; candidate selected by global Bayesian Optimization $\candidate_g$, random forest model $\surro$; minimal number of points stored in the subregion $n_{min}$; weak upper bound of points stored in the subregion $n_{max}$ 
% \STATE {\bfseries Output:} subregion $\pcs_{sub}$ extracted from $\pcs$
% \STATE {\bfseries Initialization: } $\observe_{sub} \leftarrow \emptyset $; a list of root nodes $S$ of RF~$\hat{c}$ in which $\candidate_g$ is located.
%\WHILE{$\observe_{sub}$ contains less points than $n_{min}$}
%  \FOR{each node $ s \in S $}
%    \STATE $s' \leftarrow parent(s)$
%    \STATE Let $\observe_{s'}$ be all observed points in $s'$
%    \IF{$|\observe_{sub} \cup \observe_{s'}| < n_{max}$}
%      \STATE $s \leftarrow s'$
%      \STATE $\observe_{sub} \leftarrow \ \observe_{sub} \cup  \observe_{s'}$
%    \ENDIF 
%    \IF{$n_{min} \leq |\observe_{sub}| \leq n_{max}$}
%      \STATE {\bfseries break;}
%    \ENDIF
%  \ENDFOR
%\ENDWHILE
 
% \STATE{\bfseries Return:} Let $\pcs_{sub}$ be the smallest hypercube that contains all the subspace represented by $S$
%\end{algorithmic}
%\end{algorithm}

One common issue arising from local BO is the size of the subregion. This region should not be too large, otherwise we would revert to a global optimization again. On the other hand, if the region is too small, the global optimum might be excluded from the subregion and we might only find a local minimum. 
Furthermore, the size of the subregion needs to be adapted to the number of the previous observations. A local optimizer should ideally contain all the previously evaluated points in the beginning and then gradually shrink to regions that are more likely to contain the global optimum. %This also follows the rule of the global Bayesian Optimization process: with great uncertainty in the very beginning the optimizer tends to explore more covering the whole search space. Then, as more and more points are evaluated, the optimizer tends to examine promising regions with a high probability of containing the global optimum.
BOinG achieves this by its user-defined hyperparameter $n_{min}$. If the $n_{min}$ points in the subregion are far from each other, i.e., BOinG still explores a lot and is not certain about a promising region, the selected subregion will be fairly large. In contrast if the $n_{min}$ points in the subregion are near to each other, i.e., BOinG has already found a promising region, the selected subregion will be fairly small.

\subsection{Fit Sub-Region with Gaussian Process}\label{sec:psgp}
The global model suggests a subregion that is worth being further explored. We could then utilize a GP on the points inside the subregion as an accurate model to describe the local data distribution. However, this could lead to proposed points on the boundaries of the subregion again, since it does not capture the overall trend of the data. So, a compromise between fitting a GP on all observations and on only the observations inside the subregion is required. To this end, we propose a local GP with an augmentation of the global trend that fits all points within the subregion and in addition efficiently approximates the global data distribution. 

We denote the points in the subregion and their function values as $\xin$ and $\fin$; here $\mathbf{\inside}$ refers to "inside subregions". Similarly we abbreviate the points outside the subregion as $\xout$ and $\fout$ respectively). Training a GP  with $\xin$ and $\fin$ implicitly assumes that the prior distribution of $\fin$ follows $ \prob(\fin) = \mathcal{N}(\mathbf{0}, \covmat_{\inside, \inside})$; on the contrary, when a GP model is built to fit all the previous evaluated points, $\fout$ provides a prior for $\fin$:
\begin{align}
    \prob(\fin|\fout) &= \mathcal{N}(\mu_{\fin | \fout}, \sigma_{\fin | \fout}^2) \label{eq:dis_posterior} \\
    \mu_{\fin | \fout} &= \covvect_{\outside, \inside}^{T}(\covmat_{\outside,\outside}+\sigma^2\idenmat)^{-1}\mathbf{\yout} \label{eq:mu_posterior} \\
     \sigma_{\fin | \fout}^2 &= \covmat_{\inside, \inside} - \covvect_{\outside, \inside}^{T}(\covmat_{\outside, \outside}+\sigma^2\idenmat)^{-1}\covvect_{\outside, \inside} \label{eq:var_posterior}
\end{align}

However, computing the posterior distribution with Equation \ref{eq:mu_posterior} and \ref{eq:var_posterior} for $\xin$ is expensive and inefficient if we aim for higher evaluation budgets. Luckily, we can make use of the fact that the number of $\xout$ is much greater than the $\xin$ and most of $\xout$ are far away from the subregion and thus have little influence on the subregion. Thus, we approximate Equation \ref{eq:dis_posterior} with a much cheaper proxy: a Sparse GP~\citep{candela-jmlr6a}.

  Sparse GPs introduce a set of latent variables $u$ called \textbf{inducing points} so that training points $\mathbf{x}$ and test points $\mathbf{x_*}$ are conditionally independent given $u$. We apply the same idea to approximate the posterior in Equation \ref{eq:mu_posterior} and \ref{eq:var_posterior} and compute the posterior as a prior for $\prob(\fin | \fout)$.  
Since the local GP model is augmented with a globally approximated distribution, our model is dubbed Local GP with Global Augmentation (LGPGA). 

\begin{figure}[t]
\centering
\begin{subfigure}[b]{0.2\textwidth}
\centering
\begin{tikzpicture}  
\node (u) {$\inducing$};
 \node (fin1) [below left = 0.5cm and 1.5cm of u]{$f_{\outside_1}$};
 \node (dots1) [right = 0.01cm of fin1]{$\cdots$};
  \node (finn) [right = 0.01cm of dots1]{$f_{\outside_{n_{2}}}$};
  \node (fout1) [right = 0.01cm of finn] {$f_{\inside_1}$};
  \node (dots2) [right = 0.01cm of fout1]{$\cdots$};
  \node (foutn) [right = 0.01cm of dots2]{$f_{1_{\inside_{1}}}$};
  \node (fstar) [right = 0.01cm of foutn] {$f_*$};
  \foreach \x in {fin1, finn, fout1, foutn, fstar}
  \draw[-] (u) -- (\x);
\end{tikzpicture} 
\end{subfigure}
\hfill
\begin{subfigure}[b]{0.2\textwidth}
\centering
\begin{tikzpicture}  
\node (u) {$\inducing$};
 \node (fout1) [below left = 0.5cm and 0.45cm of u]{$f_{\outside_1}$};
 \node (dots1) [right = 0.01cm of fout1]{$\cdots$};
  \node (foutn) [right = 0.01cm of dots1]{$f_{\outside_{n_{2}}}$};
  \node (fin) [right =0.5cm of u] {$\mathbf{\fin}$};
  \node (fstar) [right = 0.01cm of foutn] {$f_\star$};
  \foreach \x in {fout1, foutn, fstar}
    \foreach \y in {u}
      \draw[-] (\y) -- (\x);
    \draw[-] (fin) -- (fstar);
  \draw[-] (fin) -- (u);
\end{tikzpicture} 
\end{subfigure}

\caption {\label{fig:psgp} A comparison between FITC (left) and LGPGA~(right). In FITC, test latent function values $f_*$ can only obtain the information of $\mathbf{f}$ through the inducing variable $\inducing$. While in LGPGA, $f_\star$ can directly obtain information from the information from $\fin$.}
\end{figure}
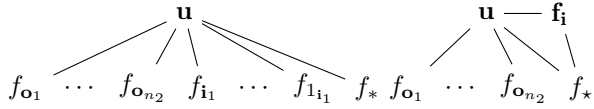

 We illustrate the difference between FITC~\citep{snelon-nips06a}) on all observations and our LGPGA with an approximation of the global trend and exact fit of the points inside the subregion in Figure~\ref{fig:psgp}.

%Taking FITC for example, we can rewrite its PSGP approximation joint distribution as:
The joint distribution of LGPGA is then computed in the following way, in this work, we apply FITC to approximate the global trend:
\begin{equation}\label{equ:psgp}
    \begin{bsmallmatrix}\mathbf{\fout}\\ \mathbf{\fin}\\ \ftest \end{bsmallmatrix} \sim
    \mathbf{\mathcal{N}}\left(\mathbf{0}, 
    \begin{bsmallmatrix}
    \mathbf{Q_{\outside,\outside} -\textrm{diag}[\mathbf{Q_{\outside,\outside} - \mathbf{K_{\outside,\outside}}}]} & \ \mathbf{Q_{\outside,\inside}} &\ \mathbf{Q_{\outside, \star}} \\
    \mathbf{Q_{\inside,\outside}} &\ \mathbf{K_{\inside,\inside}} &\ \mathbf{K_{\inside, \star}} \\
    \mathbf{Q_{\star, \outside}} &\ \mathbf{K_{\star, \inside}} &\ \mathbf{K_{\star, \star}}
    \end{bsmallmatrix}
    \right)
\end{equation}
%
%where $\fin$ denotes the set of the observations inside of the subregion while $\fout$ is the set of observations outside of the subregion. 
Where $\mathbf{Q_{a,b}}$ is defined as $\mathbf{K_{a,u}}\mathbf{K^{-1}_{u,u}}\mathbf{K_{u,b}}$ \citep{candela-jmlr6a}.

\begin{figure}
    % \centering
    % \begin{subfigure}[b]{0.3\textwidth}
    % \centering
    % \includegraphics[width=1.0\textwidth]{images/psgp/regression_fullGP.png}
    % \end{subfigure}
    % \hfill
    % \begin{subfigure}[b]{0.3\textwidth}
    % \centering
    % \includegraphics[width=1.0\textwidth]{images/psgp/regression_localGP.png}
    % \end{subfigure}
    % \hfill
    % \begin{subfigure}[b]{0.3\textwidth}
    \centering
    %\vspace{-1em}
    \begin{subfigure}[b]{0.3\textwidth}
        \includegraphics[width=1.0\textwidth]{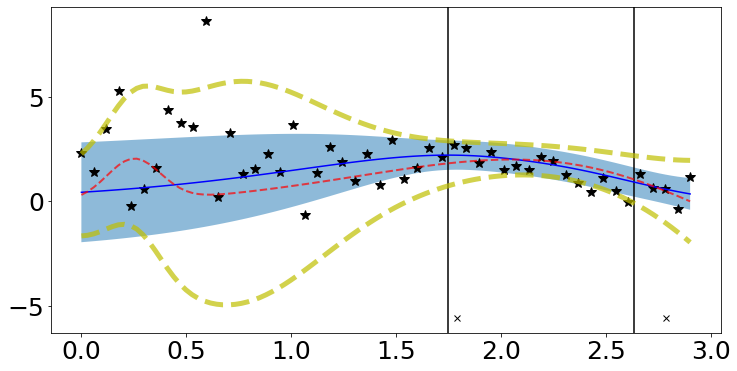}

    \end{subfigure}
    % \end{subfigure}
    % \vfill
    % \begin{subfigure}[b]{1.0\textwidth}
    % \centering
    \begin{subfigure}[b]{0.35\textwidth}

    \includegraphics[width=1.0\textwidth]{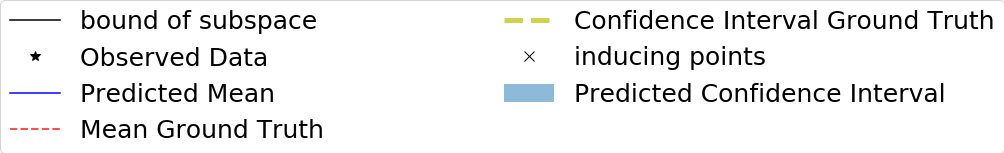}
        \end{subfigure}

    % \end{subfigure}
    \caption{
    %Regression result of different GP models. We want to approximate the data distribution inside the two red lines \textbf{Left}: A GP is utilized to fit the whole data distribution. \textbf{Middle}: A GP is utilized to fit only the data points inside the subregion. \textbf{Right}: 
    A LGPGA is utilized to fit the data points inside the subregion and still model the overall data trend also at the subregion's boundaries.}
    \label{fig:regression_gps}
\end{figure}

Two sets of LGPGA hyperparameters need to be optimized: the kernel hyperparameters and the position of inducing points. Comparing with inference time, we optimize these two sets of hyperparameters in an inverse order: we first train a GP model to fit $\fin$ to get the optimized kernel hyperparameters; thus LGPGA captures the local data distribution inside the subregion. We then train a sparse GP to fit $\fout$. The kernel hyperparameters of this sparse GP are the same as the ones in the first stage and freezed during the training process. In which case the sparse GP only optimizes the position of the inducing points. Any existing approximating strategy can be applied in the second training stage ~\citep{titsias-pmlr09a, hensman-auai2013, salimbeni-pmlr18a}.

% To illustrate how PSGP incorporates the information outside the subregion, we present a toy example in Figure~\ref{fig:regression_gps}. Following ~\cite{yuan2004}, we generate the data from a normal distribution with mean: $\mu (x) = 2(\exp (-30(x-1/4)^2)) + \sin (\pi x^2)$ and variance $\sigma^2 (x) = \exp (2\sin (2\pi x))$. This distribution has low noise level with larger $x$ values and high noise level with smaller $x$ values. Here we only use our model to fit the distribution of the right side.
% We randomly sample 50 points from $[0,1]$ and select the points indices from $35$ to $50$ as the points inside the subregion. 

As illustrated in Figure~\ref{fig:regression_gps}, LGPGA is able to model the data in the subregion accurately. %Nevertheless, it also captures the overall trend, in particular on the left side of the subregion. 
%fitting a GP on the entire data distribution will not exactly describe the noise on the right part of the data distribution. The GP fitting only the data points inside the subregion and our PSGP describe better the heteroscedastic noise inside the subregion. 
Additionally, on the right side of the subregion, LGPGA model takes the points outside the subregion into consideration and gives a more accurate prediction than a GP that only fits the data points inside the subregion. We note here, unlike previous work such as BOCK ~\citep{oh-icml18a} that explicitly assumes that the optimal points are likely to be located in the center of the subregion, LGPGA does not make any assumption about the data distribution inside the subregion. Indeed, if the data points indicate that the  boundary is still worth being explored, LGPGA would still try to sample the points near or on the boundary.

 \subsection{BOinG+: BOinG with Local TuRBO for Huge Budgets}\label{sec:BOinG+}

 Since RFs can have poor uncertainty estimates, SMAC~\citep{lindauer2021smac3} adds additional exploration by randomly sampling a new point with a certain probability and thus can provably converge to the true optimum in the limit~\citep{hutter-lion11a}. In BOinG, we can also combine SMAC's approach with the more explorative strategy of TuRBO: if we cannot make further progress with BOinG, we gradually increase the probability of switching to TuRBO instead of sticking to the subregion proposed by RF. To avoid unnecessary exploration, similar to ~\cite{wang2020}, we only do a TuRBO search inside a subregion, i.e. we randomly sample several points from $\pcs$ and extract the subregions $\pcs_{sub}$ around them with the method proposed in Section~\ref{sec:first_level}. Since we are interested in stronger exploration, we then select the subregions with the largest volume and thus with the smallest density of observed points. Then we start a new TuRBO run within this subregion whose initial points are the existing points in this subregion. Similarly, if we cannot make further progress with TuRBO, we increase the probability of switching back to BOinG. As we employ BOinG as an exploitation mechanism, we restart TuRBO earlier to allow more exploration: TuRBO restarts if the length of its subregion is smaller than $2^{-7}$, we restart TuRBO if its length is smaller than $2^{-4}$. For the details, we refer to the appendix \ref{sec:appendix_boing_detail}.
 
 All in all, BOinG (and BOinG+) is introduced to alleviate the weakness of an RF model: RF is a poor extrapolator and might fail to estimate the data distribution correctly in the regions with lower density of evaluated points~\citep{shahriari-ieee16a}. Thus we build another GP model in the vicinity of the point suggested by RF with lower evaluated budgets. Because of the poor extrapolation abilities, we extend BOinG by random search with a model-guided optimizer (TuRBO) for larger budgets where we can afford and potentially need better exploration.

\section{Experiments\label{sec:experiment}}
In this section, we first evaluate BOinG on several different functions and then perform an ablation study. 

\textbf{Experimental Setup}
For the number of included points of a subregion based on an RF, we use \mbox{$n_{min}:=5 \cdot \ndims$}, where $\ndims$ is the number of dimensions of the target function.
The local model is a GP with M\'atern $5/2$-kernel. We set the number of inducing points $n_{\inducing} := \min(10, 2 \cdot \ndims)$. Further implementation details can be found in the appendix \ref{sec:appendix_experiment_detail}.

As a baseline, we compare BOinG with RF and a GP that is trained on all the previous evaluations. Both BOinG and full GP are implemented with GPyTorch~v.1.2.1~\citep{Gardner18, balandat2020botorch}. RF and acquisition function optimizers of the above-mentioned BO models are implemented in the framework of SMAC3~v1.0.1~\citep{hutter-lion12a,lindauer2021smac3}\footnote{\url{https://github.com/automl/SMAC3}}, where a combination of random and local search for the optimization of the acquisition function is implemented. 

Since TuRBO\footnote{\url{https://github.com/uber-research/TuRBO}} and LA-MCTS\footnote{\url{https://github.com/facebookresearch/LaMCTS/tree/master/LA-MCTS}} 
are based on a similar idea of leveraging subregions, we compare BOinG against these, using their original implementations. 
LA-MCTS provides two ways of doing local optimization: BO and TuRBO; we present both approaches in our experiments, dubbed \texttt{LAMCTS-BO} and \texttt{LAMCTS-TuRBO}. TuRBO allows for batch BO; however, for a fair comparison, we only evaluate TuRBO-1 and set its batch size to 1.
%Additionally, as our method, TuRBO\cite{eriksson-nips19a} and LA-MCTS\cite{wang2020} derive from the idea of doing BO locally, we will also compare our methods against TuRBO and LA-MCTS. 

Furthermore, as baselines, we also compare against BO with GPs (\texttt{GP}) and RFs (\texttt{RF}) as surrogate models with EI~\citep{jones-jgo98a} as acquisition function,  and random search, using SMAC3. Finally, we consider TPE \citep{bergstra-nips11a} as another baseline. SMAC implemented LogEI to model heavy-tailed cost distributions~\citep{lindauer2021smac3}, our global optimizer follows the same implementation for HPO problems.

We ran all methods multiple times with different random seeds on 4 Intel Xeon E5 cores running openSUSE Leap 15.1. To ensure reproducibility, our code is publicly available at \url{https://github.com/dengdifan/SMAC3/tree/boing_tmp}%\footnote{In case this link should become invalid, we also provide the code in the supplementary material.} 

\subsection{Synthetic Function}
  
We assessed the algorithms on the following functions: Branin (2D), 
Ackley (10D) and Levy (10D).

% NeurIPS
% Branin (2D), Eggholder(2D), Ackley (2D, 5D and 10D) Hartman6 (6D) and Levy (2D, 5D and 10D). 
Branin, as probably the most studied function in GP-BO, has 3 global optima. State-of-the-art BO methods can very well pin-point one of the optima, which lie within fairly narrow regions --- unlike the landscape we expect for most HPO problems.
Ackley  
% Eggholder, and Hartman6 
has many local minima, allowing us to check against premature convergence. Levy has a large flat region near the optimum hence it tests the performance of an optimizer on a plateau. The input domain follows the suggestions in ~\cite{simulationlib}. All the optimization processes are repeated $30\times$. These synthetic functions usually do not fit the requirement of "heavy-tailed cost distributions". We simply apply EI acquisition function instead of logEI on these functions.

%NeurIPS 
  
%   \begin{figure}[!h]
%     \centering
%     \begin{subfigure}[b]{0.8\textwidth}
%     \centering
%     \includegraphics[width=1.0\textwidth]{images/res_synfunc/legend_synfunc.png}
%     \end{subfigure}
%     \vfill
%     % \begin{subfigure}[b]{0.24\textwidth}
%     % \centering
%     % \includegraphics[width=1.0\textwidth]{images/res_synfunc/branin2.png}
%     % \end{subfigure}
%     %\hfill
%     \begin{subfigure}[b]{0.32\textwidth}
%     \centering
%     \includegraphics[width=1.0\textwidth]{images/res_synfunc/branin2LargeFont.png}
%     %\includegraphics[width=1.0\textwidth]{images/res_synfunc/eggholder2.png}
%     \end{subfigure}
%     \hfill
%     \begin{subfigure}[b]{0.32\textwidth}
%     \centering
%     \includegraphics[width=1.0\textwidth]{images/res_synfunc/levy10LargeFont.png}
%     \end{subfigure}
%     \hfill
%     \begin{subfigure}[b]{0.32\textwidth}
%     \centering
%     \includegraphics[width=1.0\textwidth]{images/res_synfunc/ackley10LargeFont.png}
%     \end{subfigure}

%     \caption{Optimization on different synthetic functions, the solid lines are the mean of the best observed values with semi-transparent areas to indicate standard error.}
%     \label{fig:synethicFunc}
% \end{figure}

 \begin{figure*}[!h]
    \centering
      \vspace{-1em}

    \begin{subfigure}[b]{0.9\textwidth}
    \centering
    \includegraphics[width=0.9\textwidth]{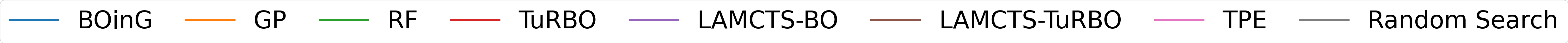}
    \end{subfigure}
    \vfill
    \begin{subfigure}[b]{0.3\textwidth}
    \centering
    \includegraphics[width=1.0\textwidth]{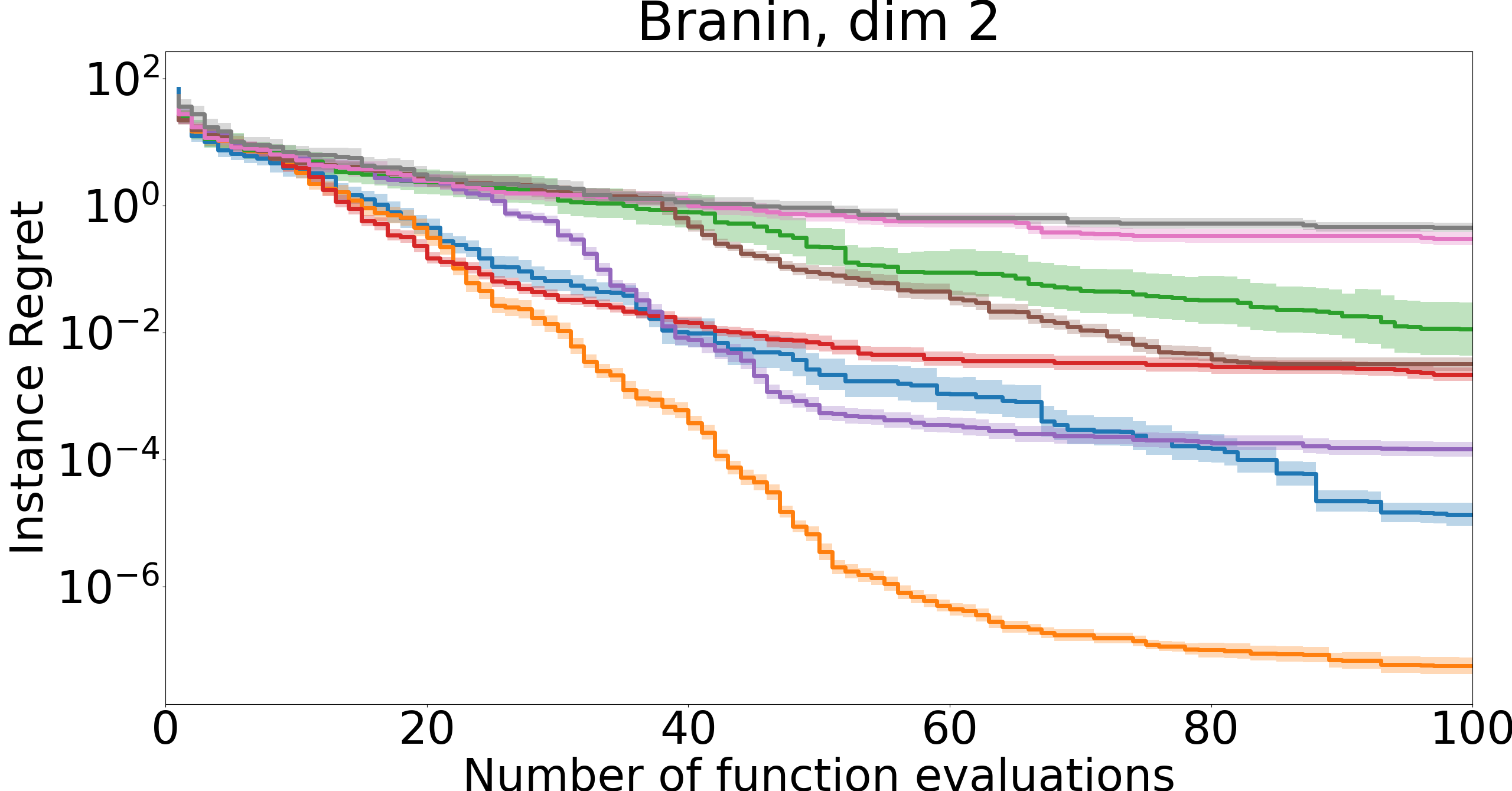}
    \end{subfigure}
    \hfill
    \begin{subfigure}[b]{0.3\textwidth}
    \centering
    \includegraphics[width=1.0\textwidth]{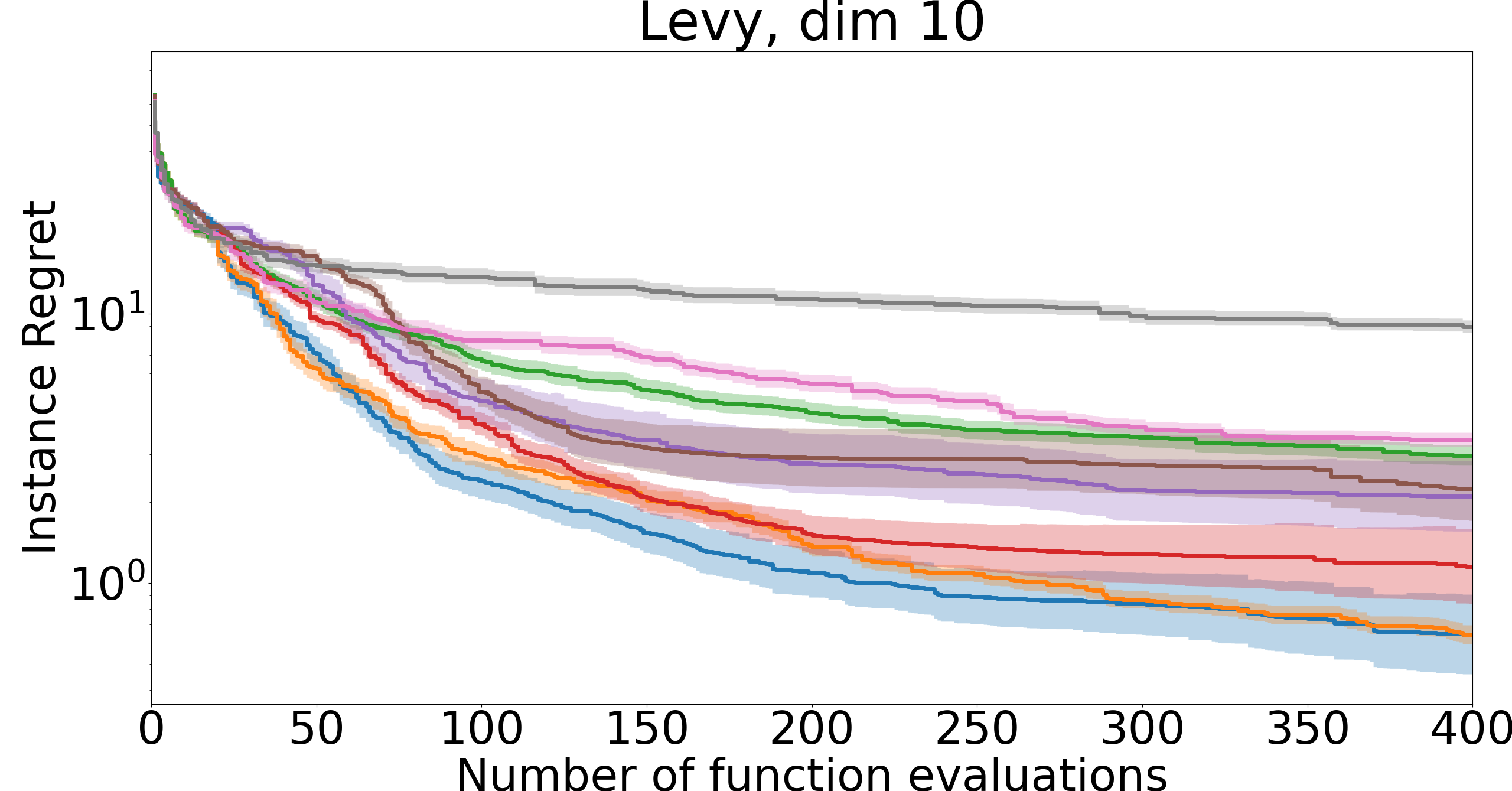}
    \end{subfigure}
    \hfill
    \begin{subfigure}[b]{0.3\textwidth}
    \centering
    \includegraphics[width=1.0\textwidth]{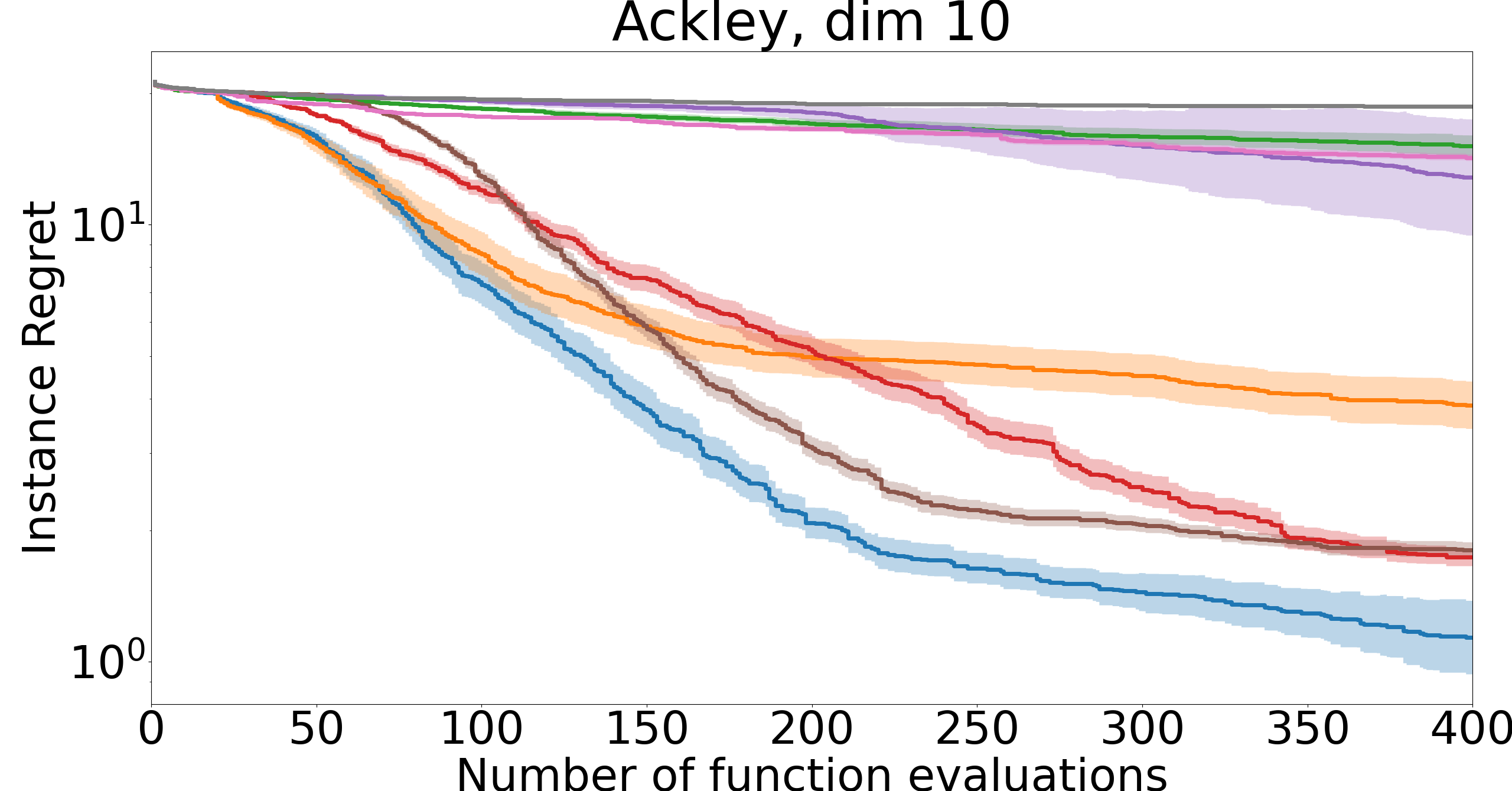}
    \end{subfigure}

    \caption{Optimization on different synthetic functions, the solid lines are the mean of the best observed values with semi-transparent areas to indicate standard error.}
    \label{fig:synethicFunc}
\end{figure*}
Figure~\ref{fig:synethicFunc} shows that--as one might expect--GP-BO is a very strong approach %on Branin and even 
for the Branin and Levy function, while on the Ackley function, the GP is not able to capture the high-dimensional, complex landscape well, and is outperformed by other approaches. Among all the compared optimizers, BOinG %shows a fairly good performance on Branin, but (NeurIPS)
is the only one showing a robust performance on higher-dimensional problems, being always at least on par with the best optimizer. %It combines the best of several worlds: different surrogate models and global and local optimization.

  \begin{figure*}[h]
  \centering
  \vspace{-1em}
  \begin{subfigure}[b]{0.8\textwidth}
    \centering
    \includegraphics[width=0.9\textwidth]{images/SynFunc/legend.png}
\end{subfigure}
    \vfill
    \centering
    %\begin{subfigure}[b]{0.32\textwidth}
    \centering
    \includegraphics[width=0.34\textwidth]{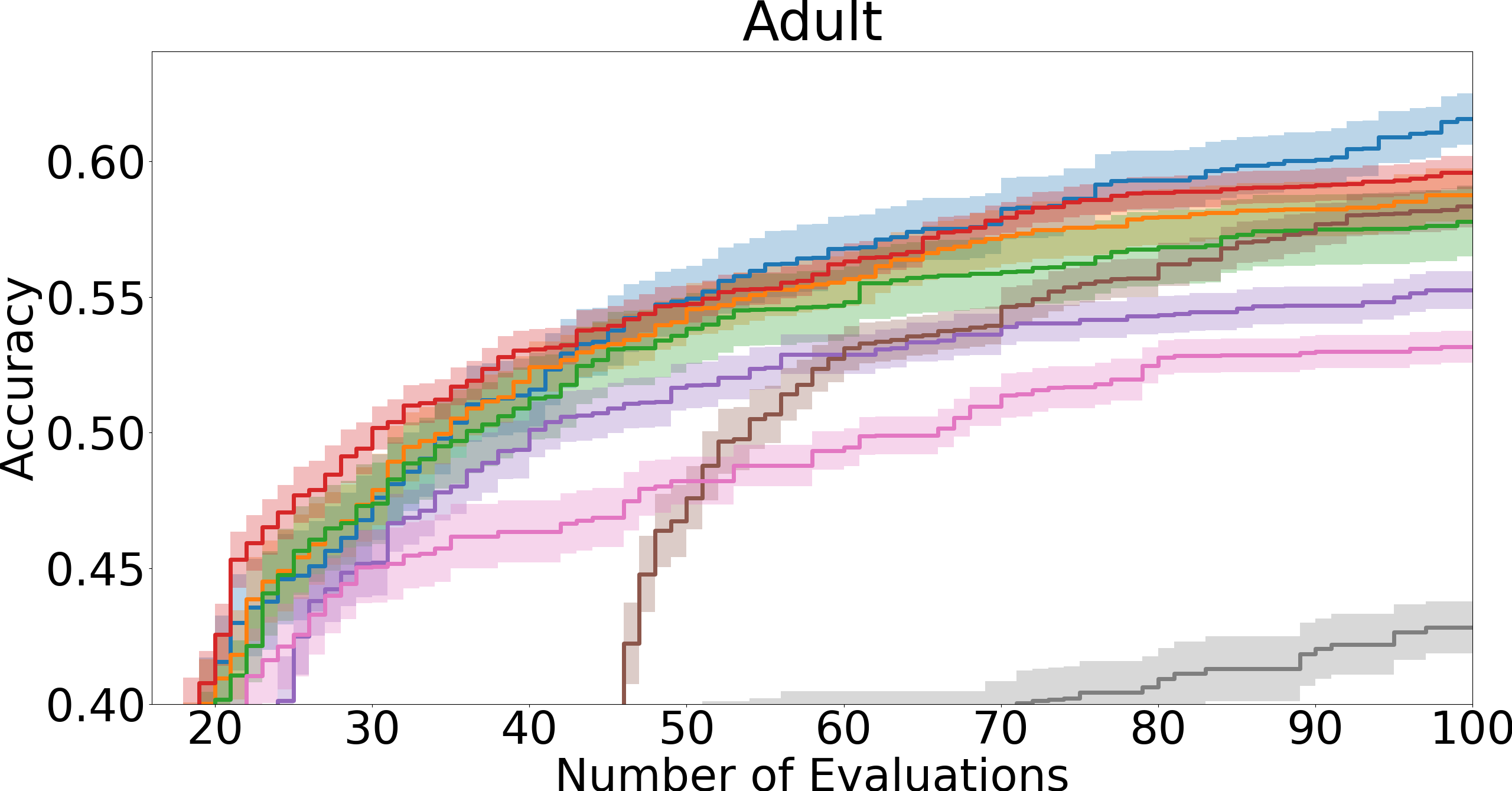}
    %\end{subfigure}
    %\begin{subfigure}[b]{0.32\textwidth}
    %\centering
    %\includegraphics[width=1.0\textwidth]{images/res_hpobench/higgs_EI.png}
    %\end{subfigure}
    %\begin{subfigure}[b]{0.32\textwidth}
    %\centering  \begin{subfigure}[b]{0.8\textwidth}
    \includegraphics[width=0.34\textwidth]{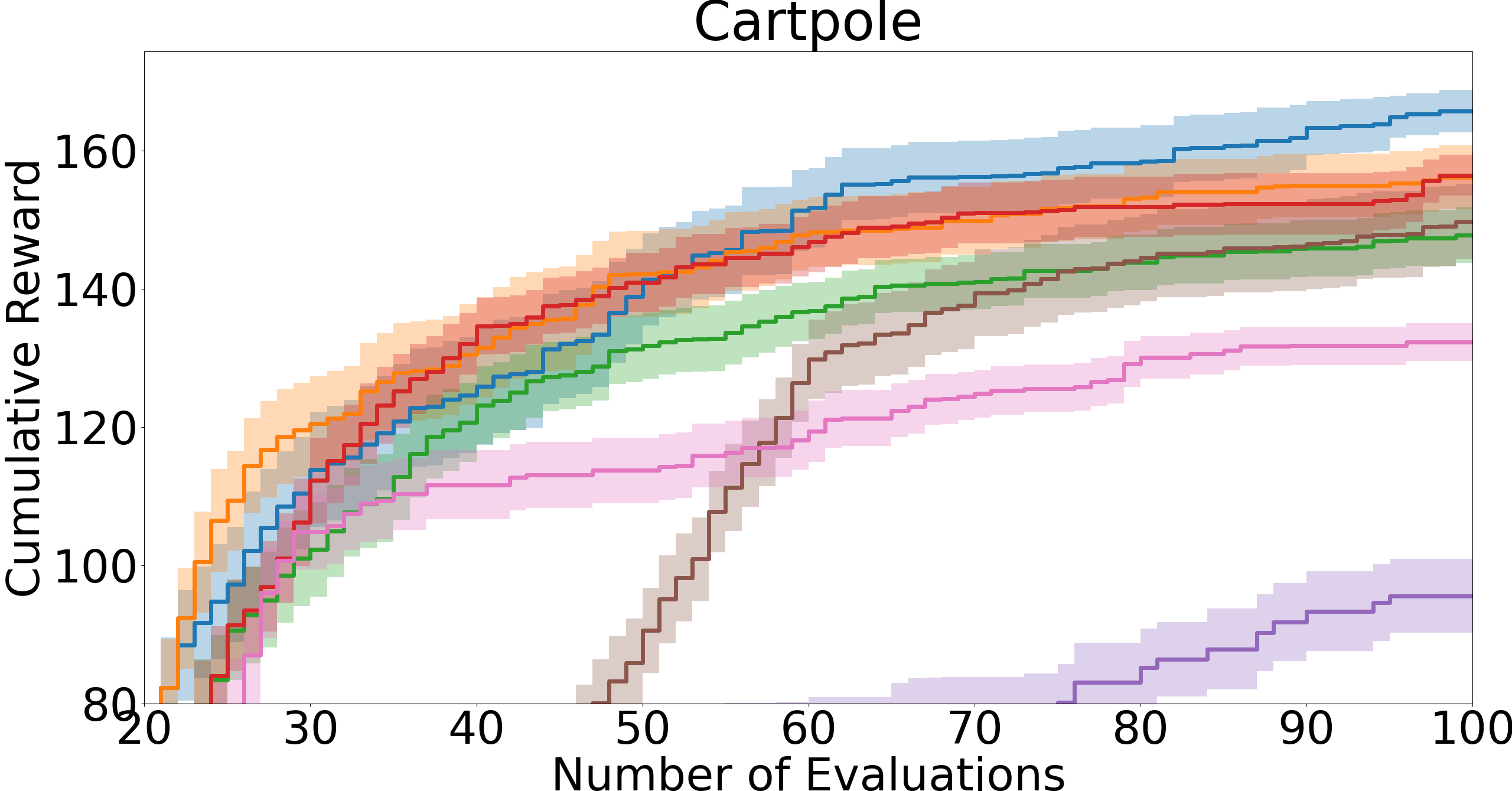}
    %\end{subfigure}
    %\begin{subfigure}[b]{0.6\textwidth}
    %\centering
    %\includegraphics[width=1.0\textwidth]{images/legend_synfunc.png}
    %\end{subfigure}
    %\begin{subfigure}[b]{0.32\textwidth}
    %\centering
    \vfill
    \begin{subfigure}[b]{0.8\textwidth}

    \centering
    \includegraphics[width=0.9\textwidth]{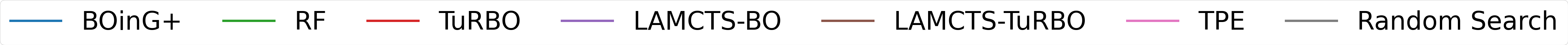}
    \end{subfigure}
    \includegraphics[width=0.34\textwidth]{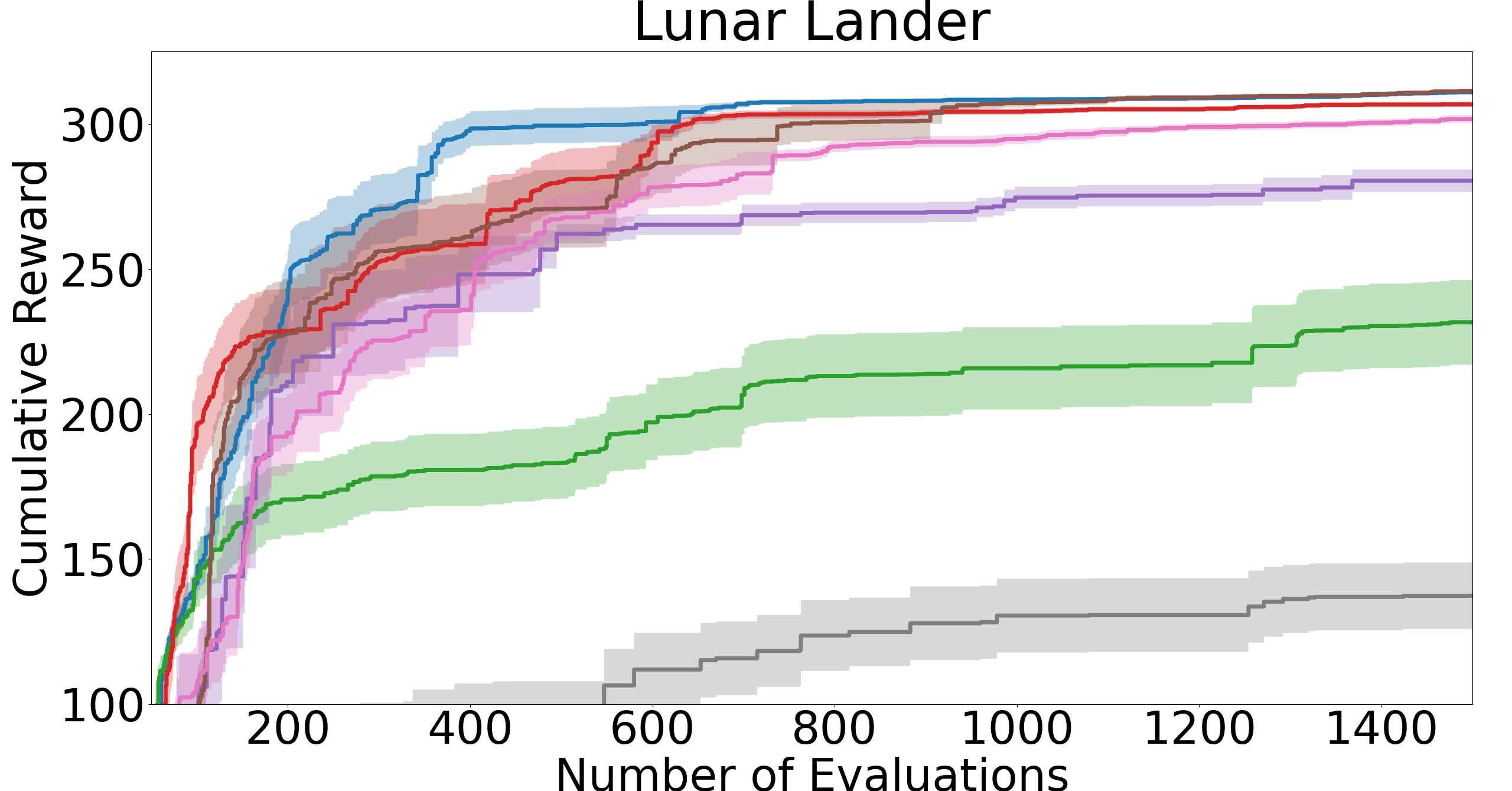}
    %\end{subfigure}
    \includegraphics[width=0.34\textwidth]{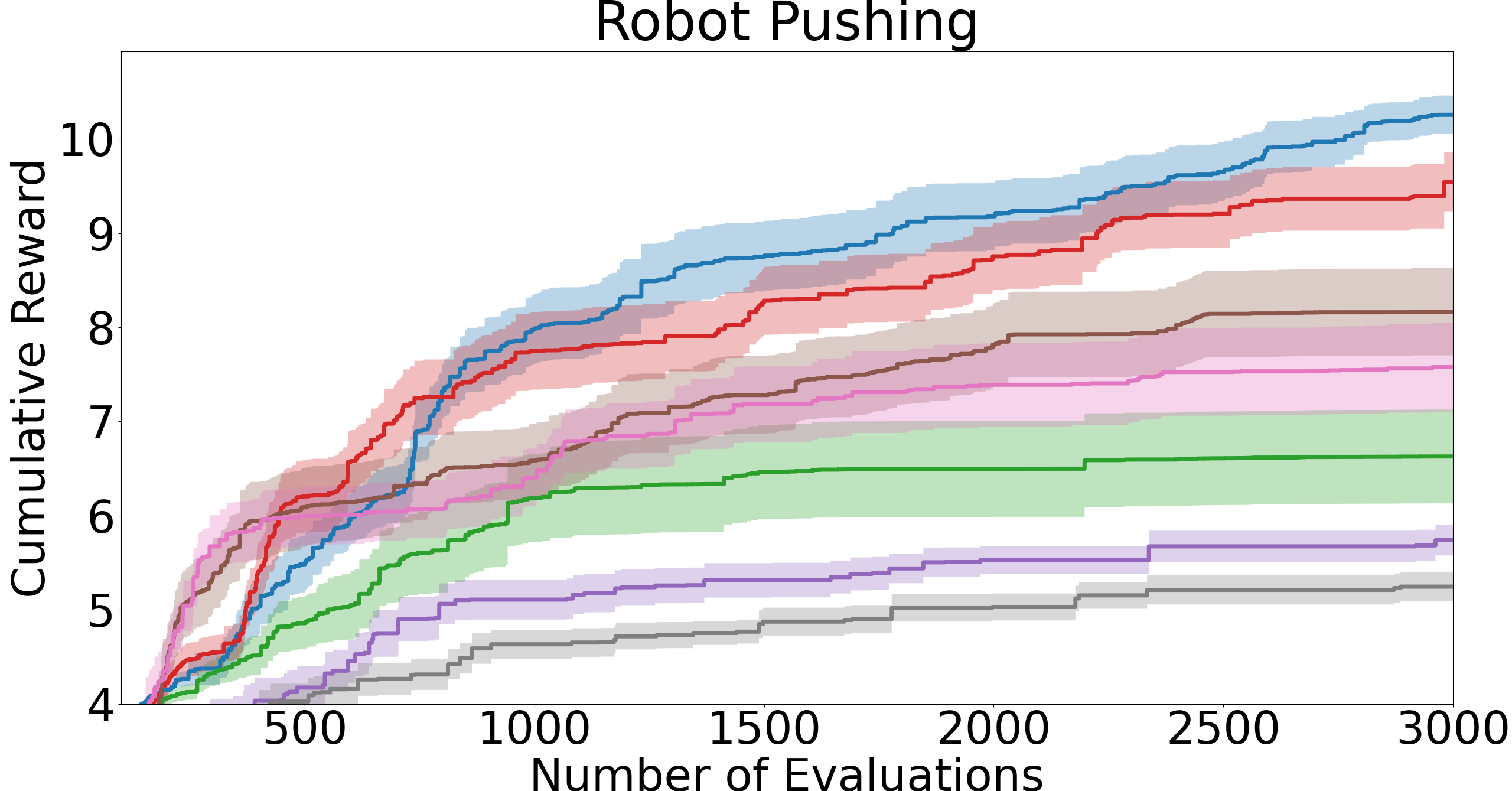}
    \caption{Results on ParamNet (Top Left), CartPole (Top Right), Lunar (Bottom Left) and Robot (Bottom Right). We show the mean performance and the standard error. Larger is better.}
    \label{fig:hpobench}
  \end{figure*}

  \subsection{Hyperparameter Optimization of ML Algorithms}
  
  %\subsubsection{HPOBench}
  
%   \begin{table*}[t]

%   \begin{tabular}{c c c c c c}
%       & $DE(Paper)$ & $BO_{GP}(Paper)$ & $BO_{RF} (Paper)$ & $BO_{RF}(local)$ & BOinG \\
%     SVM & 0.06020 & 0.05121 & 0.05206 & 0.05273 &  0.05255\\
%     LogReg & 0.00545 & 0.00172 & 0.00172 & 0.00213 & 0.00185 \\
%     RandomForest &  0.05701 & 0.01865 & 0.00107 &0.00128 & 0.01021 \\
%     XGBoost & 0.00579 & 0.00616 & 0.00601 & 0.00557 &   0.00438\\
%     MLP  & 0.01083 & 0.00902 & 0.00953 & 0.00985 &0.00844      
    
%   \end{tabular}
%  \end{table*}
  
  Although synthetic functions are a nice sanity check, we designed BOinG with HPO in mind and thus evaluated it on several HPO tasks.
  
  %\subsubsection{Hyperparameter Optimization On a Plain Configuration Space}
  
%NeurIPS
  
%   \begin{wrapfigure}[19]{r}{0.7\textwidth}
%   \centering
%   \vspace{-1em}
%   \begin{subfigure}[b]{0.7\textwidth}
%     \centering
%     \includegraphics[width=1.0\textwidth]{images/res_synfunc/legend_synfunc.png}
%     \end{subfigure}
%     \vfill
%     \centering
%     %\begin{subfigure}[b]{0.32\textwidth}
%     \centering
%     \includegraphics[width=0.345\textwidth]{images/res_hpobench/adult.png}
%     %\end{subfigure}
%     %\begin{subfigure}[b]{0.32\textwidth}
%     %\centering
%     %\includegraphics[width=1.0\textwidth]{images/res_hpobench/higgs_EI.png}
%     %\end{subfigure}
%     %\begin{subfigure}[b]{0.32\textwidth}
%     %\centering
%     \includegraphics[width=0.345\textwidth]{images/res_hpobench/cartpole.png}
%     %\end{subfigure}
%     %\begin{subfigure}[b]{0.6\textwidth}
%     %\centering
%     %\includegraphics[width=1.0\textwidth]{images/legend_synfunc.png}
%     %\end{subfigure}
%     %\begin{subfigure}[b]{0.32\textwidth}
%     %\centering
%     \hfill
%     \includegraphics[width=0.345\textwidth]{images/res_hpobench/lunar.png}
%     %\end{subfigure}
%     \includegraphics[width=0.345\textwidth]{images/res_hpobench/robot.png}
%     \caption{Results on ParamNet (Top Left), CartPole (Top Right), Lunar (Bottom Left) and Robot (Bottom Right). We show the mean performance and the standard error. Larger is better. For the two bottom figures, BOinG\_EI denotes BOinG+}
%     \label{fig:hpobench}
%   \end{wrapfigure}

%AAAI

  We utilize BOinG to optimize the following hyperparameter benchmarks introduced by ~\cite{falkner-icml18a}, provided by  HPObench~\citep{eggensperger2021hpobench}\footnote{\url{https://github.com/automl/HPOBench/}}: (i) Tuning hyperparameter of a neural network surrogate model (ParamNet) for the Adult dataset (Adult) with eight hyperparameters; (ii) Tuning seven hyperparameters of proximal policy optimization (PPO)~\citep{schulman-arxiv17a} to learn the  cartpole task. The PPO agent is implemented by Tensorforce \citep{tensorforce} and the environment is implemented in OpenAI gym \citep{brockman-arxiv16-gym}. For the details, we refer to the appendix \ref{appendix_experiment_detail}.

     Figure \ref{fig:hpobench} (top) shows the result of different optimizers on both target algorithms. BOinG achieves substantially better final performance on both benchmarks. Since BOinG shares the same model as a full GP model in the beginning of the optimization phases ($5 \cdot \ndims$), it achieves nearly the same performance as a full GP in the first few iterations and is able to identify the most promising region after $\approx 60$ function evaluations.
  
  Additionally, as BOinG introduces a local model to reduce the complexity and BOinG+ solves the poor-extrapolator issue by RF, we could now study how BOinG scales to problems with larger budgets. We optimize 12 hyperparameters of a heuristic controller for a lunar  lander implemented in the OpenAI gym and 14 hyperparameters of a controller for the robot pushing problem~\citep{wang-aistats18a}. The results are shown in the lower part of Figure~\ref{fig:hpobench}. 
  %   Both tasks are applied in Erikson et al.~\cite{eriksson-nips19a} and we set the same search space as TuRBO did\footnote{\url{https://github.com/uber-research/TuRBO}}. Here we run 20 repetitions. We give lunar lander problem a budget of 1500 function evaluations and robot pushing a budget of 3000 function evaluations. 
  %Since a full GP is too expensive for the given number of function evaluations, we will not consider it here. 
  Although TuRBO and LA-MCTS-TuRBO are able to gain some advantage in the beginning, BOinG+ is able to catch up and even outperforms TuRBO and LA-MCTS-TuRBO after $200$ function evaluations on Lunar Lander. On the robot benchmark, BOinG+ and TuRBO clearly outperform all the other methods, with a small advantage for BOinG+. Overall, the results show the strong performance of BOinG even in the large-budget setting. 
  % However, due to the higher dimensional space, a GP might "over explore" the boundary ~\cite{oh-icml18a, eriksson-nips19a} in the beginning and hence TuRBO will have a better performance. However, as BOinG could constantly exploitat the  promosing region, it achieves a better final performance compared to TuRBO.
  
%\subsubsection{Hyperparameter Optimization on a Hierarchical Configuration Space}
%An extra benefit of building a GP upon RF is that BOinG can handle a hierarchical configuration space, which is commonly made use of in modern AutoML frameworks~\cite{thornton-kdd13a,feurer-nips2015a,zimmer-tpami21a}. Since lots of configurations are inactive under a conditional hierarchy, there is no need to build a subspace within the full configuration space. For choosing the subregion, we only consider the configuration space of the model to which the global candidate $\candidate_g$ belongs. Since the configurations of one model will not directly influence the performance of another model, we only build the PSGP with the configurations evaluated for that model. 

%Here we only compare BOinG against SMAC as the canonical BO approach with RFs. We use Auto-Sklearn's configuration space~\cite{feurer-nips2015a} with 152 hyperparameters and evaluate it on the \texttt{kr vs kp} dataset~\cite{rossi-aaai2015}.
  
\subsection{Ablation Study\label{sec:ablation_study}}
Here we evaluate how different design choices affect the performance of BOinG. Further ablation studies can be found in the appendix \ref{sec:appendix_ablation}.
\subsubsection{LGPGA vs. GP}
We study the extra benefit that a LGPGA brings to BOinG. In addition to LGPGA, we train a GP model only with the evaluations inside the subregion (\textit{local GP}) and another GP model that is trained with all previous evaluations but the candidates are restrained to be selected from the subregion(\textit{full GP local}). We evaluate the local GP models on the cartpole problem.

\begin{figure}[t]
  %\vspace{-2em}
  \centering
      \begin{subfigure}[b]{0.36\textwidth}
    \centering
    \includegraphics[width=1.0\textwidth]{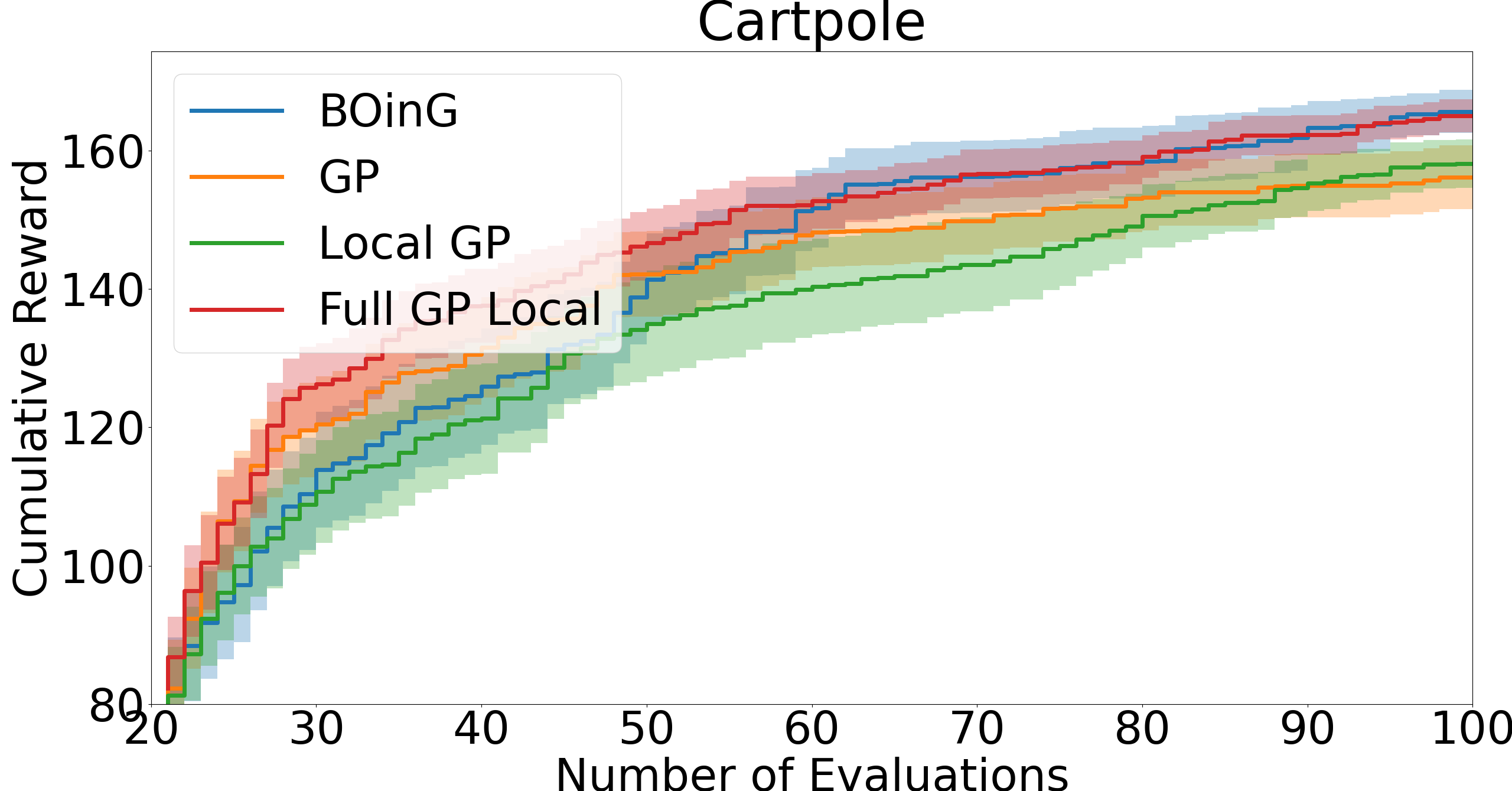}
    \end{subfigure}
    \hfill
    \begin{subfigure}[b]{0.36\textwidth}
    \centering
    \includegraphics[width=1.0\textwidth]{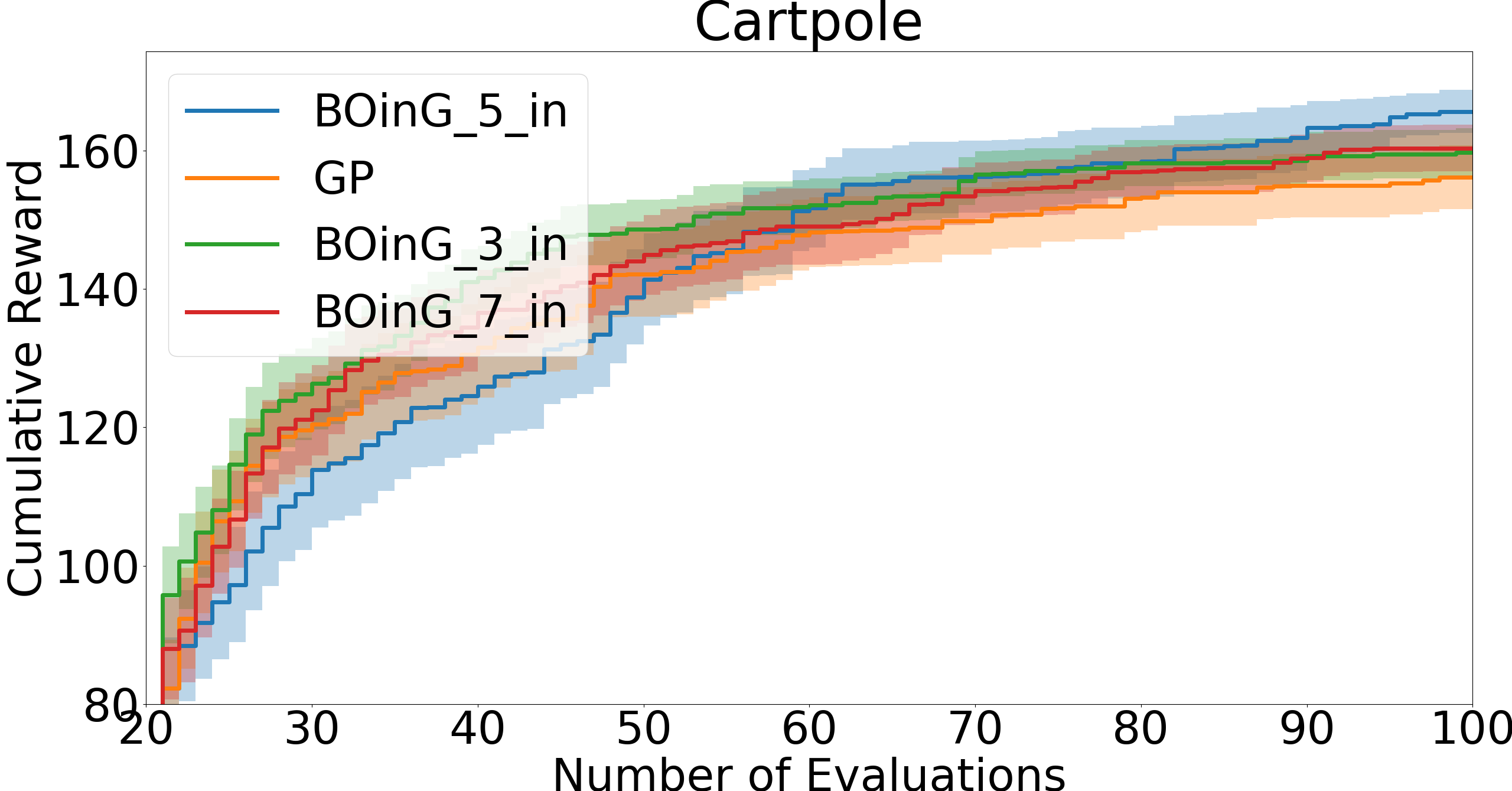}
    \end{subfigure}
    \hfill

    \caption{Ablation Study on LGPGA (Top) and $n_{min}$ (Bottom).\label{fig:ablation}}
\end{figure}

The result is shown in the left part of Figure~\ref{fig:ablation}. Both LGPGA and \textit{full GP local} has better final performances compared to a \textit{local GP}. However, consider the potential benefit that a LGPGA can bring (it requires less resources), we still suggest to use LGPGA in BOinG.
%The result is shown in the left part of Figure~\ref{fig:ablation}. PSGP achieves better any-time performances than  full GP local and local GP.  

%NeurIPS
% \subsubsection{BOinG+ vs. TuRBO}

% %NeurIPS

% %   \begin{wrapfigure}[8]{r}{0.7\textwidth}
% %   \vspace{-2em}
% %       \begin{subfigure}[b]{0.34\textwidth}
% %     \centering
% %     \includegraphics[width=1.0\textwidth]{images/ablation/ablation_psgp.png}
% %     %\includegraphics[width=1.0\textwidth]{images/res_synfunc/eggholder2.png}
% %     \end{subfigure}
% %     \hfill
% %     \begin{subfigure}[b]{0.34\textwidth}
% %     \centering
% %     \includegraphics[width=1.0\textwidth]{images/ablation/lunar_ablation.png}
% %     %\includegraphics[width=1.0\textwidth]{images/res_synfunc/eggholder2.png}
% %     \end{subfigure}
% %     \hfill

% %     \caption{Ablation Study on PSGP (left) and on BOinG+ (right).\label{fig:ablation}}
% % \end{wrapfigure}

%AAAI(TODO: FOR AAAI submission, move this to appendix)

% In Section \ref{sec:BOinG+}, we proposed to randomly switch between BOinG and TuRBO depending on their results accordingly. Here we will study if BOinG+ achieves a better exploitation-exploration tradeoff than TuRBO. 
% As we restart TuRBO earlier to allow more exploration, we first check if this strategy helps TuRBO to find a better configuration. According to Section \ref{sec:BOinG+}, we restart TuRBO if the length of the subregion is smaller than $2^{-4}$, called \textit{TuRBO\_4}. Additionally, we will study the extra benefit that BOinG+ brings to BOinG and TuRBO, we compare it with vanilla BOinG where BOinG never switches to TuRBO (\textit{BOinG\_Vanilla}) and a BOinG+ version where TuRBO never switches to BOinG (\textit{RF\_TuRBO}). The results are shown in the right part of Figure~\ref{fig:ablation}. Vanilla BOinG underperforms compared to the other variants. TuRBO\_4 becomes more explorative in the very beginning but cannot dig deeper as it restarts too early and is outperformed by RF\_TuRBO as it cannot capture the global data distribution. BOiNG+ instead allows further exploitation and finds the best configuration in the end.

%AAAI
\subsubsection{Number of points inside subregion}
BoinG introduces a special hyperparameter: $n_{min}$, the number of points inside the subregion. $n_{min}$ determines the size of the subregion to be explored in the next stage. Setting this value larger (e.g. to infinity) makes BOinG closer to a GP and thus requires more resources; reducing this value (e.g. until 0) leads BOinG to behave as an RF. As RF is often considered as a poor extrapolator, a BOinG with a small subregion might easily fall into a local minimum. Here we do an ablation study on $n_{min}$: \textit{BOinG\_$i$\_in} denotes a BOinG Optimizer that contains at least $i \cdot \ndims$ points inside its subregion. \textit{BOinG\_3\_in} achieves a better reward in the beginning but is then surpassed by \textit{BOinG\_5\_in}; \textit{BOinG\_7\_in} only starts to outperform a vanilla GP  after $\approx 60$ evaluations, which might be too late for a problem with 100-evaluation. We consider $5 \cdot \ndims$ as a plausible number of points kept in the subregion.

\section{Limitations}\label{sec:lim}
Even though BOinG is a robust approach, i.e. combining the best of GP-BO and RF-BO, GP-BO is often still the most efficient approach in low dimensional spaces. 
Furthermore, in higher dimensional spaces BO frameworks tend to suggest points near the boundary \citep{oh-icml18a} of the configuration space. Such points might not give the RF enough insights about how to extract a subregion, i.e., the subregion might still be too close to the boundary. %Additionally, BOinG does not estimate the potential of a subregion, which would require further (potentially unnecessary) exploration. One possible approach in future work might be to estimate the maximal value with Gumbel Sampling in that region~\citep{wang-pmlr17}. 
Last but not least, BOinG's combination of two surrogate models and two acquisition functions might be brittle in some applications; however in our experiments BOinG turned out to be surprisingly robust across different tasks.

\section{Discussion and Outlook\label{sec:discuss}}

In this paper, we proposed BOinG, a hierarchical approach that combines the best of a random forest on a global optimization level and a Gaussian Process on a local level. The underlying idea is that BOinG can better focus on promising subregions by having better local models and pinpoint the optimum with fewer function evaluations. Empirical experiments show that BOinG is a very robust approach and is able to outperform vanilla BO and other local BO approaches on HPO problems.
Although we focused on RFs and GPs as upper and lower level surrogate models,
BOinG can also be seen as a model-agnostic approach, which could adopt other surrogate models, such as TPE. %or BNNs~\cite{snoek-icml14a}.
In future work, we plan to study whether we can benefit from a data density estimation approach such as TPE or non-axis-aligned splits~\citep{wang2020} for better guidance on the upper level.
Another promising approach would be to combine Thompson sampling for BO~\citep{kandasamy-aistats18a} while maintaining multiple subregions~\citep{eriksson-nips19a} to perform batched BOinG efficiently.

%NeurIPS
\section{Broader Impact}\label{sec:impact}
BO is often applied for solving expensive black-box function optimization where each single evaluation consumes lots of resources. Experiments show that our method achieves a better anytime performance on different HPO problems and thus consumes less resources in the evaluation phases. Additionally, the introduction of LGPGA reduces the potential resource consumption in the optimization overhead. Furthermore, BOinG contributes to automated machine learning (AutoML) and thus to democratizing machine learning, with all its benefits and risks. Since we do not address any specific application, we do not expect any conflict with general ethical conduct.

% NeurIPS 

%\bibliographystyle{plain}
\bibliography{strings, lib, bibtex, proc}

\begin{thebibliography}{}

\bibitem[Assael et~al., 2014]{Assael14}
Assael, J., Wang, Z., and Freitas, N. (2014).
\newblock Heteroscedastic treed bayesian optimisation.
\newblock {\em CoRR}, abs/1410.7172.

\bibitem[Balandat et~al., 2020]{balandat2020botorch}
Balandat, M., Karrer, B., Jiang, D.~R., Daulton, S., Letham, B., Wilson, A.~G.,
  and Bakshy, E. (2020).
\newblock {BoTorch: A Framework for Efficient Monte-Carlo Bayesian
  Optimization}.
\newblock In {\em Advances in Neural Information Processing Systems 33}.

\bibitem[Bergstra et~al., 2011]{bergstra-nips11a}
Bergstra, J., Bardenet, R., Bengio, Y., and K{\'e}gl, B. (2011).
\newblock Algorithms for hyper-parameter optimization.
\newblock In Shawe-Taylor, J., Zemel, R., Bartlett, P., Pereira, F., and
  Weinberger, K., editors, {\em Proceedings of the 25th International
  Conference on Advances in Neural Information Processing Systems
  ({N}eur{IPS}'11)}, pages 2546--2554.

\bibitem[Bergstra and Bengio, 2012]{bergstra-jmlr12a}
Bergstra, J. and Bengio, Y. (2012).
\newblock Random search for hyper-parameter optimization.
\newblock {\em Journal of Machine Learning Research}, 13:281--305.

\bibitem[Bergstra et~al., 2013]{bergstra-icml13a}
Bergstra, J., Yamins, D., and Cox, D. (2013).
\newblock Making a science of model search: Hyperparameter optimization in
  hundreds of dimensions for vision architectures.
\newblock In Dasgupta, S. and McAllester, D., editors, {\em Proceedings of the
  30th International Conference on Machine Learning ({ICML}'13)}, pages
  115--123. Omnipress.

\bibitem[Breimann, 2001]{breimann-mlj01a}
Breimann, L. (2001).
\newblock Random forests.
\newblock {\em Machine Learning Journal}, 45:5--32.

\bibitem[Brockman et~al., 2016]{brockman-arxiv16-gym}
Brockman, G., Cheung, V., Pettersson, L., Schneider, J., Schulman, J., Tang,
  J., and Zaremba, W. (2016).
\newblock Open{AI} {G}ym.
\newblock {\em arXiv:1606.01540 [cs.LG]}.

\bibitem[Candela and Rasmussen, 2005]{candela-jmlr6a}
Candela, J. and Rasmussen, C. (2005).
\newblock A unifying view of sparse approximate gaussian process regression.
\newblock {\em J. Mach. Learn. Res.}, 6:1939--1959.

\bibitem[Dy and Krause, 2018]{icml18}
Dy, J. and Krause, A., editors (2018).
\newblock {\em Proceedings of the 35th International Conference on Machine
  Learning ({ICML}'18)}, volume~80. Proceedings of Machine Learning Research.

\bibitem[Eggensperger et~al., 2013]{eggensperger-bayesopt13}
Eggensperger, K., Feurer, M., Hutter, F., Bergstra, J., Snoek, J., Hoos, H.,
  and Leyton-Brown, K. (2013).
\newblock Towards an empirical foundation for assessing {Bayesian} optimization
  of hyperparameters.
\newblock In {\em {NeurIPS} Workshop on {B}ayesian Optimization in Theory and
  Practice (BayesOpt'13)}.

\bibitem[Eggensperger et~al., 2021]{eggensperger2021hpobench}
Eggensperger, K., Müller, P., Mallik, N., Feurer, M., Sass, R., Klein, A.,
  Awad, N., Lindauer, M., and Hutter, F. (2021).
\newblock Hpobench: A collection of reproducible multi-fidelity benchmark
  problems for hpo.

\bibitem[Eriksson et~al., 2019]{eriksson-nips19a}
Eriksson, D., Pearce, M., Gardner, J., Turner, R., and Poloczek, M. (2019).
\newblock Scalable global optimization via local bayesian optimization.
\newblock In {\em Advances in Neural Information Processing Systems},
  volume~32, pages 5496--5507.

\bibitem[Falkner et~al., 2018]{falkner-icml18a}
Falkner, S., Klein, A., and Hutter, F. (2018).
\newblock {BOHB}: Robust and efficient hyperparameter optimization at scale.
\newblock In \cite{icml18}, pages 1437--1446.

\bibitem[Feurer and Hutter, 2019]{feurer-bookchapter19a}
Feurer, M. and Hutter, F. (2019).
\newblock Hyperparameter optimization.
\newblock pages 3--38. Springer.
\newblock Available for free at http://automl.org/book.

\bibitem[Gardner et~al., 2018]{Gardner18}
Gardner, J., Pleiss, G., Weinberger, Q., Bindel, D., and Wilson, A. (2018).
\newblock Gpytorch: Blackbox matrix-matrix gaussian process inference with gpu
  acceleration.
\newblock In {\em Advances in Neural Information Processing Systems},
  volume~31, pages 7576--7586. Curran Associates, Inc.

\bibitem[Gramacy et~al., 2004]{gramacy-icml04}
Gramacy, R., Lee, H., and Macready, W. (2004).
\newblock Parameter space exploration with gaussian process trees.
\newblock In Greiner, R., editor, {\em Proceedings of the 21st International
  Conference on Machine Learning ({ICML}'04)}, pages 45--52. Omnipress.

\bibitem[Hensman et~al., 2013]{hensman-auai2013}
Hensman, J., Fusi, N., and Lawrence, N.~D. (2013).
\newblock Gaussian processes for big data.
\newblock In {\em Proceedings of the Twenty-Ninth Conference on Uncertainty in
  Artificial Intelligence}, UAI'13, page 282–290, Arlington, Virginia, USA.
  AUAI Press.

\bibitem[Hutter et~al., 2011]{hutter-lion11a}
Hutter, F., Hoos, H., and Leyton-Brown, K. (2011).
\newblock Sequential model-based optimization for general algorithm
  configuration.
\newblock In Coello, C., editor, {\em Proceedings of the Fifth International
  Conference on Learning and Intelligent Optimization ({LION}'11)}, volume 6683
  of {\em Lecture Notes in Computer Science}, pages 507--523. Springer.

\bibitem[Hutter et~al., 2012]{hutter-lion12a}
Hutter, F., Hoos, H., and Leyton-Brown, K. (2012).
\newblock Parallel algorithm configuration.
\newblock In Hamadi, Y. and Schoenauer, M., editors, {\em Proceedings of the
  Sixth International Conference on Learning and Intelligent Optimization
  ({LION}'12)}, volume 7219 of {\em Lecture Notes in Computer Science}, pages
  55--70. Springer.

\bibitem[Jones et~al., 1998]{jones-jgo98a}
Jones, D., Schonlau, M., and Welch, W. (1998).
\newblock Efficient global optimization of expensive black box functions.
\newblock {\em Journal of Global Optimization}, 13:455--492.

\bibitem[Kandasamy et~al., 2018]{kandasamy-aistats18a}
Kandasamy, K., Krishnamurthy, A., Schneider, J., and Póczos, B. (2018).
\newblock Parallelised {B}ayesian optimisation via {T}hompson sampling.
\newblock In \cite{aistats18}, pages 133--142.

\bibitem[Kandasamy et~al., 2015]{kandasamy-icml15a}
Kandasamy, K., Schneider, J., and Póczos, B. (2015).
\newblock High {Dimensional} {Bayesian} {Optimisation} and {Bandits} via
  {Additive} {Models}.
\newblock In Bach, F. and Blei, D., editors, {\em Proceedings of the 32nd
  International Conference on Machine Learning ({ICML}'15)}, volume~37, pages
  295--304. Omnipress.

\bibitem[Klein et~al., 2017]{klein-ejs17}
Klein, A., Falkner, S., Bartels, S., Hennig, P., and Hutter, F. (2017).
\newblock Fast b{a}yesian hyperparameter optimization on large datasets.
\newblock In {\em Electronic Journal of Statistics}, volume~11, page
  4945–4968.

\bibitem[Kuhnle et~al., 2017]{tensorforce}
Kuhnle, A., Schaarschmidt, M., and Fricke, K. (2017).
\newblock Tensorforce: a tensorflow library for applied reinforcement learning.
\newblock Web page.

\bibitem[Lindauer et~al., 2021]{lindauer2021smac3}
Lindauer, M., Eggensperger, K., Feurer, M., Biedenkapp, A., Deng, D.,
  Benjamins, C., Sass, R., and Hutter, F. (2021).
\newblock Smac3: A versatile bayesian optimization package for hyperparameter
  optimization.

\bibitem[McLeod et~al., 2018]{McLeodRO-icml18}
McLeod, M., Roberts, S.~J., and Osborne, M.~A. (2018).
\newblock Optimization, fast and slow: Optimally switching between local and
  bayesian optimization.
\newblock In Dy, J.~G. and Krause, A., editors, {\em Proceedings of the 35th
  International Conference on Machine Learning, {ICML} 2018}, volume~80 of {\em
  Proceedings of Machine Learning Research}, pages 3440--3449. {PMLR}.

\bibitem[Oh et~al., 2018]{oh-icml18a}
Oh, C., Gavves, E., and Welling, M. (2018).
\newblock {BOCK} : {Bayesian} {Optimization} with {Cylindrical} {Kernels}.
\newblock In \cite{icml18}, pages 3865--3874.

\bibitem[Pimenta et~al., 2020]{pimenta-evocc20a}
Pimenta, C., de~S{\'{a}}, A., Ochoa, G., and Pappa, G. (2020).
\newblock Fitness landscape analysis of automated machine learning search
  spaces.
\newblock In {\em Proceedings of Evolutionary Computation in Combinatorial
  Optimization}, pages 114--130. Springer.

\bibitem[Pushak and Hoos, 2018]{Pushak-ppsn18a}
Pushak, Y. and Hoos, H. (2018).
\newblock Algorithm configuration landscapes: - more benign than expected?
\newblock In {\em Parallel Problem Solving from Nature - {PPSN} {XV} - 15th
  International Conference, Coimbra, Portugal, September 8-12, 2018,
  Proceedings, Part {II}}, pages 271--283. Springer.

\bibitem[Salimbeni et~al., 2018]{salimbeni-pmlr18a}
Salimbeni, H., Eleftheriadis, S., and Hensman, J. (2018).
\newblock Natural gradients in practice: Non-conjugate variational inference in
  gaussian process models.
\newblock In Storkey, A. and Perez-Cruz, F., editors, {\em Proceedings of the
  Twenty-First International Conference on Artificial Intelligence and
  Statistics}, volume~84 of {\em Proceedings of Machine Learning Research},
  pages 689--697. PMLR.

\bibitem[Schulman et~al., 2017]{schulman-arxiv17a}
Schulman, J., Wolski, F., Dhariwal, P., Radford, A., and Klimov, O. (2017).
\newblock Proximal policy optimization algorithms.
\newblock {\em arXiv:1707.06347 [cs.LG]}.

\bibitem[Shahriari et~al., 2016]{shahriari-ieee16a}
Shahriari, B., Swersky, K., Wang, Z., Adams, R., and de~Freitas, N. (2016).
\newblock Taking the human out of the loop: {A} review of {B}ayesian
  optimization.
\newblock {\em Proceedings of the {IEEE}}, 104(1):148--175.

\bibitem[Snelson and Ghahramani, 2006]{snelon-nips06a}
Snelson, E. and Ghahramani, Z. (2006).
\newblock Sparse gaussian processes using pseudo-inputs.
\newblock In {\em Advances in Neural Information Processing Systems},
  volume~18, pages 1257--1264. MIT Press.

\bibitem[Snoek et~al., 2012]{snoek-nips12a}
Snoek, J., Larochelle, H., and Adams, R. (2012).
\newblock Practical {B}ayesian optimization of machine learning algorithms.
\newblock In Bartlett, P., Pereira, F., Burges, C., Bottou, L., and Weinberger,
  K., editors, {\em Proceedings of the 26th International Conference on
  Advances in Neural Information Processing Systems ({N}eur{IPS}'12)}, pages
  2960--2968.

\bibitem[Storkey and Perez-Cruz, 2018]{aistats18}
Storkey, A. and Perez-Cruz, F., editors (2018).
\newblock {\em Proceedings of the 21st International Conference on Artificial
  Intelligence and Statistics ({AISTATS})}, volume~84. Proceedings of Machine
  Learning Research.

\bibitem[Surjanovic and Bingham, 2013]{simulationlib}
Surjanovic, S. and Bingham, D. (2013).
\newblock Virtual library of simulation experiments: Test functions and
  datasets.
\newblock Retrieved January 28, 2021, from \url{http://www.sfu.ca/~ssurjano}.

\bibitem[Thornton et~al., 2013]{thornton-kdd13a}
Thornton, C., Hutter, F., Hoos, H., and Leyton-Brown, K. (2013).
\newblock {A}uto-{WEKA}: combined selection and hyperparameter optimization of
  classification algorithms.
\newblock In Dhillon, I., Koren, Y., Ghani, R., Senator, T., Bradley, P.,
  Parekh, R., He, J., Grossman, R., and Uthurusamy, R., editors, {\em The 19th
  {ACM} {SIGKDD} International Conference on Knowledge Discovery and Data
  Mining ({KDD}'13)}, pages 847--855. ACM Press.

\bibitem[Titsias, 2009]{titsias-pmlr09a}
Titsias, M. (2009).
\newblock Variational learning of inducing variables in sparse gaussian
  processes.
\newblock In van Dyk, D. and Welling, M., editors, {\em Proceedings of the
  Twelth International Conference on Artificial Intelligence and Statistics},
  volume~5 of {\em Proceedings of Machine Learning Research}, pages 567--574,
  Hilton Clearwater Beach Resort, Clearwater Beach, Florida USA. PMLR.

\bibitem[Turner, 2019]{bayesmark}
Turner, R. (2019).
\newblock The bayes opt benchmark documentation.
\newblock \url{https://bayesmark.readthedocs.io/en/latest/}.

\bibitem[Wan et~al., 2021]{wan2021think}
Wan, X., Nguyen, V., Ha, H., Ru, B., Lu, C., and Osborne, M.~A. (2021).
\newblock Think global and act local: Bayesian optimisation over
  high-dimensional categorical and mixed search spaces.

\bibitem[Wang et~al., 2020]{wang2020}
Wang, L., Fonseca, R., and Tian, Y. (2020).
\newblock Learning search space partition for black-box optimization using
  monte carlo tree search.
\newblock In {\em Advances in Neural Information Processing Systems 33: Annual
  Conference on Neural Information Processing Systems}.

\bibitem[Wang et~al., 2018]{wang-aistats18a}
Wang, Z., Gehring, C., Kohli, P., and Jegelka, S. (2018).
\newblock Batched {Large}-scale {Bayesian} {Optimization} in {High}-dimensional
  {Spaces}.
\newblock In \cite{aistats18}, pages 745--754.

\bibitem[Wang et~al., 2014]{wang-aistats14}
Wang, Z., Shakibi, B., Jin, L., and Freitas, N. (2014).
\newblock Bayesian multi-scale optimistic optimization.
\newblock In {\em Proceedings of the Seventeenth International Conference on
  Artificial Intelligence and Statistics, {AISTATS} 2014}, volume~33 of {\em
  {JMLR} Workshop and Conference Proceedings}, pages 1005--1014. JMLR.org.

\end{thebibliography}


\begin{thebibliography}{}

\bibitem[Eriksson et~al., 2019]{eriksson-nips19a}
Eriksson, D., Pearce, M., Gardner, J., Turner, R., and Poloczek, M. (2019).
\newblock Scalable global optimization via local bayesian optimization.
\newblock In {\em Advances in Neural Information Processing Systems},
  volume~32, pages 5496--5507.

\bibitem[Hensman et~al., 2013]{hensman-auai2013}
Hensman, J., Fusi, N., and Lawrence, N.~D. (2013).
\newblock Gaussian processes for big data.
\newblock In {\em Proceedings of the Twenty-Ninth Conference on Uncertainty in
  Artificial Intelligence}, UAI'13, page 282–290, Arlington, Virginia, USA.
  AUAI Press.

\bibitem[Hutter et~al., 2012]{hutter-lion12a}
Hutter, F., Hoos, H., and Leyton-Brown, K. (2012).
\newblock Parallel algorithm configuration.
\newblock In Hamadi, Y. and Schoenauer, M., editors, {\em Proceedings of the
  Sixth International Conference on Learning and Intelligent Optimization
  ({LION}'12)}, volume 7219 of {\em Lecture Notes in Computer Science}, pages
  55--70. Springer.

\bibitem[Jankowiak et~al., 2020]{jankowiak-pmlr20a}
Jankowiak, M., Pleiss, G., and Gardner, J. (2020).
\newblock Parametric {G}aussian process regressors.
\newblock In III, H.~D. and Singh, A., editors, {\em Proceedings of the 37th
  International Conference on Machine Learning}, volume 119 of {\em Proceedings
  of Machine Learning Research}, pages 4702--4712. PMLR.

\bibitem[Lindauer et~al., 2021]{lindauer2021smac3}
Lindauer, M., Eggensperger, K., Feurer, M., Biedenkapp, A., Deng, D.,
  Benjamins, C., Sass, R., and Hutter, F. (2021).
\newblock Smac3: A versatile bayesian optimization package for hyperparameter
  optimization.

\bibitem[Liu and Nocedal, 1989]{liu-lbfgs89}
Liu, D.~C. and Nocedal, J. (1989).
\newblock On the limited memory {BFGS} method for large scale optimization.
\newblock {\em Math. Program.}, 45(1-3):503--528.

\bibitem[Salimans et~al., 2017]{salimans-arxiv17a}
Salimans, T., Ho, J., Chen, X., and Sutskever, I. (2017).
\newblock Evolution strategies as a scalable alternative to reinforcement
  learning.
\newblock {\em arXiv:1703.03864 [stat.ML]}.

\bibitem[Storkey and Perez-Cruz, 2018]{aistats18}
Storkey, A. and Perez-Cruz, F., editors (2018).
\newblock {\em Proceedings of the 21st International Conference on Artificial
  Intelligence and Statistics ({AISTATS})}, volume~84. Proceedings of Machine
  Learning Research.

\bibitem[Titsias, 2009]{titsias-pmlr09a}
Titsias, M. (2009).
\newblock Variational learning of inducing variables in sparse gaussian
  processes.
\newblock In van Dyk, D. and Welling, M., editors, {\em Proceedings of the
  Twelth International Conference on Artificial Intelligence and Statistics},
  volume~5 of {\em Proceedings of Machine Learning Research}, pages 567--574,
  Hilton Clearwater Beach Resort, Clearwater Beach, Florida USA. PMLR.

\bibitem[Wang et~al., 2020]{wang2020}
Wang, L., Fonseca, R., and Tian, Y. (2020).
\newblock Learning search space partition for black-box optimization using
  monte carlo tree search.
\newblock In {\em Advances in Neural Information Processing Systems 33: Annual
  Conference on Neural Information Processing Systems}.

\bibitem[Wang et~al., 2018]{wang-aistats18a}
Wang, Z., Gehring, C., Kohli, P., and Jegelka, S. (2018).
\newblock Batched {Large}-scale {Bayesian} {Optimization} in {High}-dimensional
  {Spaces}.
\newblock In \cite{aistats18}, pages 745--754.

\bibitem[Yuan and Wahba, 2004]{yuan2004}
Yuan, M. and Wahba, G. (2004).
\newblock Doubly penalized likelihood estimator in heteroscedastic regression.
\newblock {\em Statistics \& Probability Letters}, 69(1):11 -- 20.

\end{thebibliography}

\end{document}

\appendix
\newpage

\section{Algorithms to Extract Subregions}
\begin{algorithm}[tb]
 \caption{Subregion Selection with Random Forest}\label{alg:subregion}

\begin{algorithmic}[1]
 \STATE {\bfseries Input:} Search Space $\pcs \in \mathbb{R}^n$; candidate selected by global Bayesian Optimization $\candidate_g$, random forest model $\surro$ with $n_{tree}$ trees; minimal number of points stored in the subregion $n_{min}$ %; weak upper bound of points stored in the subregion $n_{max}$ 
 \STATE {\bfseries Output:} subregion $\pcs_{sub}$ extracted from $\pcs$
 \STATE {\bfseries Initialization: } $\observe_{sub} \leftarrow \pcs $; a list of root nodes $S$ of RF~$\surro$ , a nodes indicator $I \leftarrow [False] * n_{trees}$ indicating that if we will stop deeper from the node.
\WHILE {no element in $I$ is False \label{alg:subregioniter}}
  \FOR {each node $ s_{p} \in S $}
    \STATE $s_{p}' \leftarrow child(s_{p})$ with $\candidate_g \in s_{p}'$
    \STATE Let $\observe_{s_{p}'}$ be all observed points in $s_{p}'$
    \IF {$|\observe_{sub} \cap \observe_{s_{p}'}| > n_{min}$}
      \STATE $s \leftarrow s'$
      \STATE $\observe_{sub} \leftarrow \ \observe_{sub} \cap  \observe_{s'}$
    \ELSE \label{alg:subregionstop}
      \STATE $I_p \leftarrow True$ 
    \ENDIF 
  \ENDFOR
  %\IF {$|\observe_{sub} \cap \observe_{s_{p}'}| > n_{max}$}
    %\STATE repeat line \ref{alg:subregioniter}  except that in line  \ref{alg:subregionstop}, we go further deeper without shrinking the subregion
  %\ENDIF
\ENDWHILE
 
 \STATE{\bfseries Return:} An extracted subregion $\pcs_{sub}$ 
\end{algorithmic}
\end{algorithm}

\section{Implementation Details}

We start with a Gaussian Process that is trained on all the previous evaluated data points until $n_{min}$ points exist in the dataset. Then we start to extract a subregion, as described in Algorithm \ref{alg:subregion}. It makes no sense to introduce LGPGA if the number of $\fout$ is smaller than $n_\inducing$. In such a case, we still train a full GP model on all the previously evaluated points but only optimize the acquisition function values in the subregion. If we have more observations, we apply LGPGA to reduce the potential computation time.

\subsection{LGPGA Details}
We train our LGPGA model with the following manner: we first train a GP model to fit $\fin$ to acquire the kernel hyperparameters. Then we use this optimized kernel to initialize a sparse GP and keep its kernel hyperparameters fixed, i.e. we approximate $\fout$ by only optimizing the position of the inducing points. The hyperparameters of LGPGA thus captures both $\fin$ (the kernel hyperparameters) and $\fout$ (the inducing points positions). We apply variational GP~\citep{titsias-pmlr09a, hensman-auai2013} to approximate $\fout$ with its hyperparameters optimized with natural gradient descend~\cite{salimans-arxiv17a}.  We train the sparse GP to approximate the predictive variational evidence lower bound (ELBO), as proposed in~\cite{jankowiak-pmlr20a}. A further advantage of a variational GP is that it allows stochastic variational inference (SVI), i.e. we could scale variational GP to even larger dataset as the number of evaluations grows. However, in this paper, we only consider batch optimization.

We set number of inducing points to be at least $min(2 \times n_{dims}, 10)$ and it grows linearly as number of total evaluations grows ($min(\frac{n_{\observe}}{20})$ where $\observe$ denote all the previous evaluations) and up to $50$ points.

\subsection{BOinG+ Details}
When working with larger budgets (e.g. larger than 500 evaluations), as Random Forest might converge to a local minimum, we combine TuRBO~\citep{eriksson-nips19a} and BOinG in the following ways: we start with BOinG and set a failure counter $\failcounter$. We increase it every $n_{dim}$ times when we have not found a better configuration or decrease it if a better configuration was found. The probability of switching to TuRBO is computed as:
\begin{equation}\label{eq:switch_prob}
    \prob_{switch} = 0.1 * \failcounter
\end{equation}

As BOinG potentially acts as an exploitation mechanism, we will let TuRBO focus more on exploration: we randomly sample 20 configurations $\candidates_{random} \in \pcs$ and extract their subregions accordingly with Algorithm~\ref{alg:subregion}. We take the subspace $\pcs_{sub}$ with the largest volume and build a TuRBO optimizer inside $\pcs_{sub}$. The TuRBO optimizer is initialized with the points inside this subspace.  \cite{eriksson-nips19a} restarts TuRBO if the length of the subregion is smaller than $2^{-7}$. We restart TuRBO if the length of the subregion is smaller than $2^{-4}$ instead to allow more repetitions. Similarly, we adjust the probability of switching to BOinG with Equation \ref{eq:switch_prob}. However, if TuRBO finds a better configuration, we switch back directly to BOinG for further exploitation. Each time when we switch between BOinG and TuRBO, we halve their failure counts $\failcounter$ accordingly to avoid too frequent switching.

\section{Experiments Details}
  For Adult and Cartpole problems, we allow $2\cdot n_{dims}$ initial points and they are initialized with a Sobol sequence, except for TuRBO and LA-MCTS. The deterministic Sobol sequence is not applicable to TuRBO and LA-MCTS since they require the randomness by initialization to restart with different points. All the optimizer runs are repeated 30 times.
  
  For the CartPole benchmark, we optimized the reward value achieved by the agent after a maximum of 200 episodes or 6000 steps---whichever was reached first. The final performance is the mean reward of 20 episodes by the trained agent. To reduce the impact of noise, for each hyperparameter configuration, we repeatedly evaluate 9 runs and return the mean value of the final cost value. 
  
   Lunar Lander and Robot Pushing tasks are applied in ~\cite{eriksson-nips19a}. We set the same search space as TuRBO did\footnote{\url{https://github.com/uber-research/TuRBO}} due to the large amount of time required for each evaluation. We only run 20 repetitions for these two benchmarks. For the lunar lander problem, we allocate a budget of 1500 function evaluations and for robot pushing, a budget of 3000 function evaluations. 
   
   All the related data can be found under \url{https://figshare.com/s/757b2b74d77586690458}.

\section{Ablation Study}
In this section,  we first continue our ablation study in Section 4.3 of the main paper but on different tasks. Then we will discuss the choices that we adapted to different tasks. 

\subsection{Ablation Study on Adult}
    \begin{figure}[ht]
        \centering
        \begin{subfigure}[b]{0.48\textwidth}
        \centering
          \includegraphics[width=1.0\textwidth]{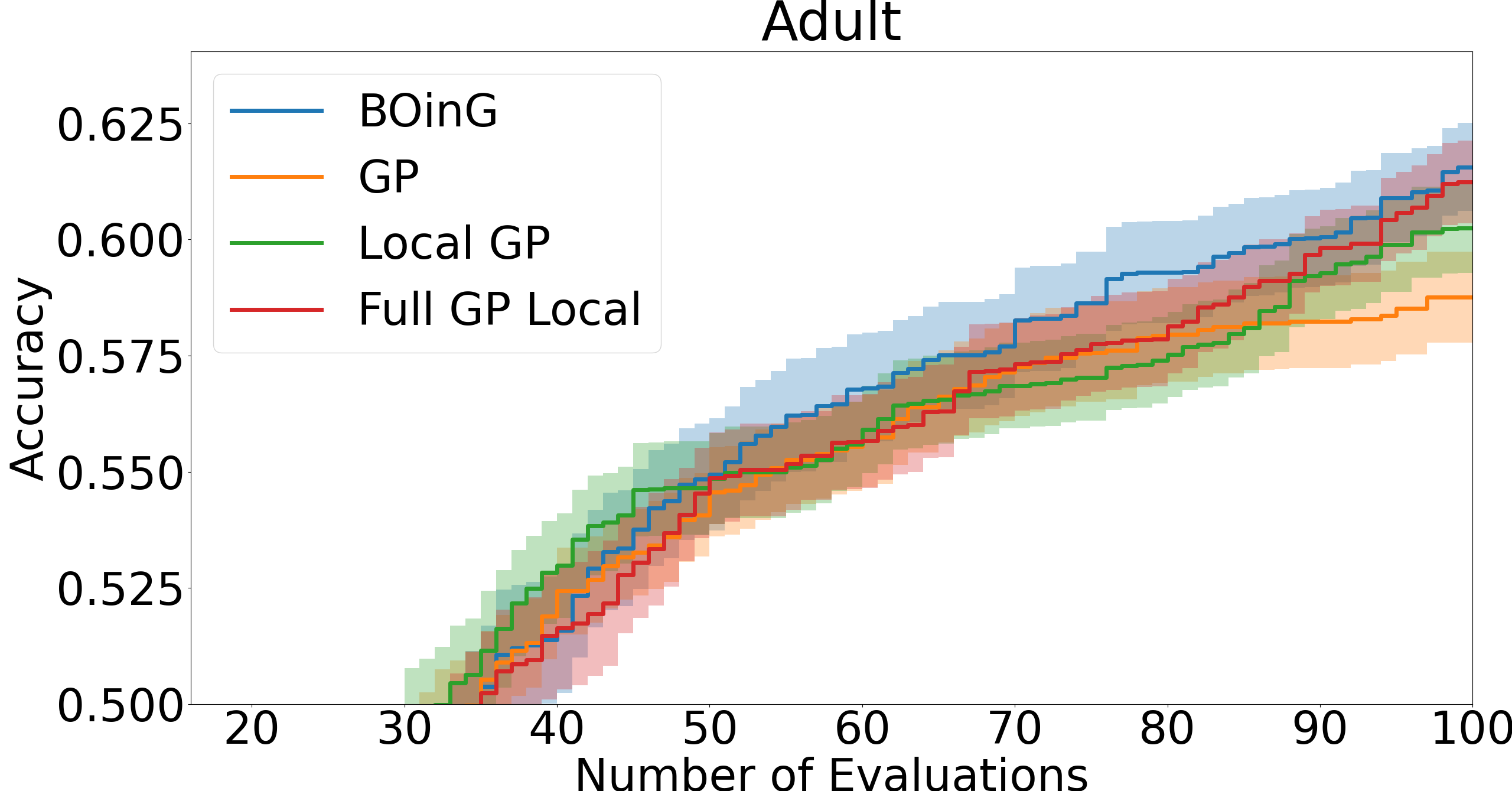}
        \end{subfigure}
        \begin{subfigure}[b]{0.48\textwidth}
        \centering
          \includegraphics[width=1.0\textwidth]{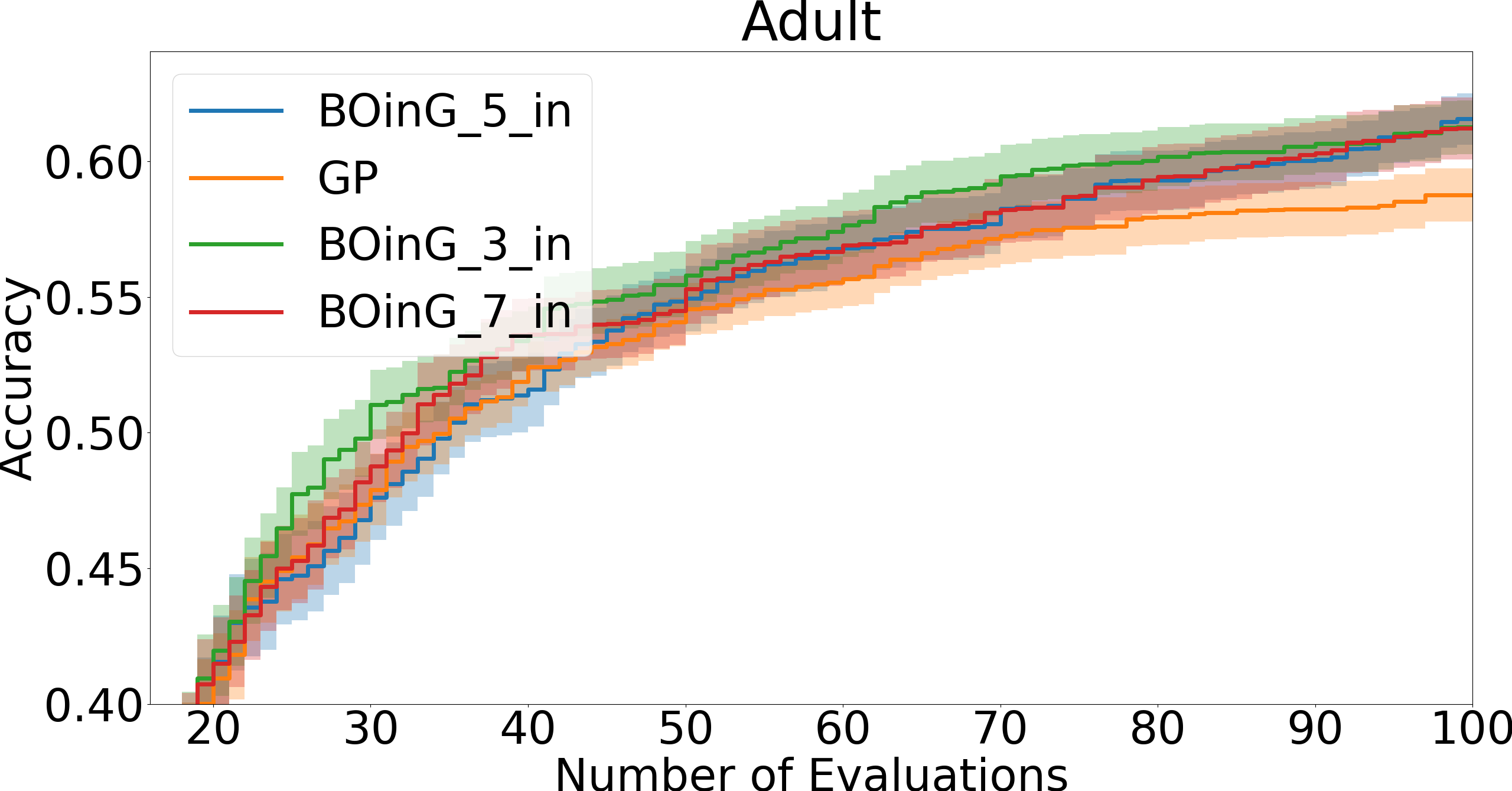}
        \end{subfigure}
        \caption{Ablation Study on LGPGA (Top) and $n_{min}$ (Bottom) on the Adult task.\label{fig:ablation_adult}}
    \end{figure}
Figure~\ref{fig:ablation_adult} shows how different choices of GPs and $n_{min}$ influences the optimization process on the Adult task. LGPGA and \textit{full GP local} still outperform \textit{Local GP} and acquire similar final performance. However, different choices of $n_{min}$ does not affect the performance on the Adult task too much, all the BOinG-variation has similar performance. This also indicates that the landscape of the target problem might also influence the final performance of the optimizer.

\subsection{Ablation Study on the choice of acquisition function at the global level}

\begin{figure}[ht]
        \centering
        \begin{subfigure}[b]{0.48\textwidth}
        \centering
          \includegraphics[width=1.0\textwidth]{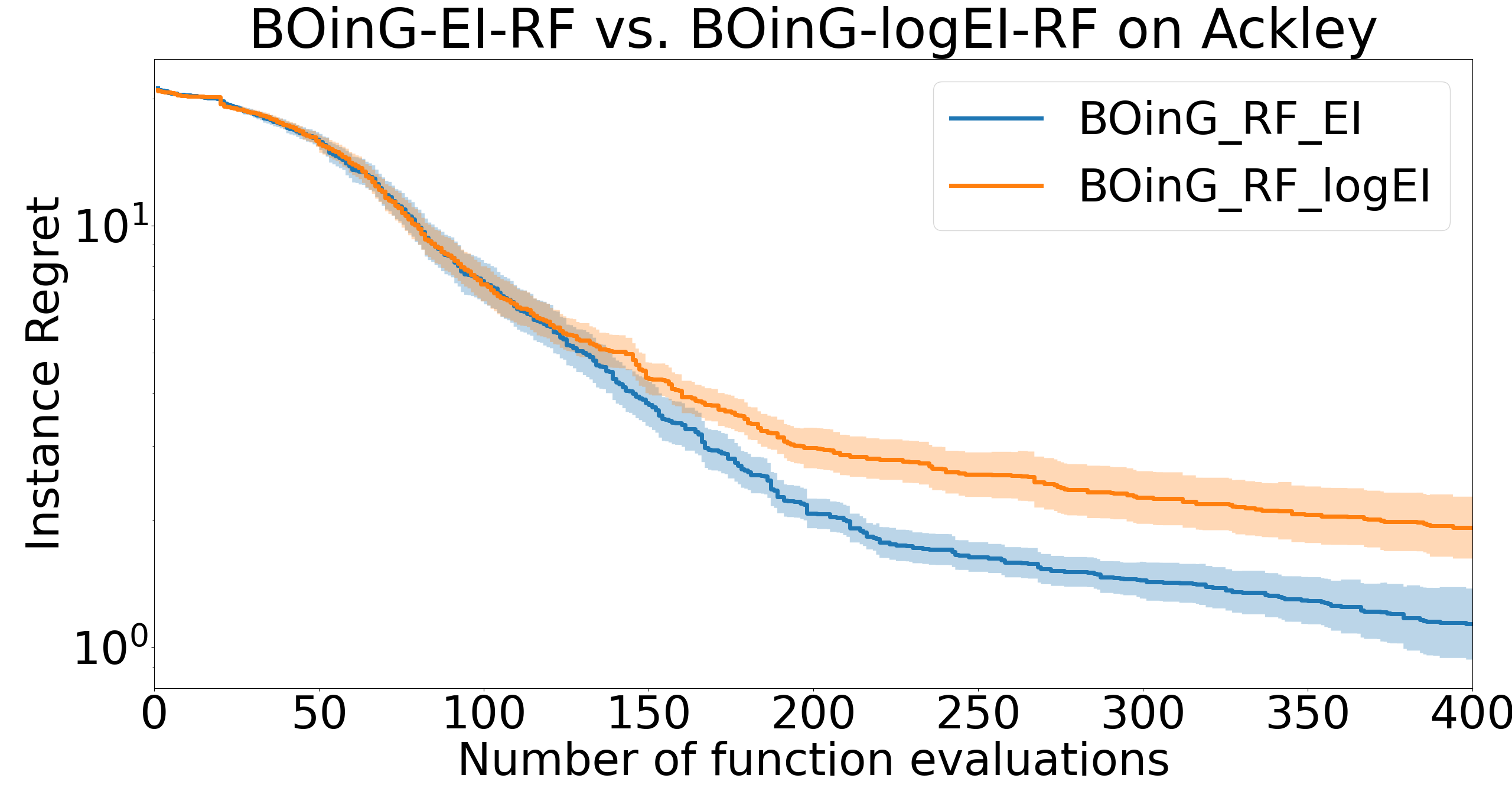}
        \end{subfigure}
        \begin{subfigure}[b]{0.48\textwidth}
        \centering
          \includegraphics[width=1.0\textwidth]{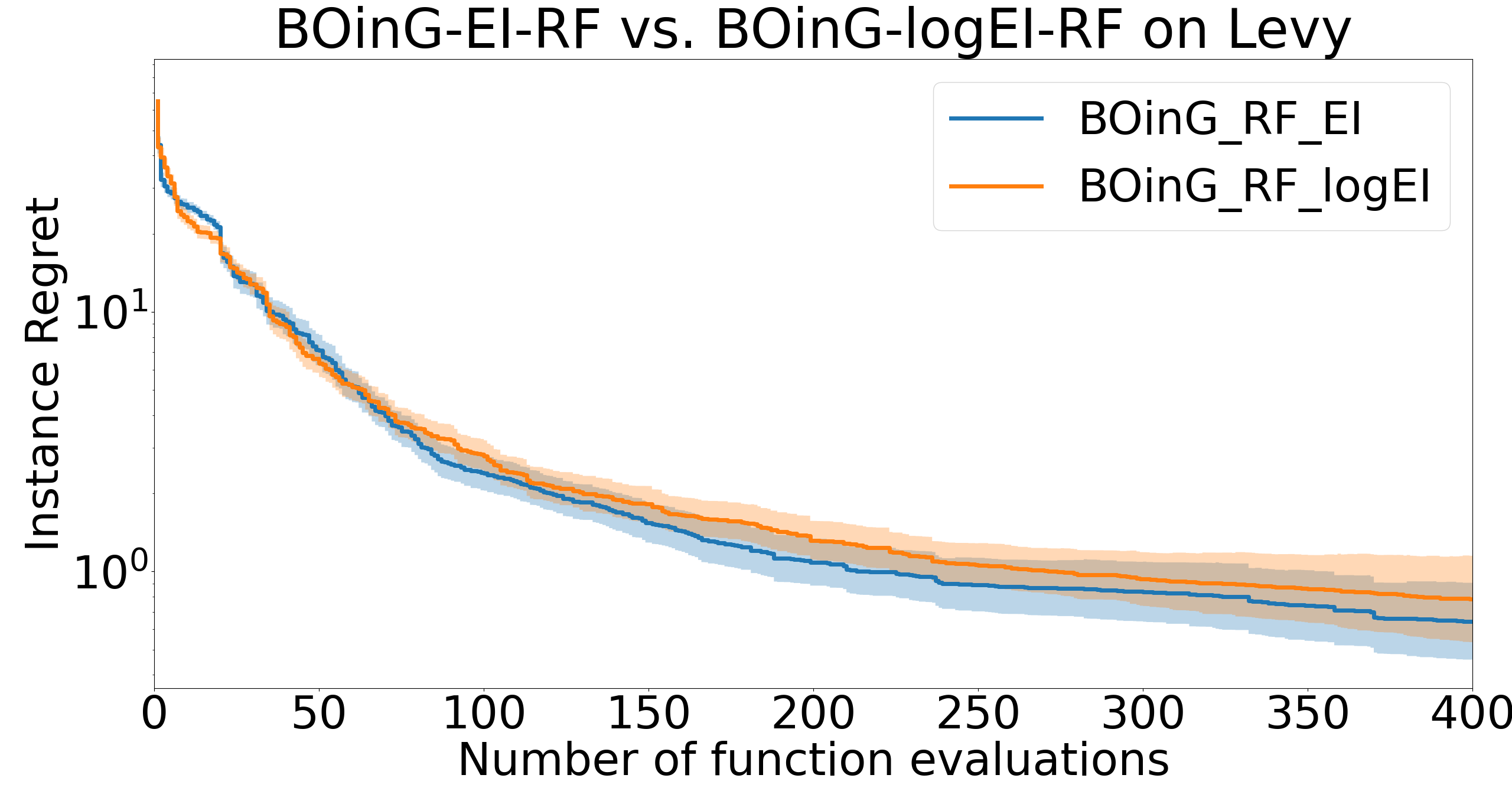}
        \end{subfigure}
        \caption{Ablation Study on the choice of the acquisition function at the global level on ackley (top) and levy(middle) function.\label{fig:ablation_logei}}
    \end{figure}
Instead of the default setting of logEI~\citep{hutter-lion12a}, we apply vanilla EI as the acquisition function at the global level on synthetic functions. These synthetic functions usually do not fit the requirement of "heavy-tailed cost distributions"~\citep{lindauer2021smac3} and possess several local minima that an optimizer might easily fall into. Applying EI allows the optimizer to explore the unknown regions more and thus provides the optimizer a higher probability to find the global optimum. The result is illustrated in Figure~\ref{fig:ablation_logei}. BOinG equipped with EI as the acquisition function at the global level generally has a better any time performance.

\subsection{Ablation study on BOinG+}

\begin{figure}[ht]
        \centering

        \begin{subfigure}[b]{0.48\textwidth}
        \centering
          \includegraphics[width=1.0\textwidth]{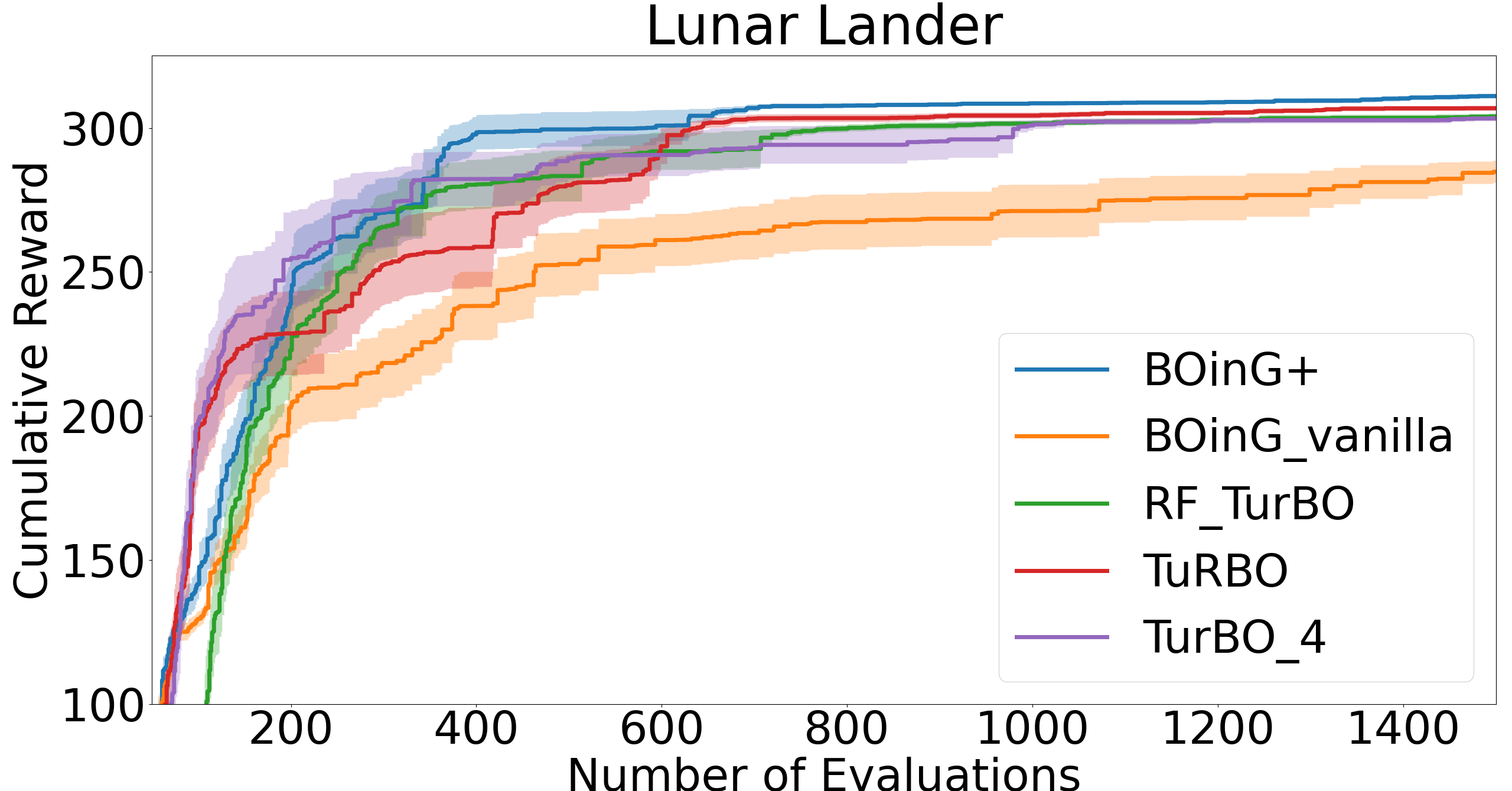}
        \end{subfigure}
        \caption{Ablation Study on BOinG+.\label{fig:ablation_boing+}}
    \end{figure}

In Section 3.3, we proposed to randomly switch between BOinG and TuRBO depending on their results accordingly. Here we will study if BOinG+ achieves a better exploitation-exploration tradeoff than TuRBO. 
As we restart TuRBO earlier to allow more exploration, we first check if this strategy helps TuRBO to find a better configuration. According to Section 3.3, we restart TuRBO if the length of the subregion is smaller than $2^{-4}$, called \textit{TuRBO\_4}. Additionally, we will study the extra benefit that BOinG+ brings to BOinG and TuRBO, we compare it with vanilla BOinG where BOinG never switches to TuRBO (\textit{BOinG\_Vanilla}) and a BOinG+ version where TuRBO never switches to BOinG (\textit{RF\_TuRBO}). The results are shown in Figure~\ref{fig:ablation_boing+}. Vanilla BOinG underperforms compared to the other variants. TuRBO\_4 becomes more explorative in the very beginning but cannot dig deeper as it restarts too early and is outperformed by RF\_TuRBO as it cannot capture the global data distribution. BOiNG+ instead allows further exploitation and finds the best configuration in the end.

\section{Scalability\label{sec:scalbility}}
  One drawback of a Gaussian Process is its cubic complexity with respect to the number of the points ever evaluated. Similar to other local BO approaches~\citep{eriksson-nips19a, wang-aistats18a, wang2020}, our approach alleviates this problem by developing a model only with a subset of the previous evaluations. However, a local GP still scales cubically w.r.t. the number of points inside the subregion (and does not capture the overall trend). Hence the complexity of our algorithms mainly depends on the number of points inside the subregion. 
  
  \subsection{Complexity of Partial Sparse GPs}\label{app:complexity}
  
Assuming that the size of $\fout$, $\fin$ and $u$ are $n_g$, $n_l$ and $m$ respectively, the complexity of fitting a model in the first and second stages are  $\mathcal{O}(m^2n_g)$ and $\mathcal{O}((m+n_l)^3)$ respectively. Predicting the mean and variance takes the complexity of $\mathcal{O}(m+n_g)$ and $\mathcal{O}((m+n_l)^2)$. Normally we have \mbox{$m\ll n_g \ll n_l$}, although the posterior distribution needs to be computed twice, we can still save a lot of resources by introducing LGPGA without loss of precision. Additionally, a sparse GP can only be trained with $\fin$ and $\fout$ jointly, which might underfit $\fin$. Later we will show how different components of LGPGA emphasise different parts of training data.

    \begin{figure}[ht]
        \centering
        \begin{subfigure}[b]{0.48\textwidth}
        \centering
          \includegraphics[width=1.0\textwidth]{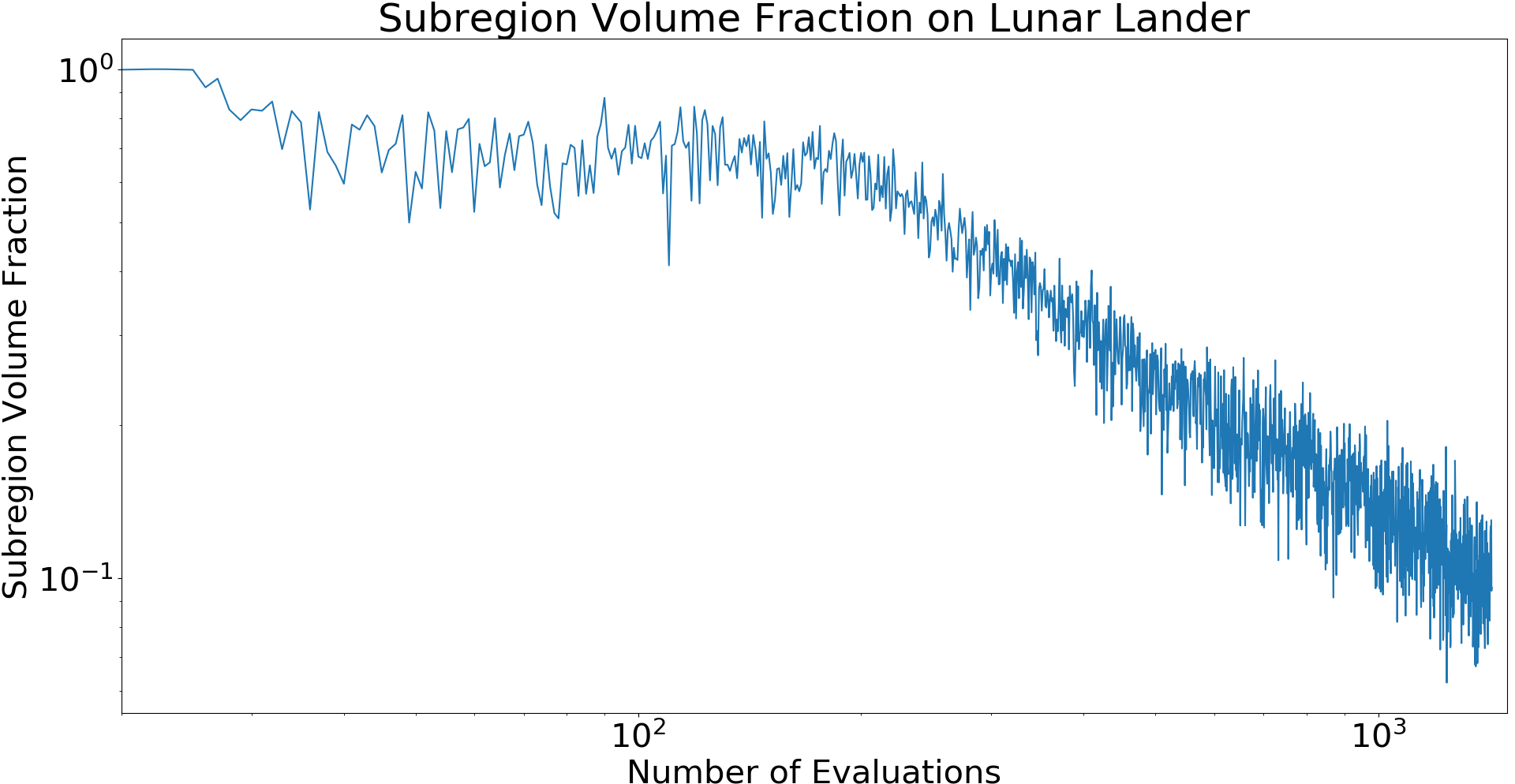}
        \end{subfigure}
        \begin{subfigure}[b]{0.48\textwidth}
        \centering
          \includegraphics[width=1.0\textwidth]{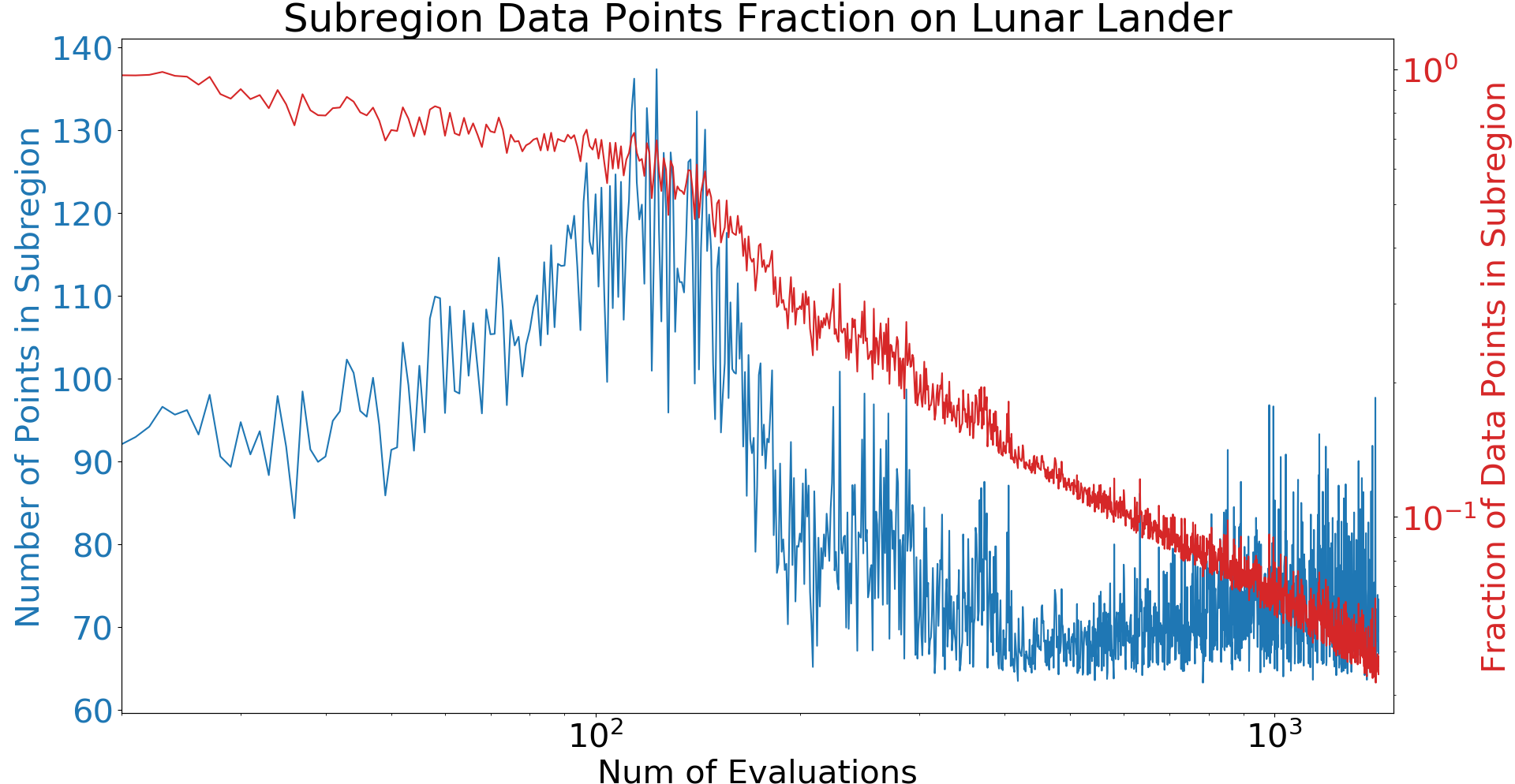}
        \end{subfigure}
        \caption{Top: fraction of subregion volume versus the entire region. Bottom: number of points inside subregion versus number of evaluated points.}
        \label{fig:num_points_in_ss}
    \end{figure}
  
  Figure \ref{fig:num_points_in_ss} illustrates how the number of data points varies as the number of evaluations grows. Here we take the evaluation result on the lunar lander problem as an example, where $n_{dims} = 14$ and thus the minimal number of points inside the subregion is $70$. We could see that the number of points inside the subregion stays nearly constant as the number of evaluations grows. The subregion quickly shrinks after about 100 evaluations and hence our local model could focus more on the most promising region. However, this trend does not hold for the following evaluations. The subregion only shrinks smoothly in the following evaluations, which will prevent BOinG from converging to a local minimum too early. 
  
\begin{figure}[ht]
        \centering
        \begin{subfigure}[b]{0.48\textwidth}
        \includegraphics[width=1.0\textwidth]{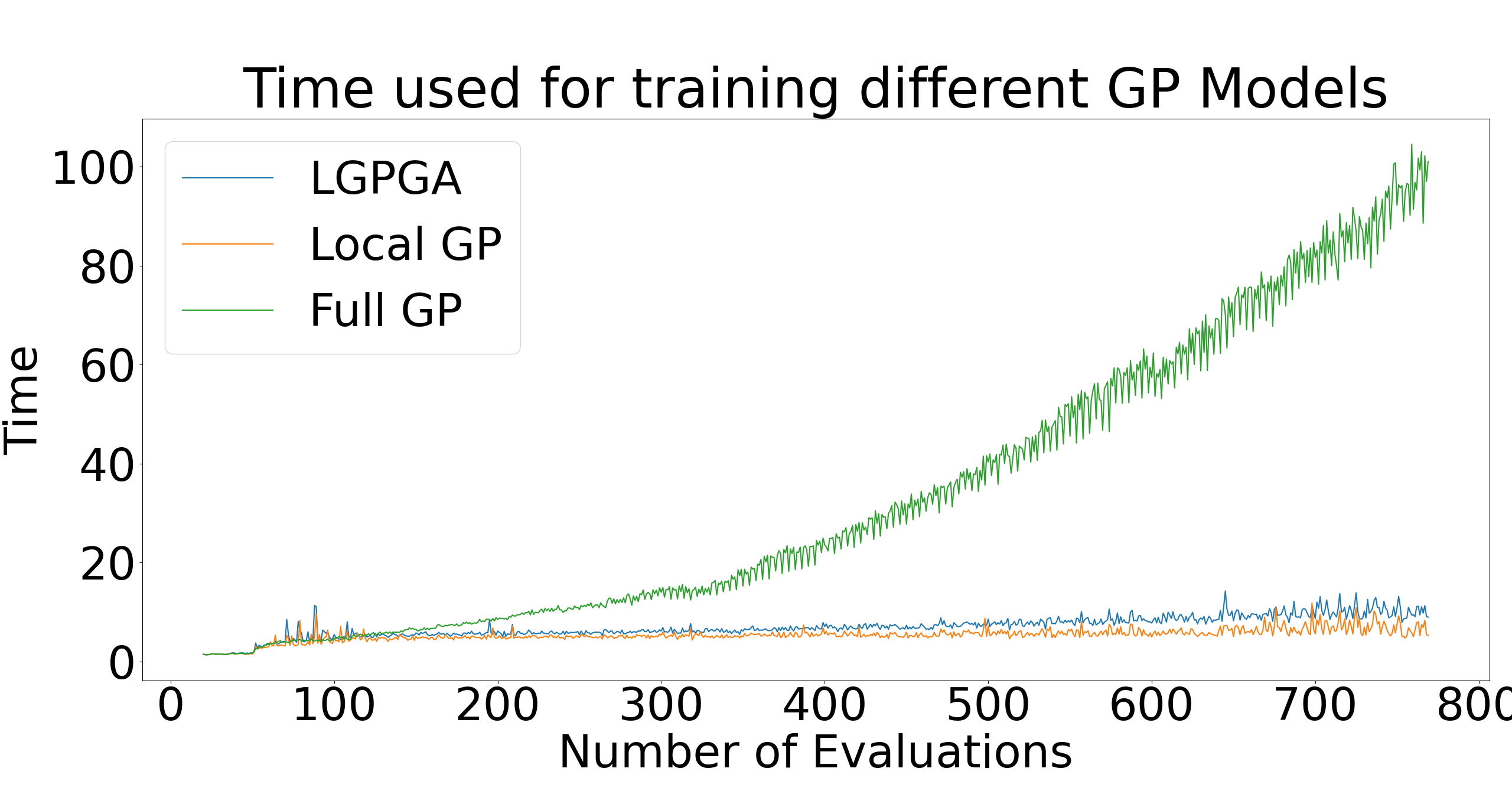}
        \end{subfigure}
        \begin{subfigure}[b]{0.48\textwidth}
        \includegraphics[width=1.0\textwidth]{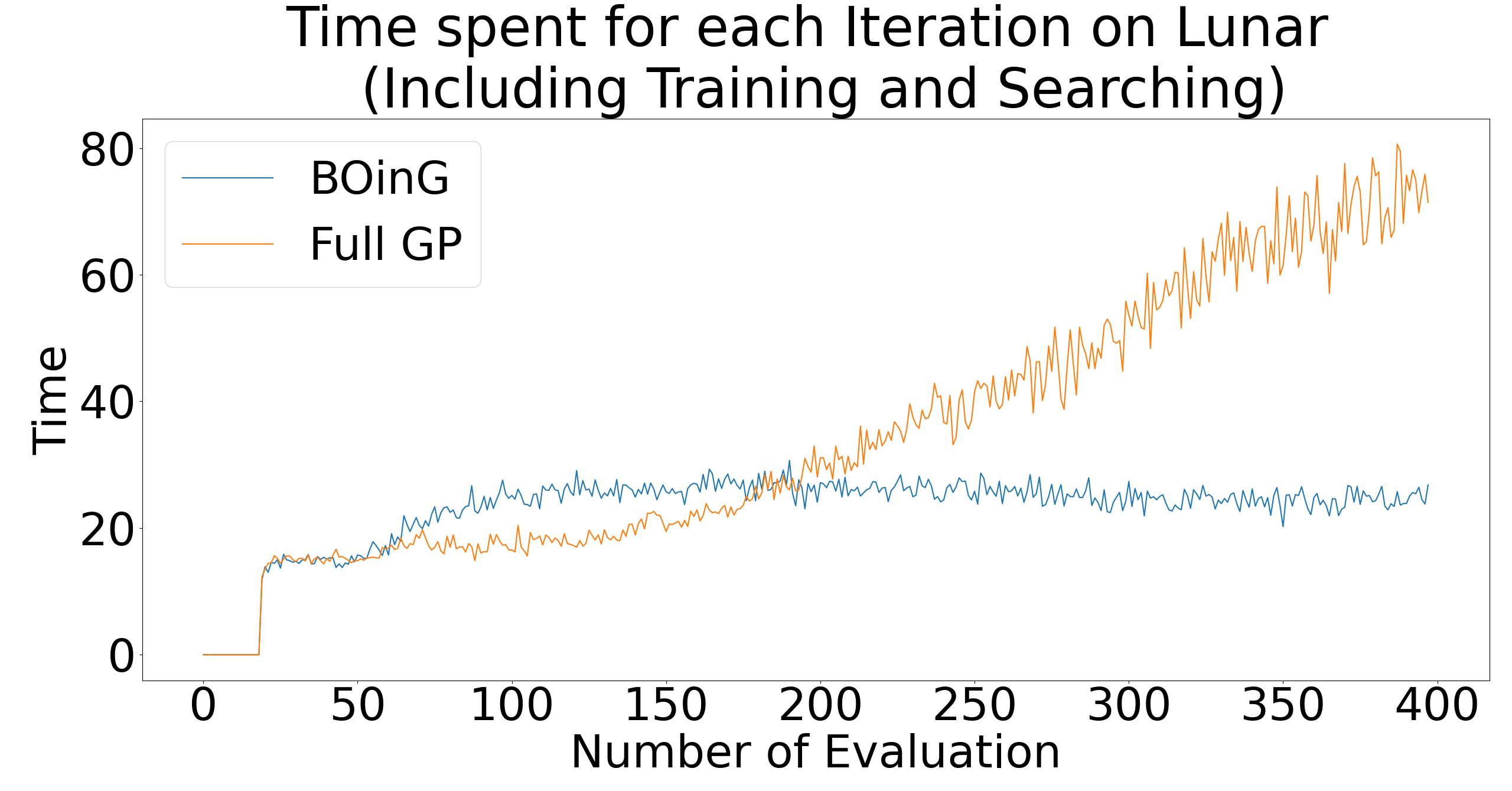}
        \end{subfigure}
        \begin{subfigure}[b]{0.48\textwidth}
        \includegraphics[width=1.0\textwidth]{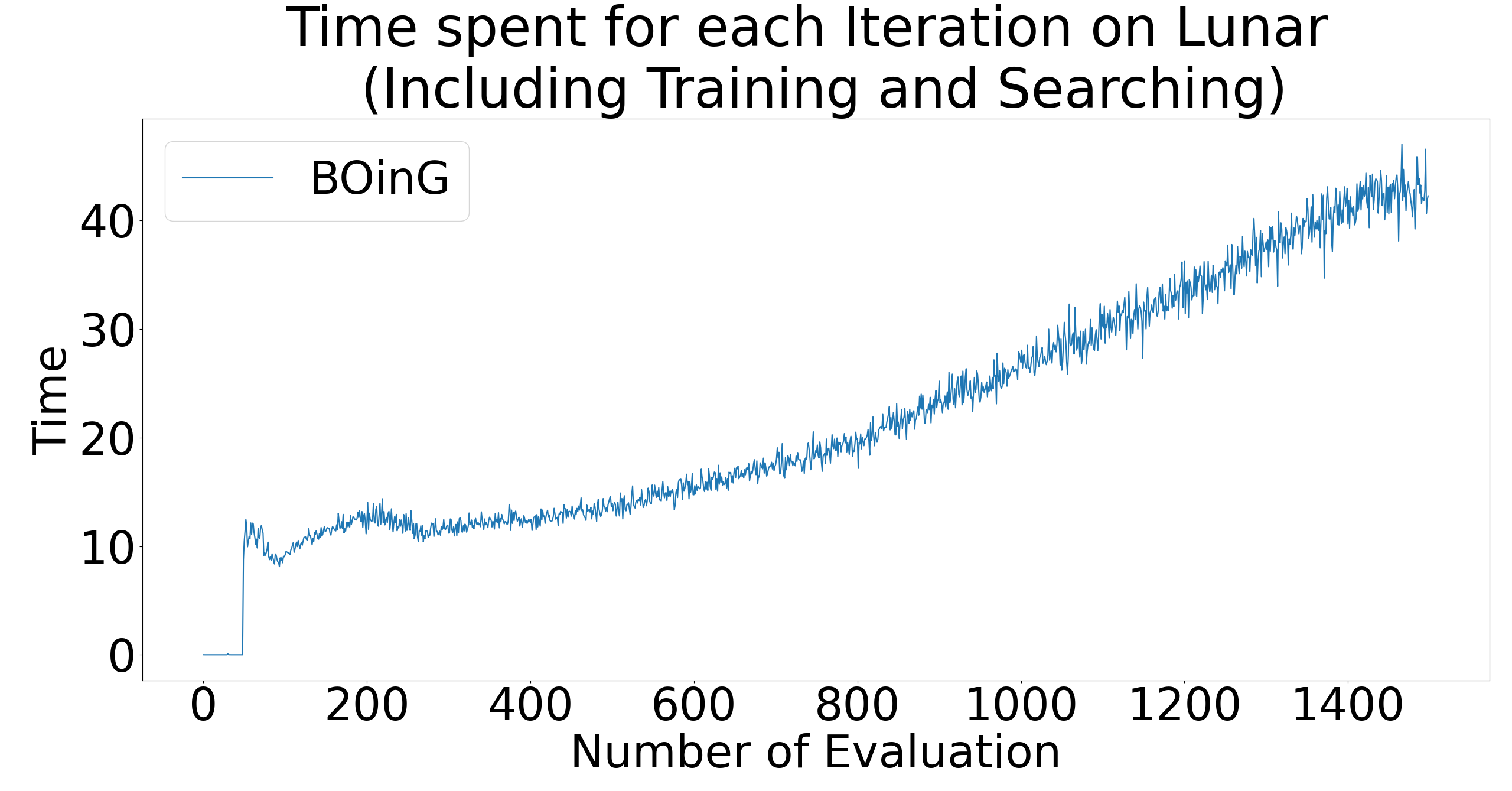}
        \end{subfigure}
        \caption{Scalability Analysis, we note that we optimize GP's hyerparameters with L-BFGS~\citep{liu-lbfgs89} optimizer and repeat optimization for 10 repetitions with different initial points on Lunar Lander (14D). \textbf{Top}: time spent for training different GP models. \textbf{Middle}: Time Spent by BOinG and Full GP for each BO iterations on Ackley (10D).  \textbf{Bottom}: Time Spent for each BO iterations on Lunar Lander (14D).}
        \label{fig:psgp_opt_time}
    \end{figure}

  Figure \ref{fig:psgp_opt_time} illustrates the time spent for a BOinG iteration as the number of evaluations grows. Starting from the 200th evaluations on ackley 10D, BOinG takes less time in each iteration compared to a full GP. From the right plot in Figure~\ref{fig:psgp_opt_time}, we could see that BOinG scales nearly linearly as the number of evaluations grows until up to 1500 evaluations.  
      
  Next we evaluate the saved resources and the additional overload that LGPGA brings to global and local GP models. Again, we train $3$ different models on the data that we obtained when we optimize the hyperparameters on lunar task: full GP denotes a GP model that is trained on all the previous evaluated points; local GP denotes a model that is trained only with the points inside the subregion; and finally our LGPGA model. Here, the GP's hyperparameters are optimized with 10 repetitions. The result is illustrated in the top part of Figure~\ref{fig:psgp_opt_time}. Compared to a full GP model, LGPGA requires much less resources and scales nearly linearly as the number of the previous evaluations grows.

  \section{How LGPGA Guides the Search in Subspaces}
\begin{figure}[h]
    \centering
    \begin{subfigure}[b]{0.48\textwidth}
    \centering
    \includegraphics[width=1.0\textwidth]{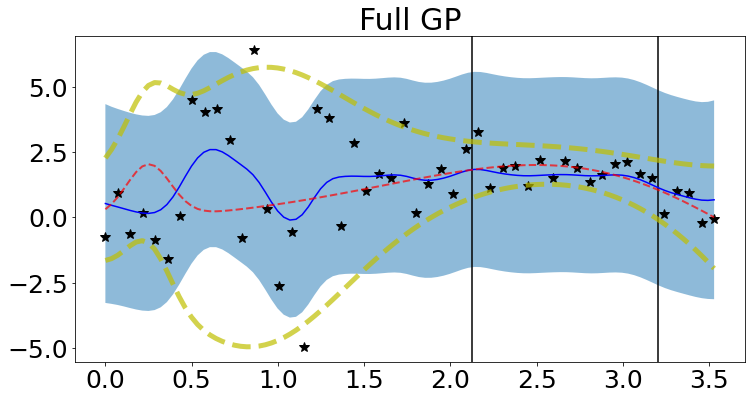}
    \end{subfigure}\hfill
    \begin{subfigure}[b]{0.48\textwidth}
    \centering
    \includegraphics[width=1.0\textwidth]{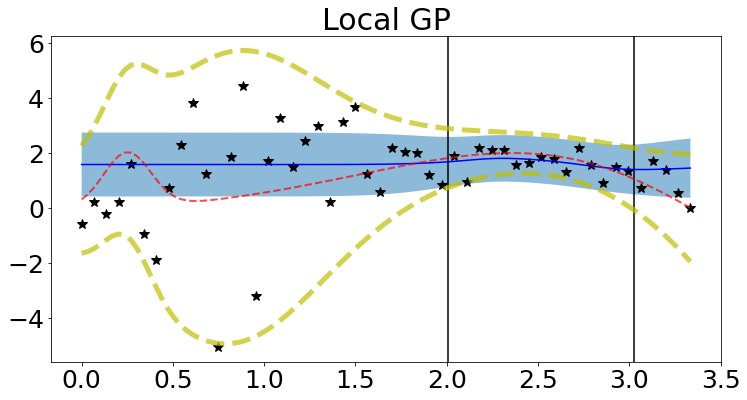}
    \end{subfigure}\hfill
    \begin{subfigure}[b]{0.48\textwidth}
    \centering
    \includegraphics[width=1.0\textwidth]{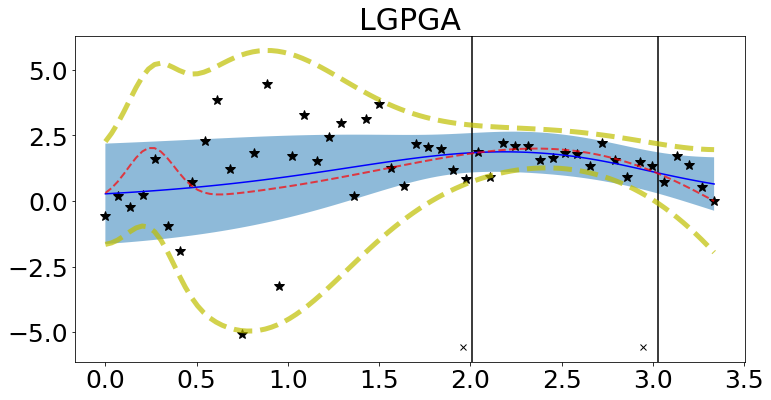}
    \end{subfigure}
    \begin{subfigure}[b]{0.48\textwidth}
    \centering
    \includegraphics[width=1.0\textwidth]{images/psgp/psgp_legend_twosides.png}
    \end{subfigure}
    \caption{Predicted mean and variance of a \textbf{Top}: full GP \textbf{Middle}:  local GP \textbf{Bottom}: LGPGA}
    \label{fig:res_psgp}
\end{figure}

  In Section 3.2, we used a toy example to show how LGPGA handles heteroscedastic noise. The data is generated according to~\cite{yuan2004} and follows a normal distribution with mean: $\mu (x) = 2(\exp (-30(x-1/4)^2)) + \sin (\pi x^2)$ and variance $\sigma^2 (x) = \exp (2\sin (2\pi x))$. This distribution has low noise level with larger $x$ values and high noise level with smaller $x$ values. Here we only use our model to fit the distribution of the right side. We randomly sample 50 points from $[0,1]$ and select the points indices from $35$ to $45$ as the points inside the subregion. The predicted mean and variances are illustrated in Figure~\ref{fig:res_psgp}. Fitting a GP on the entire data distribution will not exactly describe the noise on the right part of the data distribution. The GP fitting only the data points inside the subregion and our LGPGA describe better the heteroscedastic noise inside the subregion. 
  
  Additionally,  we illustrate how LGPGA influences the acquisition function value landscape and guides the search. 
\begin{figure*}[h]
    \centering
    \begin{subfigure}[b]{0.48\textwidth}
    \centering
    \includegraphics[width=1.0\textwidth]{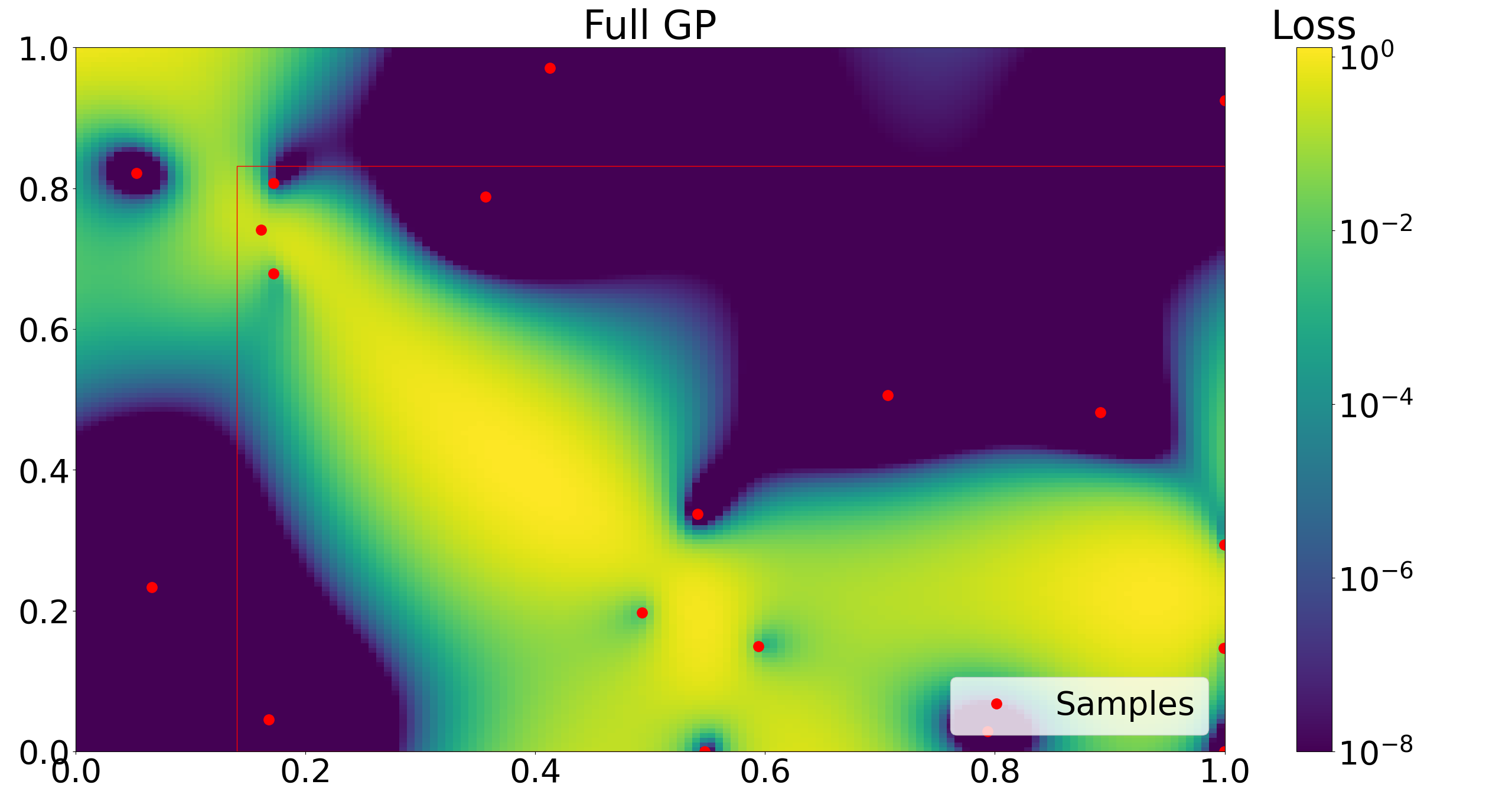}
    \end{subfigure}
    \begin{subfigure}[b]{0.48\textwidth}
    \centering
    \includegraphics[width=1.0\textwidth]{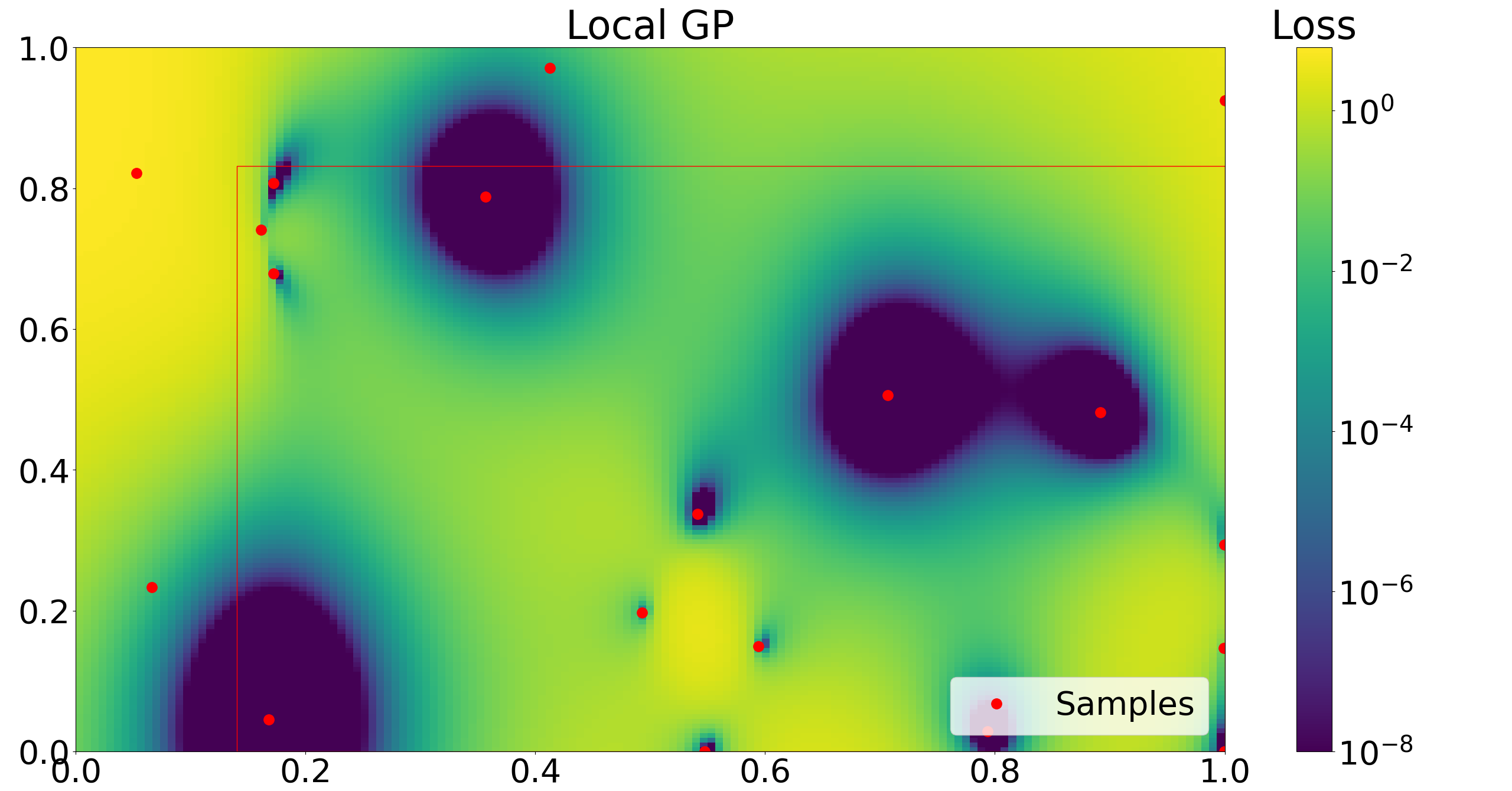}
    \end{subfigure}
    \begin{subfigure}[b]{0.48\textwidth}
    \centering
    \includegraphics[width=1.0\textwidth]{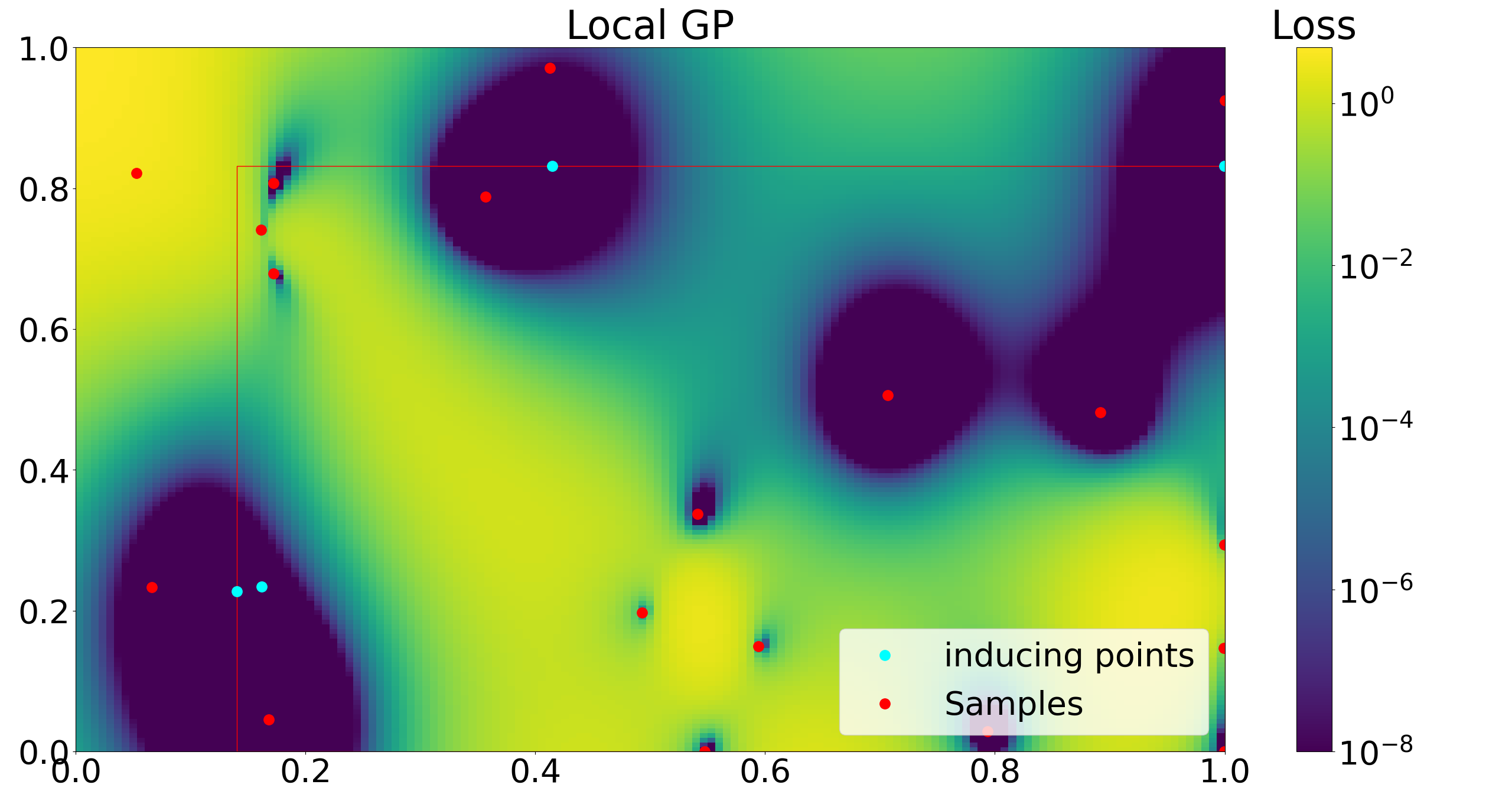}
    \end{subfigure}
    \begin{subfigure}[b]{0.48\textwidth}
    \centering
    \includegraphics[width=1.0\textwidth]{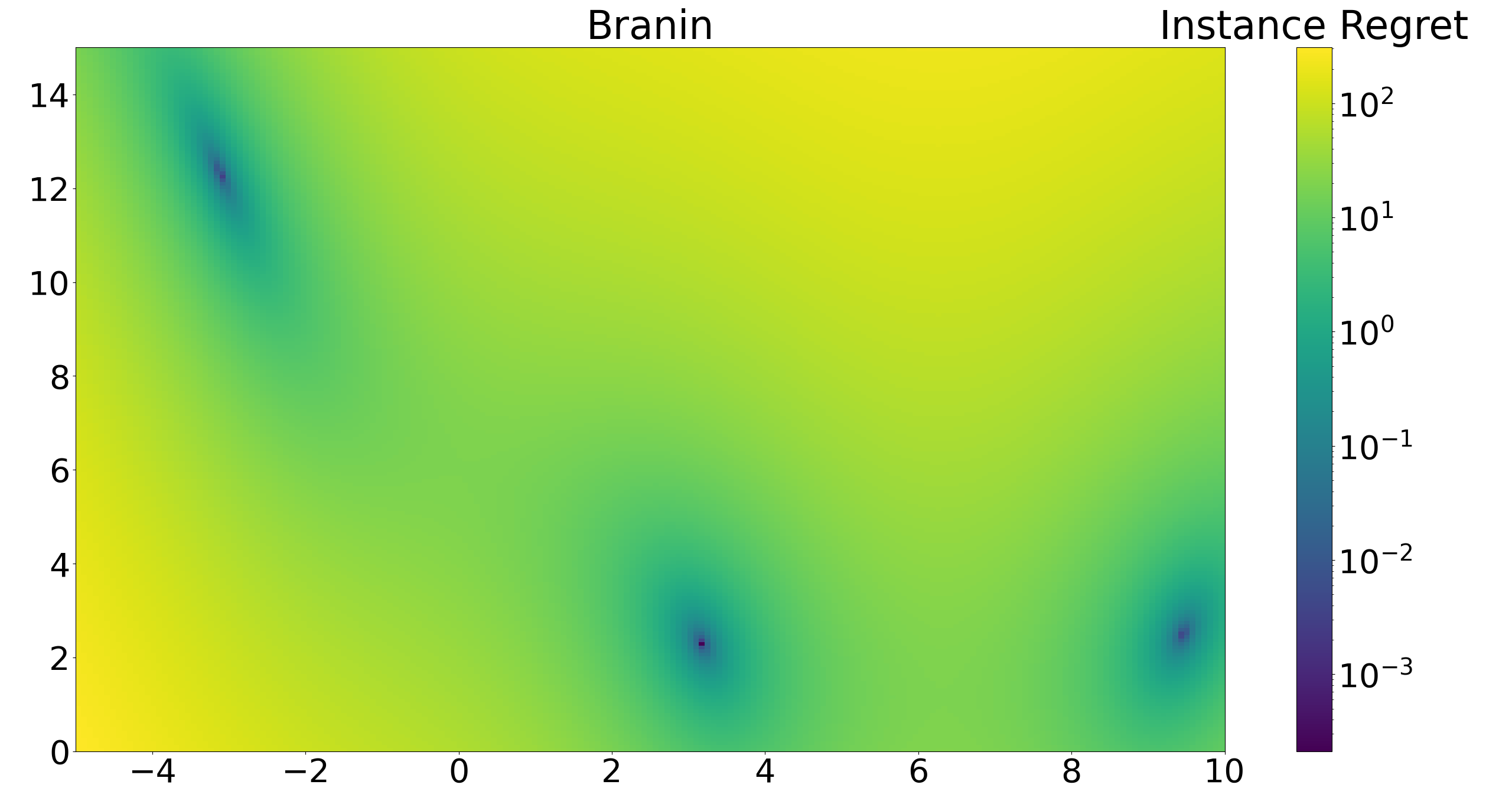}
    \end{subfigure}
    \caption{Predicted EI value of different GP models, red rectangles are the boundary of the subregion. From top left to bottom right: a GP model that is trained on all the previous evaluations; a GP model that is trained to only fit the distribution inside the subregion (the red rectangle); a LGPGA model trained with all the previous evaluations; the loss landscape of branin function.}
    \label{fig:ei_dis}
\end{figure*}

Similar to section \ref{sec:scalbility}, we train full GP, local GP and LGPGA on the same data distribution. The subregion is bounded by the rectangle. Without knowing the information of the points outside the subregion, the local GP will have a large chance to sample the points on the top right of the subregion for exploration. However, the high loss of the samples on the top right indicates that it might not be a good choice to sample a new point in this direction, as illustrated by the acquisition value loss landscape of the full GP model. However, LGPGA optimizes its inducing points to approximate the distribution outside the subregion and we see that two inducing points are located on the bottom left of the subregion while another two lay on the top side. Thus we could avoid unnecessary exploration on the bottom left and focus more on the region near the optimum or the direction that is still not fully explored. 

%On the contrast, a Full GP makes use of all the points that are previously evaluated. However, the previously evaluated points are not uniformly distributed: they tend to gather towards the incumbent values due to sampling bias. a full GP might no longer be the best fit for such sorts of distribution, we illustrate a toy example here:

\section{Will BOinG+ Give Better Suggestions compared to TuRBO?}
BOinG+ switches between TuRBO and BOinG randomly according to their failure counts. In the ablation study, we show that BOinG+ has a better final performance compared to different variation of TuRBO. However, it is still unclear where the incumbent configuration comes from, i.e., if BOinG+ works as we expect: explore with TuRBO and exploit with BOinG? To answer this question, we show the fraction of the incumbents' origin as the number of evaluations grows on the two tasks where BOinG+ is applied. 

\begin{figure}[ht]
    \centering
    \begin{subfigure}{0.48\textwidth}
    \centering
    \includegraphics[width=1.0\textwidth]{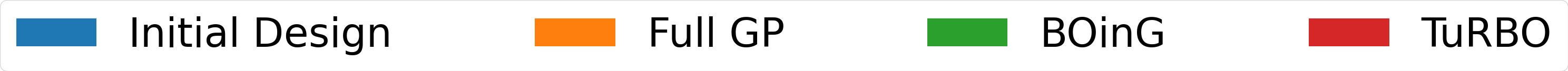}
    \end{subfigure}
    \begin{subfigure}{0.48\textwidth}
    \centering
    \includegraphics[width=1.0\textwidth]{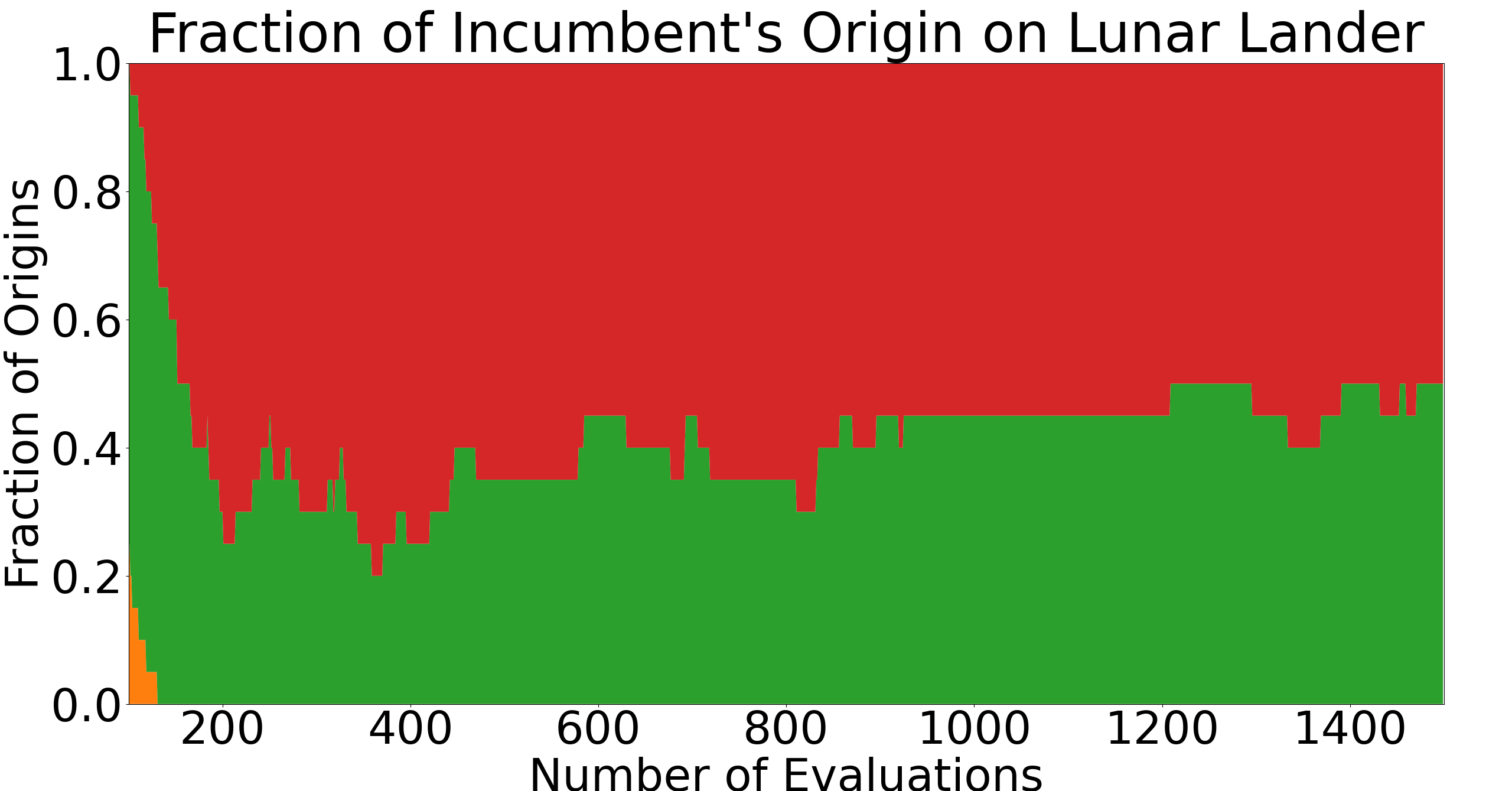}
    \end{subfigure}
        \begin{subfigure}{0.48\textwidth}
    \centering
    \includegraphics[width=1.0\textwidth]{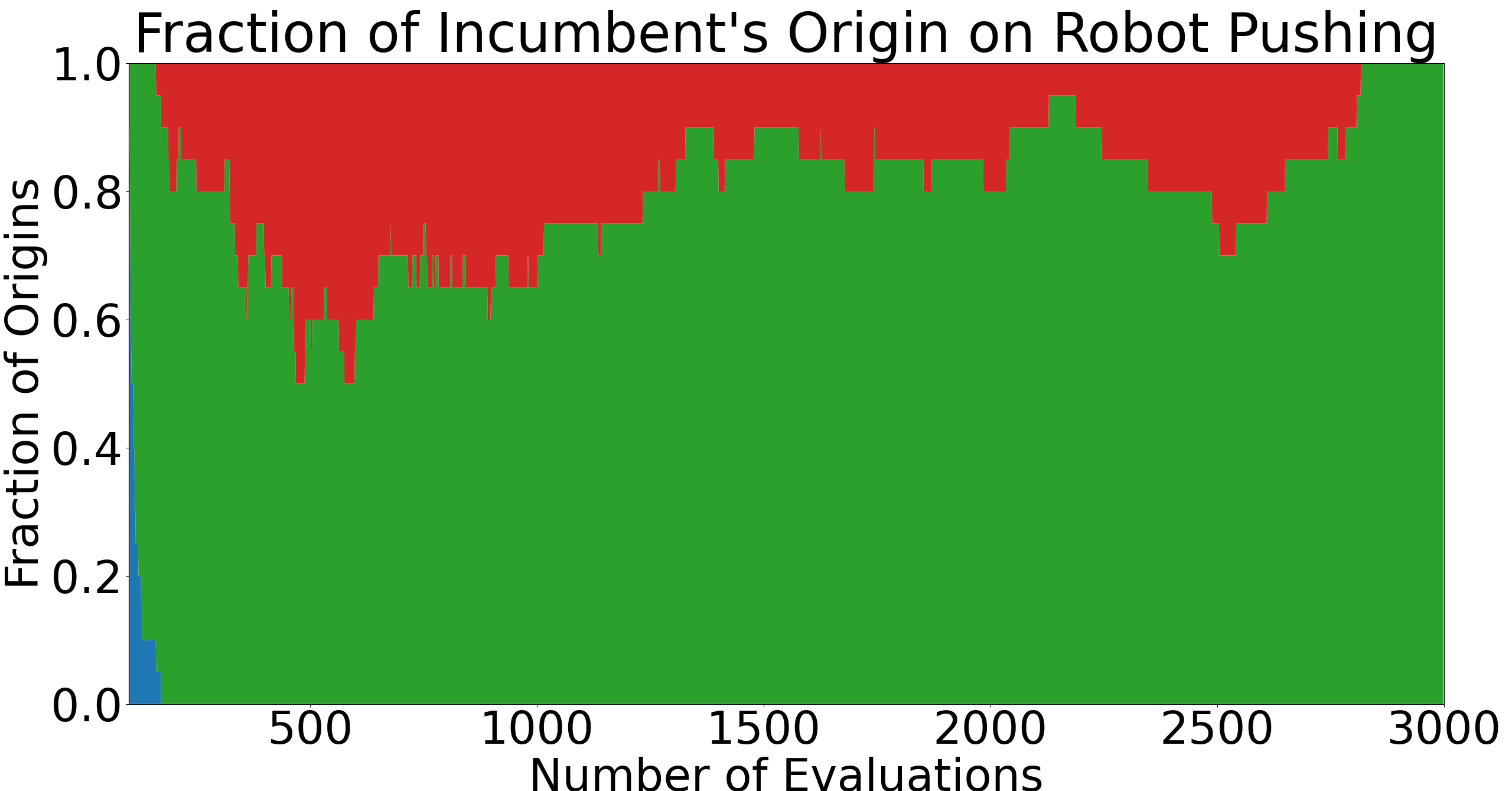}
    \end{subfigure}
    \caption{Share of different incumbents' origins}
    \label{fig:area_boing}
\end{figure}

The results are shown in Figure~\ref{fig:area_boing}. Depending on the task, the performances are quite different. This shows that BOinG+ could adjust to different sorts of landscapes and not get stuck at one single optimizer. The share of TuRBO reaches peak at roughly 500 evaluations and then more incumbents are suggested by BOinG. Thus, as expected, TuRBO explores more in the mid term of the optimization process and thus finds more incumbents during this period; while BOinG exploits more in the most promising regions and thus gives more incumbents at the end.

\bibliography{strings, lib, bibtex, proc}